\documentclass{article}

\usepackage{PRIMEarxiv}
\usepackage[labelfont=bf]{caption}
\usepackage{amsmath}
\usepackage[utf8]{inputenc} 
\usepackage[T1]{fontenc}    
\usepackage{hyperref}       
\usepackage{url}            
\usepackage{booktabs}       
\usepackage{amsfonts}       
\usepackage{nicefrac}       
\usepackage{microtype}      
\usepackage{lipsum}
\usepackage{fancyhdr}       
\usepackage{graphicx}       
\usepackage[affil-it]{authblk}
\usepackage{afterpage}   
\usepackage{multirow}
\allowdisplaybreaks 
\usepackage{makecell}
\vspace{-15mm}

\newgeometry{textwidth=7in}

\title{Target Specific De Novo Design of Drug Candidate Molecules with Graph Transformer-based Generative Adversarial Networks}

\author[1,2\thanks{Equally contributing co-first authors.}]{Atabey Ünlü}
\author[1,2\textsuperscript{*}]{Elif Çevrim}
\author[1,3\textsuperscript{*}]{Melih Gökay Yiğit}
\author[4,5]{Ahmet Sarıgün}
\author[1]{Hayriye Çelikbilek}
\author[6]{Osman Bayram}
\author[7]{Deniz Cansen Kahraman}
\author[8,9]{Abdurrahman Olğaç}
\author[10,11]{Ahmet Sureyya Rifaioğlu}
\author[8]{Erden Banoğlu}
\author[1,2,12\thanks{Corresponding author email address: tuncadogan@gmail.com}]{Tunca Doğan}

\affil[1]{Biological Data Science Lab, Dept. of Computer Engineering, Hacettepe University}
\affil[2]{Dept. of Bioinformatics, Graduate School of Health Sciences, Hacettepe University}
\affil[3]{Dept. of Computer Engineering, Middle East Technical University}
\affil[4]{Dept. of Chemistry, Middle East Technical University}
\affil[5]{Dept. of Physics, Middle East Technical University}
\affil[6]{Dept. Of Artificial Intelligence Engineering, Bahcesehir University}
\affil[7]{Cancer Systems Biology Lab, Graduate School of Informatics, Middle East Technical University}
\affil[8]{Dept. of Pharmaceutical Chemistry, Faculty of Pharmacy, Gazi University}
\affil[9]{Laboratory of Molecular Modeling, Evias Pharmaceutical R\&D Ltd.}
\affil[10]{Dept. of Electrical and Electronics Engineering, İskenderun Technical University}
\affil[11]{Institute for Computational Biomedicine, Heidelberg University}
\affil[12]{Dept. of Health Informatics, Institute of Informatics, Hacettepe University, 06800, Ankara, Turkey}
\begin{document}

\maketitle


\begin{abstract}
Discovering novel drug candidate molecules is one of the most fundamental and critical steps in drug development. Generative deep learning models, which create synthetic data given a probability distribution, offer a high potential for designing de novo molecules. However, to be utilisable in real-life drug development pipelines, these models should be able to design drug-like and target-centric molecules. In this study, we propose an end-to-end generative system, DrugGEN, for the de novo design of drug candidate molecules that interact with intended target proteins. The proposed method represents molecules as graphs and processes them via a generative adversarial network comprising graph transformer layers. The system is trained using a large dataset of drug-like compounds and target-specific bioactive molecules to design effective inhibitory molecules against the AKT1 protein, which is critically important in developing treatments for various types of cancer. We conducted molecular docking and dynamics to assess the target-centric generation performance of the model, as well as attention score visualisation to examine model interpretability. In parallel, selected compounds were chemically synthesised and evaluated in the context of in vitro enzymatic assays, which identified two bioactive molecules that inhibited AKT1 at low micromolar concentrations. These results indicate that DrugGEN’s de novo molecules have a high potential for interacting with the AKT1 protein at the level of its native ligands. Using the open-access DrugGEN codebase, it is possible to easily train models for other druggable proteins, given a dataset of experimentally known bioactive molecules.
\end{abstract}

\vspace{2em}


\section{Introduction}

The development of a new drug is a long-term and costly process. It entails the identification of bioactive compounds against predefined biomolecular targets as one of its initial and most essential steps. Advancements in high-throughput screening technologies now enable the simultaneous screening of tens of thousands of compounds. However, it is still impossible to analyse the entire chemical and target spaces due to their huge sizes \cite{Rifaioglu_Atas_Martin_Cetin-Atalay_Atalay_Doğan_2019}, which usually prevents the discovery of the best candidate molecules. The majority of “non-ideal” drug candidates are discontinued in the late stages of the development process, such as clinical trials, due to high toxicity or low efficacy, which is the primary reason for the low success rates lately observed in drug development \cite{Paul_Mytelka_Dunwiddie_Persinger_Munos_Lindborg_Schacht_2010}. \\

Small-molecule drugs developed so far exhibit low levels of structural diversity and can only target certain protein families \cite{Bhisetti_Fang_2022}. It is possible to argue that a similar diversity problem is also valid for large virtual chemical libraries. Thus, there is a need for truly novel, i.e. structurally diverse, small molecule drug candidates to target druggable proteins in the human proteome, including their clinically significant variants  \cite{Elton_Boukouvalas_Fuge_Chung_2019}. Within the vast theoretical space of small molecules, estimated to be between  $10^{30}$ and $10^{60}$ \cite{Walters_2018} , compounds that have the potential to interact specifically and effectively with each targetable biomolecule may exist. The main challenge is identifying the correct molecular structures within this unexplored territory. An approach called "de novo drug design" is utilised for this. Its purpose is to design truly novel candidate molecules that are not mere derivatives of previously established molecular structures, especially to overcome hard-to-target biomolecules \cite{Mouchlis_Afantitis_Serra_Fratello_Papadiamantis_Aidinis_Lynch_Greco_Melagraki_2021}. \\

To address problems associated with conventional de novo drug design, such as long development durations, elevated costs, and a high number of unknown variables regarding the efficacy and safety of the designed compounds, artificial intelligence (AI) driven methods (e.g., deep generative modelling/AI) are starting to emerge in the field \cite{Kingma_Welling_2013,Gómez-Bombarelli_Wei_Duvenaud_Hernández-Lobato_Sánchez-Lengeling_Sheberla_Aguilera-Iparraguirre_Hirzel_Adams_Aspuru-Guzik_2018,Goodfellow_Pouget-Abadie_Mirza_Xu_Warde-Farley_Ozair_Courville_Bengio_2020,De_Cao_Kipf_2018,Zou_Yu_Hu_Zhao_Shi_2023,Mahmood_Mansimov_Bonneau_Cho_2021,Ho_Jain_Abbeel_2020,Sohl-Dickstein_Weiss_Maheswaranathan_Ganguli_2015, Hoogeboom_Satorras_Vignac_Welling_2022,peng2022pocket2mol,schneuing2022structure}. A short review of this topic can be found in the Supplementary Material (S1). \\

Deep generative models have also been used to design molecules with desired properties via conditioning the training or the prediction procedure(s). Most of these models have utilised condition vectors as a tool for property injection into the generative process.  VAEs \cite{Mitton_Senn_Wynne_Murray-Smith_2021,Nemoto_Kaneko_2023,Richards_Groener_2022}, GANs \cite{De_Cao_Kipf_2018,Kadurin_Nikolenko_Khrabrov_Aliper_Zhavoronkov_2017,Xie_Valiente_Kim_2023}, sequence-based (language) models \cite{Arús-Pous_Johansson_Prykhodko_Bjerrum_Tyrchan_Reymond_Chen_Engkvist_2019,Bagal_Aggarwal_Vinod_Priyakumar_2022,Blaschke_Arús-Pous_Chen_Margreitter_Tyrchan_Engkvist_Papadopoulos_Patronov_2020,Wang_Gao_Han_Li_Chen_Rodríguez_Patón_Wang_Zheng_2023,yang2023cmgn}, geometric models (e.g., 3D molecule generation) \cite{zhang2023resgen, guan20233d, peng2022pocket2mol} and diffusion models \cite{Hoogeboom_Satorras_Vignac_Welling_2022, Guan_Qian_Peng_Su_Peng_Ma_2023} have been utilised for conditional molecule generation tasks. Reinforcement learning (RL) has also been employed for this purpose, with reward-penalty functions guiding models towards desired molecular characteristics in the respective latent space  \cite{Blaschke_Arús-Pous_Chen_Margreitter_Tyrchan_Engkvist_Papadopoulos_Patronov_2020,Perron_Mirguet_Tajmouati_Skiredj_Rojas_Gohier_Ducrot_Bourguignon_Sansilvestri‐Morel_Do_Huu_2022, zhou2019optimization}. These approaches result in optimised molecule generation; however, one of the fundamental objectives of drug design is developing small molecules that will physically interact with the desired target, and obtaining de novo molecules with optimised properties alone is insufficient to satisfy this goal. Although a few recent studies present prototype models \cite{Gebauer_Gastegger_Hessmann_Müller_Schütt_2022,Li_Zhang_Wang_Zou_Yang_Luo_Wu_Yang_Tian_Xu_2022,Liu_Luo_Uchino_Maruhashi_Ji_2022,Rozenberg_Freedman_2023,Shi_Singha_Srivastava_Pu_Ramanujam_Brylinski_2022,Uludoğan_Ozkirimli_Ulgen_Karalı_Özgür_2022,Wang_Hsieh_Wang_Wang_Weng_Shen_Yao_Bing_Li_Cao_et_al._2022,Zhang_Li_Xing_Yuan_He_Sun_2023}, AI-driven target-centric drug design is a highly novel and understudied area with great potential to contribute to rational drug design. \\

In this study, we propose DrugGEN, a new de novo small molecule design system, an end-to-end framework, that generates target-centric drug-like molecules using the GAN and transformer architectures \cite{Vaswani_Shazeer_Parmar_Uszkoreit_Jones_Gomez_Kaiser_Polosukhin_2017}, incorporating graph representation learning \cite{Kipf_Welling_2016}. The workflow of the study is depicted in Figure 1. The process (i) starts with the preparation of small molecule datasets (including their bioactivities/interactions) and their encoding as graphs to be used as training and test data (Figure 1A), (ii) proceeds with the design and implementation of the DrugGEN model together with the training and evaluation of the system in the context of performance comparison with the state-of-the-art and an ablation study (Figure 1B), and (iii) concludes with the downstream analysis of generated molecules and selection of the most promising candidates to effectively target the selected protein (Figure 1C). In item iii, we generated and evaluated novel inhibitors for the “RAC-alpha serine/threonine-protein kinase” (AKT1) protein (Figure S1). Additional information about AKT1 is provided in Supplementary Material (S2).

\begin{figure*}
\vspace{-3em}
    \centering
    \includegraphics[width=\textwidth,clip=True]{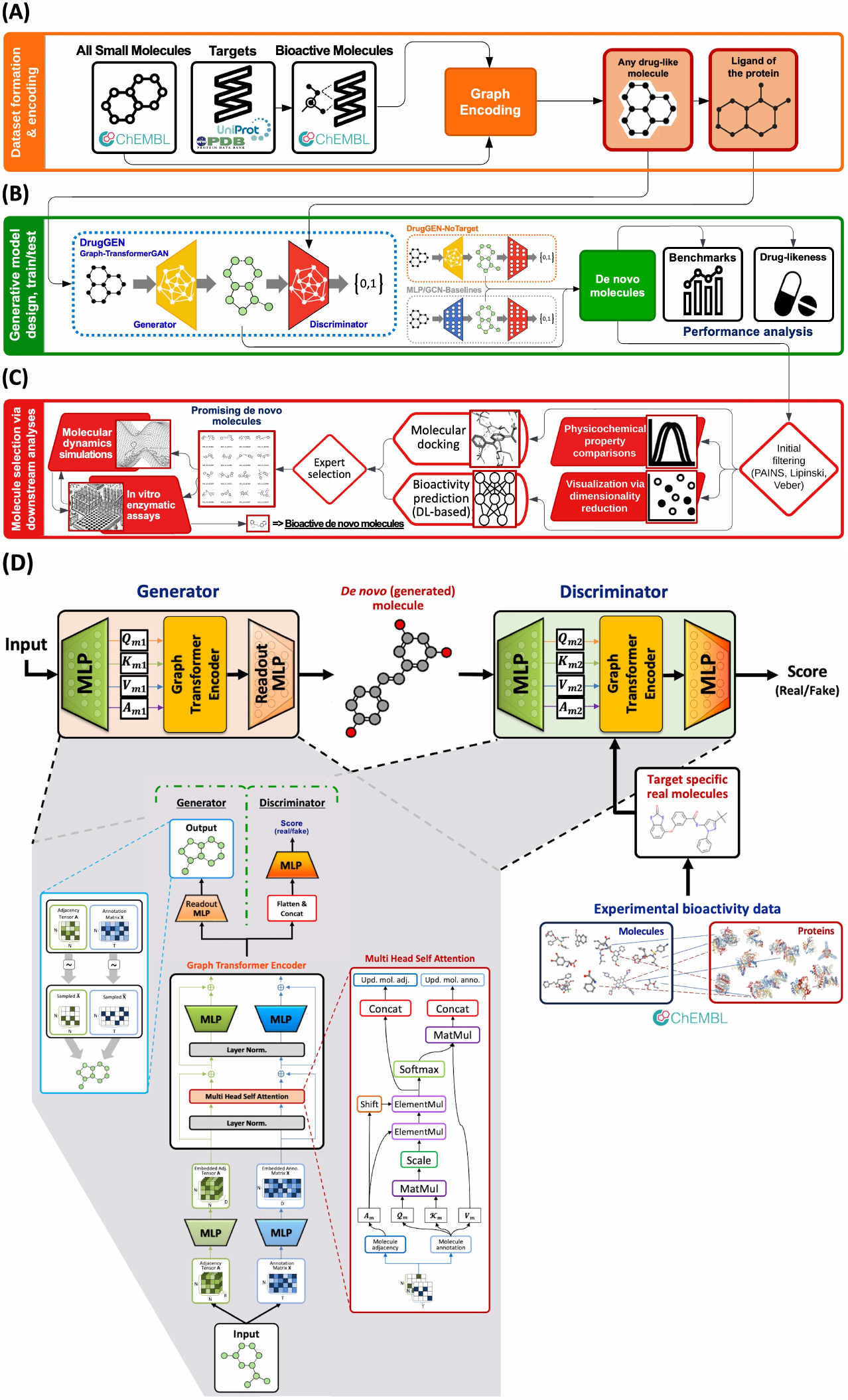}

    \label{fig:img1}
\end{figure*}

\clearpage

\captionof{figure}{The workflow of the study. \textbf{(A)} Preparation of datasets including molecules and bioactivities, together with the graph-based encoding of samples in datasets, \textbf{(B)} the graph transformer GAN-based architecture of DrugGEN, model training and evaluation via subjecting de novo molecules to fundamental benchmarks and drug-likeness-related metrics, and \textbf{(C)} subsequent selection of molecules via a series of in silico experiments to identify promising candidates that effectively target the selected protein, \textbf{(D)} the schematic representation of the architecture of the DrugGEN model with powerful graph transformer encoder modules in both generator and discriminator networks. The generator module transforms the given input into a new molecular representation. The discriminator compares the generated de novo molecules to the known inhibitors of the given target protein, scoring them for their assignment to the groups of “real” and “fake” molecules (abbreviations; MLP: multi-layer perceptron, Norm: normalisation, Concat: concatenation, MatMul: matrix multiplication, ElementMul: element-wise multiplication, Upd: updated, N: total \# of heavy atoms, T: \# of atom types, B: \# of bond types, D: hidden dimension size).
} 

\vspace{1.5em}

\section{Results}

\subsection{The Overview of DrugGEN}

DrugGEN aims to learn the distributions of the physicochemical properties and topological attributes of drugs and drug candidate compounds from the given data to generate drug-like small molecules that are valid and novel. The discriminator network takes the de novo molecules designed by the generator and compares them with the known (real) bioactive ligands (inhibitors) of the selected target protein so that the structural differences between the de novo and real molecules can be learned by the model. This information is then shared with the generator network in an adversarial setting. This approach directs the generation process toward ligands interacting with the selected target. One of the technical novelties in DrugGEN is the incorporation of graph transformers into the generator and discriminator modules of the generative adversarial network (GAN) (Figure 1D). The reason behind employing graph transformers, as opposed to conventional architectures, e.g., the multi-layer perceptron (MLP), is to overcome problems related to handling the complexity of the target-centric generation process, especially considering the large sizes of drug-like molecules, by effectively learning the long-range dependencies and interactions between atoms/bonds. Graph transformers alone have been used in the literature for molecular property prediction \cite{li2021effective, rong2020self, li2022kpgt} however, to the best of our knowledge, this is the first study to employ molecular graph transformers within a GAN pipeline. Furthermore, we modified the attention mechanism of the graph transformer in both the generator and discriminator modules by enhancing the incorporation of edge information (i.e., atomic bonds), thereby increasing the influence of graph connectivity during molecular learning (please see Methods for details). Finally, in traditional GANs, the input is typically random noise; however, in our model, as in some studies reported in the literature (e.g., CycleGAN \cite{zhu2017unpaired}, and DiscoGAN \cite{kim2017learning}), real samples are used as input. Due to incorporating an adversarial training framework, we have characterised the architecture of our model as a GAN derivative. \\

DrugGEN is trained with small molecule data, including random drug-like molecules and known target-specific inhibitors (from ChEMBL), and applied for designing de novo ligands of (i) “RAC-alpha serine/threonine-protein kinase” (AKT1), and (ii) “Cyclin-dependent kinase 2” CDK2 (included as a second target to assess reproducibility), and its effectiveness was assessed through a series of downstream analyses (Figure 1B and C). The performance of DrugGEN was evaluated via known benchmarking metrics. DrugGEN’s molecules were also explored in comparison to real molecules via (i) physicochemical property value distribution plots and (ii) UMAP / t-SNE projections and visualisation in 2-D. Additionally, target-centric properties of the generated molecules were evaluated further via in silico experiments, such as molecular docking and deep learning-based drug-target interaction prediction. In our tests, we also compared DrugGEN with other generative models. Furthermore, we conducted an explainable AI-centric analysis via attention score visualisation to interpret the model’s output. We selected 30 promising candidate de novo molecules according to the results of these computational analyses and expert curation. Subsequently, we chemically synthesised five chosen molecules (with 95$\geq$\% purity) and experimented in vitro to validate the target specificity of the model. Finally, we performed molecular dynamics simulations with (i) the native ligand in the co-crystal structure of AKT1 (PDB ID: 4GV1\cite{Addie_Ballard_Buttar_Crafter_Currie_Davies_Debreczeni_Dry_Dudley_Greenwood_et_al._2013}), and (ii) chosen de novo generated molecules to compare their binding characteristics within the AKT1 binding site. Each of these analyses is described in detail in subsequent sections.  \\

\subsection{Overall performance evaluation and comparison}

We compared DrugGEN’s performance with methods from the molecular generative AI literature. For this, we generated 10,000 de novo molecules (using our trained model in the inference mode) and subjected these molecules to MOSES benchmarking \cite{Polykovskiy_Zhebrak_Sanchez_Lengeling_Golovanov_Tatanov_Belyaev_Kurbanov_Artamonov_Aladinskiy_Veselov_et_al._2020}. Here, we report the generative performance of models using the widely known metrics of validity, uniqueness, novelty, internal diversity (IntDiv), synthetic accessibility (SA) and quantitative estimate of drug-likeness (QED) (see “Performance metrics” in Methods for details). It is important to note that these benchmarking metrics only provide clues about the performance of a generative model on a fundamental level. They do not offer a comprehensive evaluation, which is a difficult task since the main aims of these models are different from each other (e.g., designing a valid molecule, optimising molecules over a physicochemical property, etc.). Target-specific molecule generation, which is the primary goal of DrugGEN, cannot be assessed by these metrics; therefore, this property has been evaluated at later steps via ML-based bioactivity prediction and physics-based analyses, such as molecular docking and dynamics. \\

In Table 1, DrugGEN’s performance is shown together with other models from the literature, categorised based on their objectives: non-targeted design models (e.g., REINVENT, BIMODAL, MolGPT, ORGAN, CMGN, EDM, QADD, MARS, MGM, molDQN, and STAGAN) and targeted design models (e.g., RELATION, TRIOMPHE-BOA, ResGen, TargetDiff, and Pocket2Mol). The competing models were selected based on the underlying algorithms and training datasets to cover a wide range of approaches. Here, we report the original results from the published results (unless stated otherwise) due to resource and time constraints and the fact that re-training could risk deviating from the model's optimised performance. Nevertheless, we retrained the equivariant diffusion model (EDM) \cite{Hoogeboom_Satorras_Vignac_Welling_2022} using the ChEMBL dataset since its original training data differed. Table 1 includes baseline metrics commonly used for evaluating generative models. Targeted design models were evaluated based on AKT1-specific molecule generation. Since this analysis has multiple metrics and numerous methods, it was not straightforward to make a generalised comparison. To solve this issue, we calculated an overall ranking by considering all the metrics together (where rank 1 is the best). For this, we divided the methods into three groups: the best, medium and low, and scored them with 1, 2 and 3, respectively, independently for each metric. The methods are ranked from lowest to highest average score. The underlying continuous metric values are considered to resolve ties in the score-based ranking. DrugGEN and DrugGEN-NoTarget became the top models in the targeted and non-targeted model rankings, respectively. Additional comments can be found in Supplementary Material (S3). \\

\renewcommand{\theadfont}{\fontsize{7}{11}}
\begin{table}
\setlength\tabcolsep{2pt}
    \caption{Molecule generation performance of DrugGEN and other methods: RELATION \cite{Wang_Hsieh_Wang_Wang_Weng_Shen_Yao_Bing_Li_Cao_et_al._2022},MGM \cite{Mahmood_Mansimov_Bonneau_Cho_2021}, MolGPT \cite{Bagal_Aggarwal_Vinod_Priyakumar_2022}, ORGAN \cite{Guimaraes_Sanchez-Lengeling_Outeiral_Farias_Aspuru-Guzik_2018}, QADD \cite{Fang_Pan_Shen_2023}, molDQN \cite{zhou2019optimization}, BIMODAL \cite{Grisoni_Moret_Lingwood_Schneider_2020}, MARS \cite{Xie_Shi_Zhou_Yang_Zhang_Yu_Li_2021}, REINVENT \cite{Olivecrona_Blaschke_Engkvist_Chen_2017}, ResGen \cite{zhang2023resgen}, CMGN \cite{yang2023cmgn},TRIOMPHE-BOA \cite{matsukiyo2023novo}, TargetDiff \cite{guan20233d}, Pocket2Mol \cite{peng2022pocket2mol}, and STAGAN \cite{Zou_Yu_Hu_Zhao_Shi_2023}, measured in terms of fundamental benchmarking metrics of validity, novelty, uniqueness (Uniq.), internal diversity (IntDiv), QED and the overall rank of the models, based on all given metrics. We directly obtained the scores of competing models from their respective articles, except for ResGen, for which we used the trained model to generate molecules and calculate metrics, and EDM, for which we utilised the ChEMBL dataset to train a new model. Performance values missing from the original studies and uncomputable by us are indicated by “-”. The ORGAN and STAGAN models are trained on a curated ZINC dataset, whereas the RELATION and MolDQN models utilise a combination of ChEMBL and ZINC data for training. The remaining models are trained exclusively on the ChEMBL data. Arrows show the direction in which the performance increases. The best performances are displayed in bold font, separately for non-targeted and targeted design model groups (all scores within 1\% of the highest value are included, as the differences among these close values are negligible). Metrics that contain a distribution of values are reported as mean ± SD, while metrics yielding a single value are reported as such.}
    \centering
    \fontsize{7}{11}\selectfont
    \makebox[\linewidth]{
    \begin{tabular}{ccccccccccc}
    \hline
    \thead{Model \\ objective} & \thead{Model \\ Name} & \thead{Input data \\ modality} & \thead{Architecture \\ details \textsuperscript{*}} & \thead{Dataset} & \thead{Validity \\ ($\uparrow$)} & \thead{Novelty \\ ($\uparrow$)} & \thead{Uniq. \\ ($\uparrow$)} & \thead{IntDiv \\ ($\uparrow$)} & \thead{QED \\ ($\uparrow$)} & \thead{Overall \\ ranks ($\uparrow$)} \\

    \hline
    \multirow{16}{*}{\thead{Non-\\targeted\\design}}     
    & REINVENT & Text & RNN + RL & ChEMBL & 0.940 & 0.307 & - & 0.755 & 0.525 & 10 \\
    & BIMODAL & Text & RNN & ChEMBL & \textbf{0.997} & 0.314 & - & 0.720 & 0.541 & 9 \\
    & MolGPT & Text & Transformer & MOSES & \textbf{0.996} & 0.782 & \textbf{1.000} & 0.857 & - & 2 \\
    & ORGAN & Text & GAN & ZINC & 0.379 & 0.687 & 0.841 & - & 0.520 & 12 \\
    & CMGN & Text & Transformer & \thead{ZINC + \\ ChEMBL} & 0.989 & 0.585 & 0.409 & 0.682 & - & 11 \\
    & EDM & 3D coord.* & Diffusion models & ChEMBL & 0.106 & \textbf{1.000} & 0.964 & \textbf{0.905} & 0.223 & 5 \\
    & QADD & Graph & GNN + RL & ChEMBL & \textbf{1.000} & 0.341 & - & 0.614 ± 0.029 & 0.785 ± 0.008 & 6 \\
    & MARS & Graph & GNN + MCMC & ChEMBL & \textbf{0.997} & 0.333 & - & 0.641 & 0.746 & 8 \\
    & MGM & Graph & MPNN & ChEMBL & 0.849 & 0.722 & \textbf{1.000} & - & 0.582 & 4 \\
    & molDQN & Graph & Q-Learning (RL) & \thead{ZINC + \\ ChEMBL} & \textbf{1.000} & 0.360 & - & 0.531 & 0.761 & 7 \\
    & STAGAN & Graph & GAN & ZINC-250k & 0.482 & \textbf{1.000} & \textbf{1.000} & - & - & 3 \\
    & \thead{DrugGEN-\\NoTarget\\(ours)} & Graph & \thead{Graph \\ transformer\\GAN} & ChEMBL & 0.913 & \textbf{0.990} & \textbf{1.000} & 0.874 ± 0.020 & 0.502 ± 0.214 & \textbf{1} \\
     \hline
    
    \multirow{8}{*}{\thead{Targeted\\design}}
    & RELATION & Text & BiTL & \thead{ChEMBL + \\ PDBBind} & 0.854 & \textbf{1.000} & \textbf{0.999} & 0.857 ± 0.016 & 0.558 ± 0.187 & 2 \\
    & \thead{TRIOMPHE-\\BOA} & Text & VAE + Attention & ExCAPE & - & - & 0.445 & 0.586 ± 0.156 & \textbf{0.819 ± 0.116} & 5 \\
    & ResGen & 3D coord. & GVP & CrossDocked2020 & - & - & 0.289 & 0.547 ± 0.092 & 0.218 ± 0.096 & 6 \\
    & TargetDiff & 3D coord. & SE(3) Diffusion & CrossDocked2020 & 0.839 & - & \textbf{1.000} & \textbf{0.879 ± 0.012} & 0.462 ± 0.190 & 3 \\
    & Pocket2Mol & 3D coord. & E3GNN + MCMC & CrossDocked2020 & 0.962 & - & 0.365 & 0.844 ± 0.022 & 0.400 ± 0.188 & 4 \\
    & \thead{DrugGEN\\(ours)} & Graph & \thead{Graph transformer\\GAN} & ChEMBL & 0.931 & \textbf{0.991} & \textbf{1.000} & 0.866 ± 0.018 & 0.507 ± 0.213 & \textbf{1} \\

    \hline
    
    \label{tab:table1}
    \vspace{-1em}
    \end{tabular}}
{\raggedright *RNN: Recurrent Neural Networks, RL: Reinforcement Learning, BiTL: Bidirectional Transfer Learning, GAN: Generative Adversarial Networks, GNN: Graph Neural Networks, MCMC: Markov chain Monte Carlo, MPNN: Message-passing Neural Networks, GVP: Geometric Vector Perceptron. coord.: coordinates\par}  
    \vspace{0.3em}
    \hrule

\end{table}

\subsection{DrugGEN generates target-specific drug-like molecules}

The DrugGEN model was trained and tested on two distinct targets, AKT1 and CDK2, to evaluate its performance across different proteins. The evaluation was mainly conducted for the AKT1 protein, with CDK2 included as an additional target to assess the reproducibility of our system. We selected CDK2 because it is frequently deregulated in various cancers, making it a critical target for de novo drug development \cite{tadesse2020targeting}. Additionally, there are CDK2 targeting models in the literature that can be compared with DrugGEN. \\

In this analysis, we evaluated targeted de novo design models, also presented in Table 1, namely RELATION, TRIOMPHE-BOA, ResGen, TargetDiff, Pocket2Mol and DrugGEN, using docking scores against AKT1 and CDK2, together with drug-likeness-related metrics, including quantitative estimate of drug-likeness (QED), synthetic accessibility (SA) \cite{Polykovskiy_Zhebrak_Sanchez_Lengeling_Golovanov_Tatanov_Belyaev_Kurbanov_Artamonov_Aladinskiy_Veselov_et_al._2020} Frechet ChemNet Distance (FCD), Lipinski, Veber, and PAINS. We also added fragment and scaffold similarities to compare de novo molecules with real inhibitors of the selected targets (see “Performance metrics” in Methods for details). Finally, we calculated these metrics for the training dataset (real) molecules to serve as a reference. Table 2 presents the results of this comparative analysis. Also, Figure 2A shows docking results with bar graphs (percentages shown on top of each bar represent docking performance of generative models compared to the real AKT1/CDK2 inhibitors, i.e., as the percentage of the docking performance of real inhibitors). \\

In AKT1-targeted de novo molecule generation, DrugGEN scored the highest docking performance with -8.386 kcal/mol and 99.98\% (Table 2 and Figure 2A), exhibiting a noteworthy contrast to the other models. Furthermore, the difference between the DrugGEN and the next best model, RELATION (i.e., -8.386 and -8.100 kcal/mol, respectively), was found to be statistically significant according to the Mann-Whitney U rank test (p-value < 0.0001). Following DrugGEN and RELATION, TargetDiff and Pocket2Mol exhibit satisfactory docking scores (91.59\% and -7.981 kcal/mol for TargetDiff and 91.10\% and -7.957 kcal/mol for Pocket2Mol). DrugGEN also achieved a well-balanced profile across QED (0.507), SA (3.510), and structural similarity metrics. \\

When DrugGEN is trained for targeting the CDK2 protein, it achieved a top 10\% median docking score of -8.747 kcal/mol (86.99\%), indicating superior binding affinity compared to TargetDiff while remaining competitive with Pocket2Mol (Table 2). Our model generated drug-like molecules with an average QED score of 0.547, reflecting its capability to design compounds that adhere to key drug-likeness criteria, further supported by its high performance on the Lipinski, Veber, and PAINS filters (Table 2). On the other hand, Pocket2Mol showed suboptimal drug-likeness characteristics, with lower QED and higher SA scores (Table 2), indicating challenging synthetic complexities. More details on docking and drug-likeness analysis can be found in Supplementary Material (S4). \\

Furthermore, we measured the docking binding free energy of the capivasertib-AKT1 complex (i.e., the bound ligand in our template PDB structure: 4GV1), as well as other experimentally proven inhibitors that reached clinical trials, namely ipatasertib and uprosertib, at -9.517, -9.681 and -8.569 kcal/mol, respectively. DrugGEN was able to generate 40 molecules that surpass the binding score of the native ligand capivasertib, which are shown in Figure S2. 

\renewcommand{\theadfont}{\fontsize{6}{11}}
\begin{table} [h]
\setlength\tabcolsep{2.5pt}
\caption{Drug-likeness related and target-centric performance of DrugGEN and other methods: RELATION\cite{Wang_Hsieh_Wang_Wang_Weng_Shen_Yao_Bing_Li_Cao_et_al._2022}, ResGen \cite{zhang2023resgen}, TRIOMPHE-BOA \cite{matsukiyo2023novo}, TargetDiff \cite{guan20233d}, and Pocket2Mol \cite{peng2022pocket2mol}, measured in terms of of QED, synthetic accessibility (SA), FCD, fragment similarity, scaffold similarity, and adherence to Lipinski, Veber, and PAINS filters, together with docking scores (median kcal/mol values of the top 10\% molecules in terms of docking scores), for the AKT1 and CDK2 targeting tasks, separately. Values for the training dataset (real molecules) are also provided. Arrows show the direction in which the performance increases. The best performances are displayed in bold font, separately for AKT1 and CDK2 targeting models (all within \%1 of the highest score are included due to very close numbers without a significant difference). Performance values that could not be calculated are indicated by “-”. Metrics that contain a distribution of values are reported as mean ± SD, while metrics yielding a single value are reported as such.}
    \centering
    \fontsize{6}{11}\selectfont
    \makebox[\linewidth]{
    \begin{tabular}{ccccccccccc}
    
    \hline
    \thead{Model objective /\\ data property} & \thead{Model/dataset\\ name} & \thead{Docking\\ score ($\downarrow$)} & \thead{QED\\ ($\uparrow$)} & \thead{SA\\ ($\downarrow$)} & \thead{FCD\\ ($\downarrow$)} & \thead{Fragment\\ Sim. ($\uparrow$)} & \thead{Scaffold\\ Sim. ($\uparrow$)} & \thead{Lipinski\\ ($\uparrow$)} & \thead{Veber\\ ($\uparrow$)} & \thead{PAINS\\ ($\uparrow$)} \\
    \hline
    \multirow{3}{*}{\thead{Training Dataset\\(real molecules)}}
    & Random ChEMBL mol.s* & - & 0.537 ± 0.223 & 3.008 ± 0.965 & - & - & - & 4.645 ± 0.801 & 1.827 ± 0.474 & 0.935 \\
    & Real AKT1 inhibitors & -8.387 ± 0.048 & 0.552 ± 0.189 & 3.225 ± 0.833 & - & - & - & 4.736 ± 0.507 & 1.951 ± 0.236 & 0.947 \\
    & Real CDK2 inhibitors & -9.352 ± 0.041 & 0.580 ± 0.148 & 2.911 ± 0.591 & - & - & - & 4.927 ± 0.289 & 1.954 ± 0.213 & 0.910 \\

    \hline
    \multirow{7}{*}{\thead{AKT1\\targeting}}
    & RELATION & -8.100 ± 0.038 & 0.558 ± 0.187 & 2.711 ± 0.546 & \textbf{20.186} & \textbf{0.770} & \textbf{0.267} & 4.854 ± 0.394 & \textbf{1.961 ± 0.198} & 0.948 \\
    & TRIOMPHE-BOA & -6.920 ± 0.061 & \textbf{0.819 ± 0.116} & \textbf{2.534 ± 0.497} & 69.638 & 0.201 & 0.000 & \textbf{4.997 ± 0.056} & \textbf{1.976 ± 0.153} & 0.525 \\
    & ResGen & -6.468 ± 0.101 & 0.218 ± 0.096 & 5.680 ± 1.208 & 42.100 & 0.185 & 0.000 & 3.012 ± 0.941 & 1.171 ± 0.377 & 0.491 \\
    & Pocket2Mol & -7.957 ± 0.033 & 0.400 ± 0.188 & 4.414 ± 1.134 & 26.413 & 0.638 & 0.000 & 4.225 ± 0.688 & 1.74 ± 0.441 & 0.915 \\
    & TargetDiff & -7.981 ± 0.025 & 0.462 ± 0.19 & 4.952 ± 0.857 & 30.663 & 0.598 & 0.003 & 4.673 ± 0.681 & 1.724 ± 0.498 & \textbf{0.964} \\
    & MLP* baseline & -6.367 ± 0.017 & 0.545 ± 0.187 & 3.117 ± 0.923 & 36.227 & 0.519 & 0.003 & 4.815 ± 0.496 & 1.895 ± 0.319 & \textbf{0.966} \\
    & DrugGEN (ours) & \textbf{-8.386 ± 0.018} & 0.507 ± 0.213 & 3.510 ± 0.809 & \textbf{20.416} & 0.721 & 0.246 & 4.599 ± 0.815 & 1.834 ± 0.437 & 0.932 \\
    \hline
    \multirow{3}{*}{\thead{CDK2\\targeting}}
    & Pocket2Mol & \textbf{-8.803 ± 0.030} & 0.387 ± 0.187 & 3.996 ± 0.902 & 30.119 & 0.484 & 0.000 & 4.421 ± 0.63 & 1.614 ± 0.487 & 0.907 \\
    & TargetDiff & -8.353 ± 0.024 & \textbf{0.556 ± 0.185} & 5.017 ± 1.008 & 26.483 & 0.766 & 0.000 & \textbf{4.833 ± 0.508} & \textbf{1.888 ± 0.334} & \textbf{0.967} \\
    & DrugGEN (ours) & \textbf{-8.747 ± 0.022} & \textbf{0.547± 0.180} & \textbf{3.409 ± 0.925} & \textbf{10.611} & \textbf{0.931} & \textbf{0.311} & \textbf{4.860 ± 0.442} & \textbf{1.906 ± 0.314} & 0.940 \\
    \hline    
    \vspace{-1.3em}
    \end{tabular}}
    
 {\raggedright *MLP: Multilayer perceptron, mol.s: molecules\par}  
 \vspace{0.3em}
    \hrule
    \label{tab:table2}
\end{table}

\begin{figure*}
    \vspace{-2.5em}
    \centering
    \includegraphics[width=13cm]{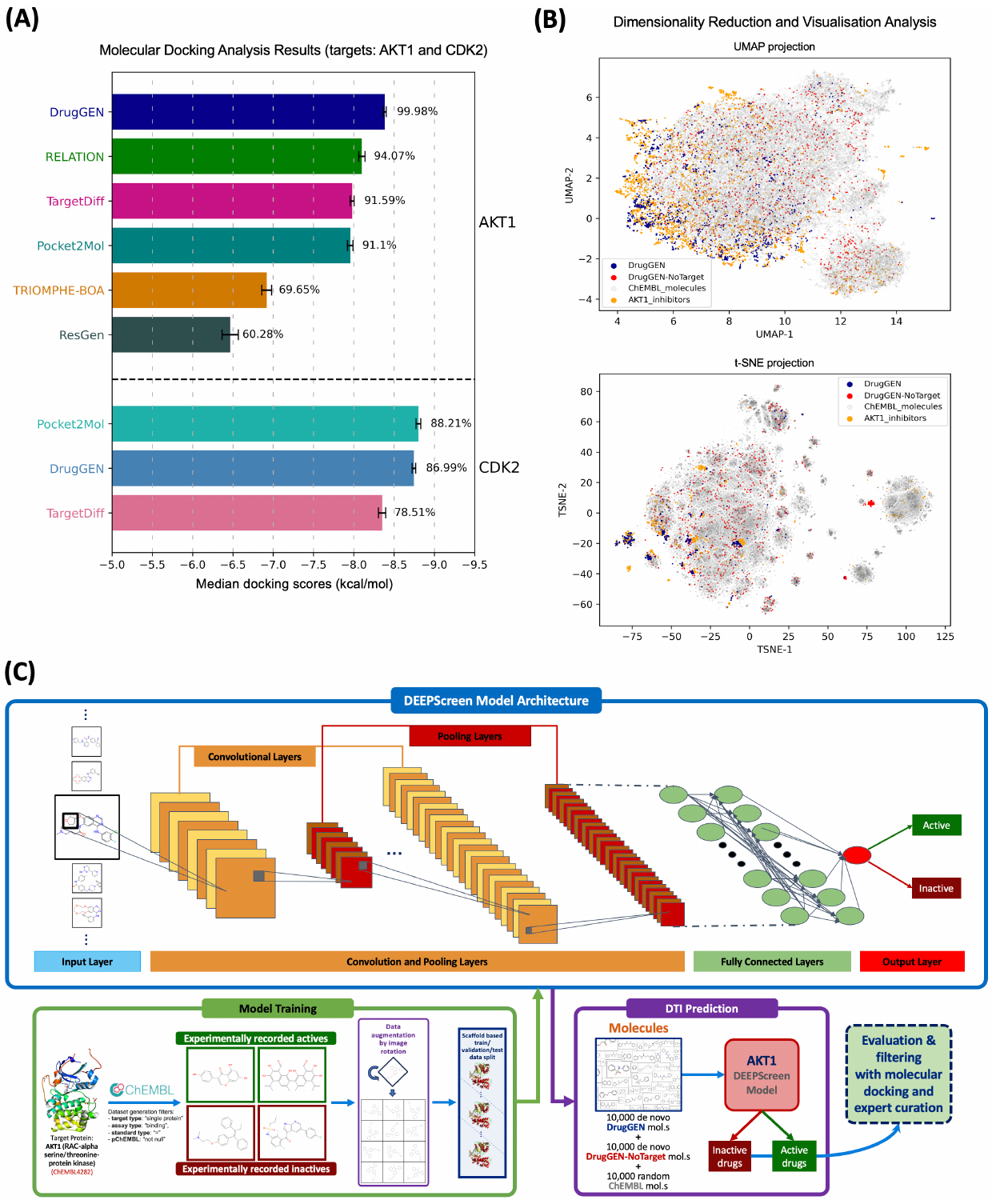}
    \vspace{-3em}
    \caption{Exploration of de novo molecules via downstream analysis. \textbf{(A)} Bar plots displaying the median of binding free energies measured in the docking analysis of de novo molecules generated by DrugGEN, RELATION, TargetDiff, Pocket2Mol, TRIOMPHE-BOA, and ResGen models (molecules are docked into the binding pocket of the AKT1/CDK2 protein structure). The whiskers on the bars represent the standard error of the median (median values are utilised due to the non-normal distribution of data). The percentages above each bar represent docking scores normalised with respect to the scores of real AKT1/CDK2 inhibitors, with the top real AKT1/CDK2 inhibitors set at 100\%. \textbf{(B)} 2-D visualisation of the molecules generated by DrugGEN and DrugGEN-NoTarget models, real AKT1 inhibitors and randomly selected ChEMBL molecules via UMAP and t-SNE projections. \textbf{(C)} Deep learning-based bioactivity prediction analysis. Top: the workflow of DEEPScreen, which uses 2-D pixel-based (image) structural representations of molecules as input and processes them via deep convolutional neural networks \cite{Rifaioglu_Nalbat_Atalay_Martin_Cetin-Atalay_Doğan_2020}. Bottom-left: the DEEPScreen AKT1 model is trained with binarised experimental bioactivity data points of the AKT1 protein (obtained from ChEMBL binding assays). Bottom-right: molecules (i.e., DrugGEN and DrugGEN-NoTarget molecules and randomly selected ChEMBL molecules) are submitted to DrugGEN AKT1 DTI prediction model for inference.
}
    \label{fig:img2}
\end{figure*}

\clearpage
Additionally, with the aim of assessing the generative performance of DrugGEN from the perspective of input data, we measured how the type of input molecules (i.e., random ChEMBL molecules vs. real AKT1 inhibitors) given to the model during inference changed the output. The results are provided in Supplementary Material (S5) with 3 example molecules in Figure S3.\\

Finally, we conducted physicochemical property comparisons between DrugGEN, other target-based generation models, and real molecule datasets (Supplementary Material S7 and Figure S5); and further assessed the drug-related properties of DrugGEN molecules by subjecting them to Lipinski, Veber and PAINS filters (Supplementary Material S8). \\

Overall, the results demonstrate that the generative and drug-likeness metrics of DrugGEN are consistent across different targets, indicating the versatility of the generation pipeline for diverse protein targets. Therefore, it is possible to state that DrugGEN was successful in its target-specific and drug-like molecular design routine. \\ 

\subsection{Ablation study}

In an extensive ablation study, we compared several variations of our model to assess the effectiveness of the graph transformer architecture and the modified attention mechanism in DrugGEN for target-specific molecule generation. Specifically, we evaluated the default DrugGEN model and its non-targeted variant (DrugGEN-NoTarget) against baseline models that use simpler architectures—multi-layer perceptron (MLP) or graph convolutional network (GCN) modules—in both the generator and discriminator modules. Additionally, we analysed the attention mechanism to investigate various strategies for incorporating edge information (i.e., atomic bonds). For each model variant, 10,000 de novo molecules were generated and evaluated using the same fundamental generative metrics and drug-likeness scores as in Tables 1 and 2. Overall, the ablation study confirms that the use of graph transformer blocks with the modified attention, involving an additional multiplication by a shifted adjacency matrix, i.e., $Att \bigodot A_m \bigodot (A_m+1)$, significantly enhances the model’s ability to generate valid, diverse, and novel molecules with high drug likeness, demonstrating superior performance compared to the MLP- and GCN-based baselines and attention variants. Please see Supplementary Material (S6) and Table S1 for detailed descriptions and complete results. \\

\subsection{Exploration of Molecules via Dimensionality Reduction}

We visualized both real and de novo molecules from the previous analyses using UMAP \cite{McInnes_Healy_Melville_2018} and t-SNE \cite{Van_der_Maaten_Hinton_2008} algorithms. algorithms (please see Methods for details). In the UMAP and t-SNE plots in Figure 2B, molecules generated by the DrugGEN model are clustered together with real AKT1 inhibitors, indicating a higher structural resemblance (regarding topological fingerprint-based similarities). On the other hand, molecules generated by DrugGEN-NoTarget exhibit a nearly homogeneous distribution across the entire plane due to the non-targeted generation approach. Our conclusion is that the DrugGEN model has learned the structural distribution of the AKT1 inhibitor molecules. \\

\subsection{Deep Learning-based Bioactivity Prediction}

To evaluate the target interaction-related properties of DrugGEN’s de novo molecules in parallel to molecular docking, we carried out a deep learning-based drug/compound–target interaction (DTI) prediction analysis against the AKT1 protein, using our previously developed DEEPScreen system \cite{Rifaioglu_Nalbat_Atalay_Martin_Cetin-Atalay_Doğan_2020}. DEEPScreen is a supervised discriminative deep learning system composed of target-centric bioactivity prediction models that classify input molecules as either active (interacting) or inactive (non-interacting) against the selected protein (Figure 2C). For this analysis, we trained and tested a DEEPScreen model using the experimentally proven active and inactive small molecules of AKT1  (please see “Methods” for details). \\

When tested on the scaffold-split hold-out dataset composed of 484 known AKT1 binder molecules, DEEPScreen AKT1 model accurately predicted 475 as active molecules based on a standard confidence score threshold of 0.5 (i.e., at least half of the augmented molecule images are predicted as active), resulting in precision: 0.96, recall: 0.94, F1-score: 0.95, and MCC: 0.91, which was considered highly satisfactory. Among 484 known AKT1 binders, 421 were classified with a high confidence score ( 0.85), indicating the reliability of the model. \\

We tested 10,000 de novo molecules generated by the DrugGEN model using DEEPScreen, which resulted in 1,382 active predictions based on the standard confidence threshold ($\geq$0.5). Of these, 748 molecules were predicted with high confidence ($\geq$ 0.85). A complete histogram of the confidence scores for active predictions is provided in Figure S6. DrugGEN molecules, on average, exhibit a lower average molecular similarity (Tanimoto) to the training set of DEEPScreen when compared with the hold-out test dataset (0.309 and 0.637, respectively), which may explain why the proportion of de novo molecules predicted to be positive is not as high. We also analysed 10,000 random ChEMBL compounds and 10,000 DrugGEN-NoTarget molecules with DEEPScreen for comparison, which received 122 and 69 high-confidence ($\geq$ 0.85) active predictions against AKT1, respectively. Considering the two non-targeted molecule groups (i.e., DrugGEN-NoTarget and random ChEMBL molecules), the difference in the number of high-confidence active predictions (748 vs. 122 and 69) indicates the effectiveness of DrugGEN in generating target-based molecules. \\

\subsection{Selection of thirty promising molecules}

We manually picked 30 promising de novo molecules (mainly from molecules with docking scores < -8 kcal/mol and DEEPScreen predictions as “active”) via expert curation and presented them as our best candidates to target AKT1 (Figure 3). Many of these selected molecules can be considered readily synthesisable and exhibit favorable drug-like properties as indicated by their SA scores (< 6 \cite{ertl2009estimation}), and QED scores (> 0.5 \cite{bickerton2012quantifying}), while the others display less conventional features, such as increased polarity or molecular weight. Compliance with Lipinski’s Rule of Five was also confirmed, with hydrogen-bond donors ($\leq$ 5) and acceptors ($\geq$ 10) within acceptable limits, further supporting their potential as drug candidates. \\

Figure S7 shows a subset of those 30 de novo molecules alongside the real AKT1 inhibitory molecules (from the training dataset) most similar to them (two real molecules for each de novo design). As observed from Figure S7, de novo molecules are structurally diverse even from their most similar counterpart in the training set. Yet, most of these molecules obtained convincing docking scores (i.e., < -8 kcal/mol) against AKT1. \\

\begin{figure*}
    \centering
    \includegraphics[width=\textwidth,clip=True]{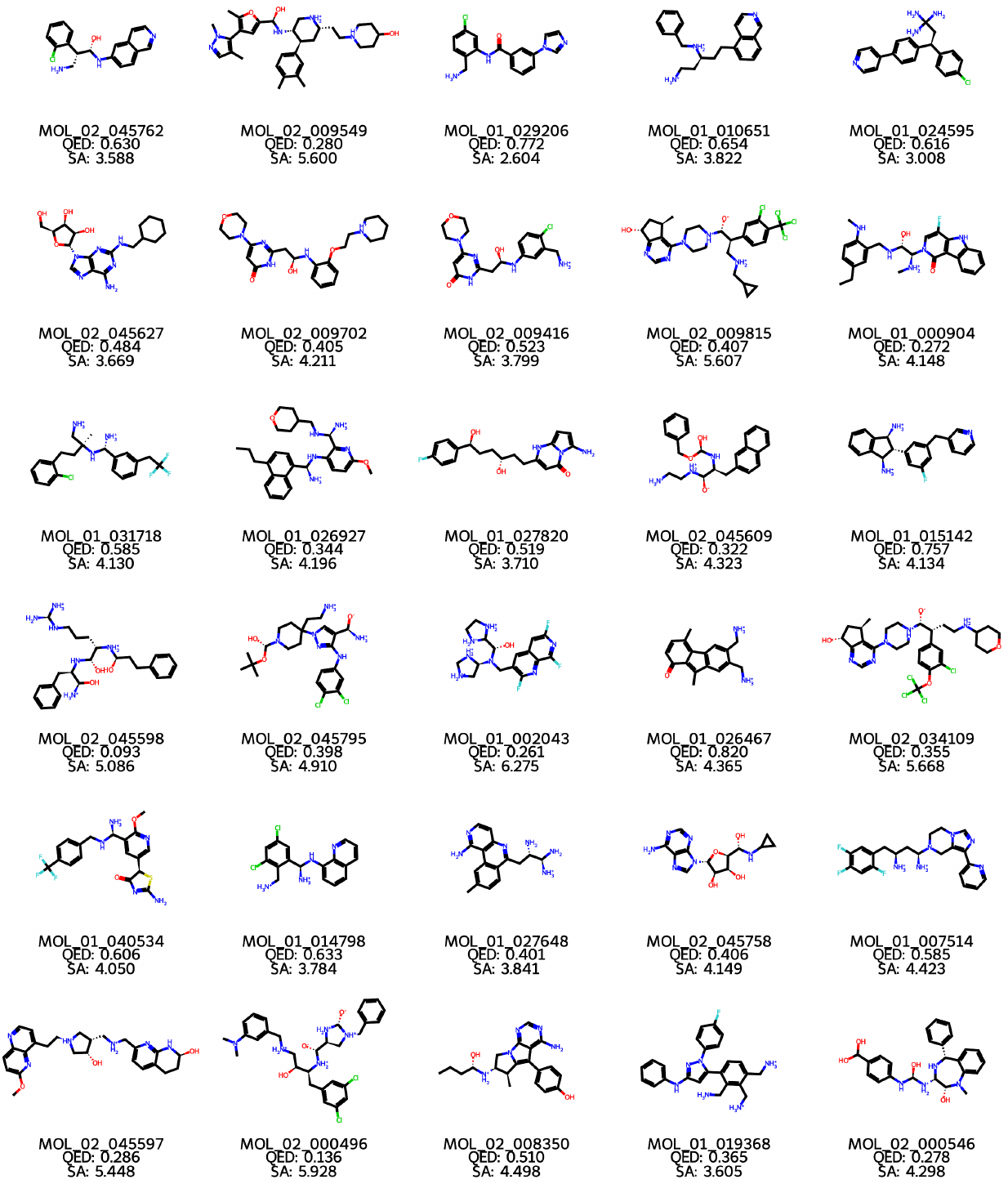}
    \vspace{-3em}
    \caption{30 promising de novo molecules to effectively target AKT1 protein (generated by DrugGEN model), selected via expert curation from the dataset of molecules with sufficiently low binding free energies (< -8 kcal/mol) in the molecular docking experiment and deep learning-based DTI predictions  \cite{Rifaioglu_Nalbat_Atalay_Martin_Cetin-Atalay_Doğan_2020}) as “active”.
}
    \label{fig:img3}
\end{figure*}

\subsection{Experimental validation of de novo molecules for AKT1 inhibition}

Following the computational analyses, chemical synthesis and in vitro enzymatic assays were conducted for experimental validation. Among the 30 promising de novo compounds given in Figure 3, five structurally diverse compounds were selected to be synthesised (Figure 4A). These compounds were chosen based on their economic synthetic accessibility and low synthesis and purification turnaround times. Figure 4B shows the best docking poses of Molecules 1 ($MOL\_02\_045762$), 2 ($MOL\_02\_045795$), 3 ($MOL\_02\_045758$), 4 ($MOL\_02\_045609$), and 5 ($MOL\_02\_045627$) with binding free energies of -8.948, -7.686, -8.655, -8.383, and -5.341 kcal/mol, respectively. Additionally, Figures 4A and 4B include the reference crystal structure ligand, capivasertib, for comparison, with a docking score of -9.517 kcal/mol. Details about the docking results of synthesised molecules are provided in Supplementary Material (S9). \\

Selected compounds were synthesised by commercial organic chemistry laboratories. The analytical proof for the purities of the compounds can be found in Figures S12-16. Subsequently, in vitro enzymatic assays were conducted on these compounds, which rely on transferring a radioactive \textsuperscript{33}P-labeled phosphate group from ATP to the target kinase (AKT1) substrate. The study yielded two out of five chemically different compounds, showing IC\textsubscript{50} values of 1.89 µM (Molecule 1) and 48.6 µM (Molecule 2) against AKT1 (Table S3). However, Molecules 3, 4, and 5 exhibited insufficient activity up to 100 µM. Corresponding dose–response curves are displayed in Figure S17. To rationalise the experimental results, molecular dynamics simulations were performed and explained below. \\

\begin{figure*}

    \centering
    \includegraphics[width=\textwidth,clip=True]{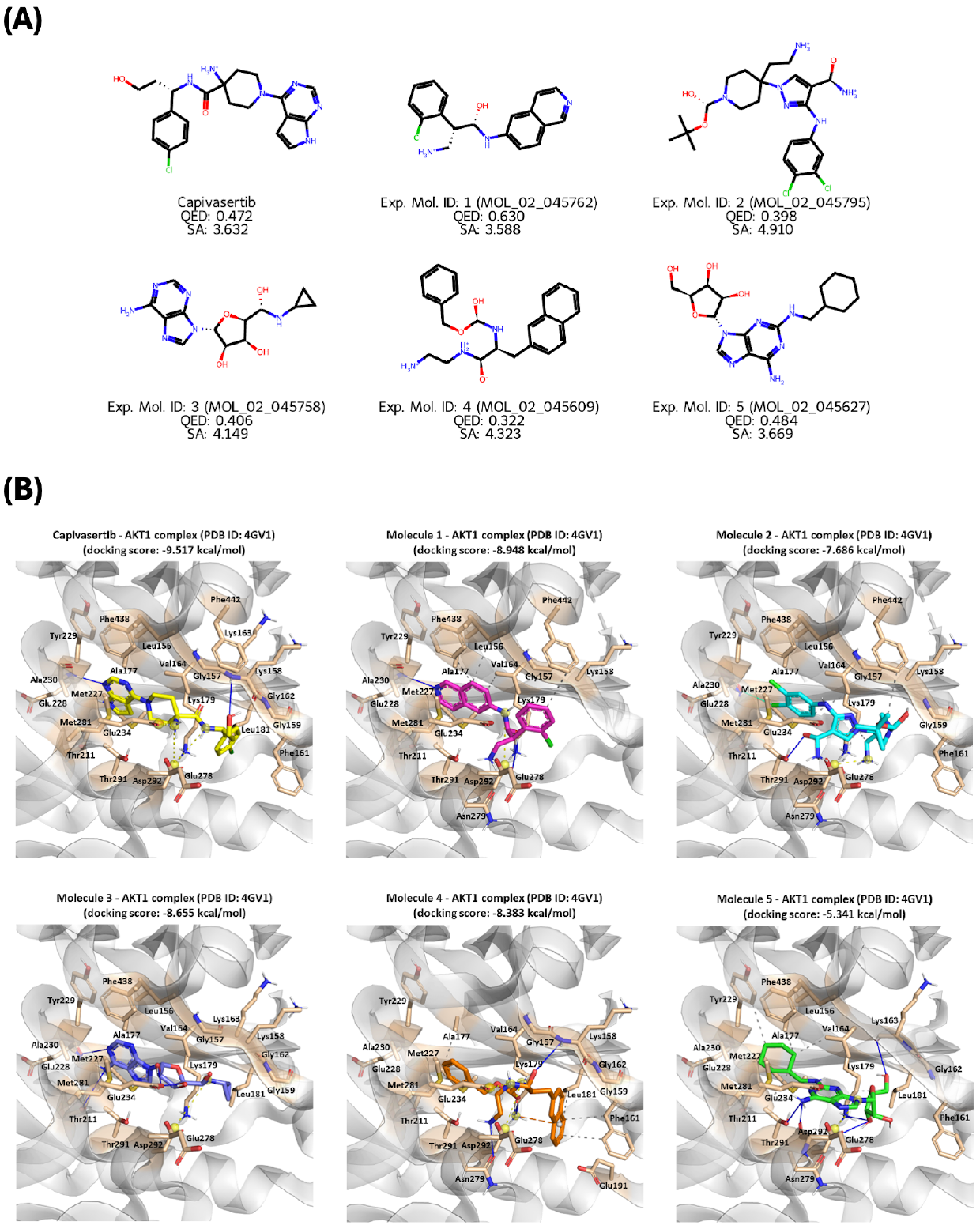}
\end{figure*}

\begin{figure*}
\vspace{-9.5em}
    \centering
    \includegraphics[width=\textwidth,clip=True]{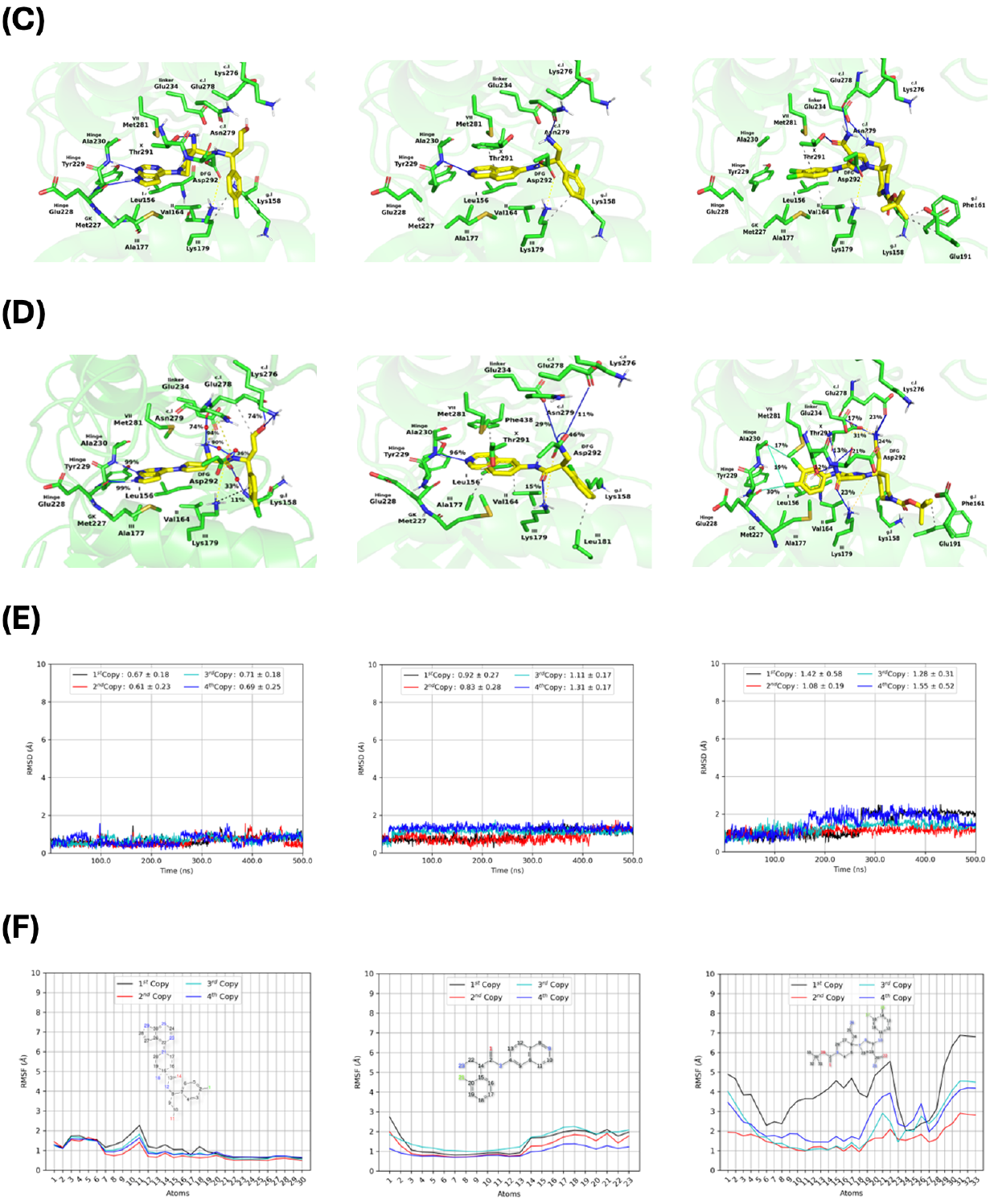}

\end{figure*} 

\clearpage
\captionof{figure}{Structural analysis of capivasertib (bound ligand in 4GV1) and five de novo generated molecules (Mol. ID 1–5) that were selected for experimental validation. \textbf{(A)} 2D visualisation of capivasertib and Molecules 1–5, with their unique identifiers assigned during molecule generation, QED and SA values. \textbf{(B)} 3D protein-ligand interactions of capivasertib and Molecules 1–5 after molecular docking with AKT1 (PDB ID: 4GV1). The interactions are visualised with active site residues of AKT1 highlighted in beige color. Different interaction types are depicted as follows: blue lines represent hydrogen bonds, yellow dashed lines indicate salt bridges, gray dashed lines denote hydrophobic interactions, orange dashed lines highlight $\pi$-cation interactions, and green lines represent halogen bonds. \textbf{(C)} Initial binding orientations of capivasertib, Molecules 1 and 2 at the starting point of molecular dynamics (MD) simulations. \textbf{(D)} Key protein-ligand interactions observed during MD simulations, visualised with interacting residues and interaction types. The depicted poses represent the most populated conformations from each simulation. \textbf{(E)} Root-mean-square deviation (RMSD) values of capivasertib, Molecules 1 and 2 in complex with AKT1. \textbf{(F)} Root-mean-square fluctuation (RMSF) values of ligand atoms in the same complexes. Abbreviations: Exp. Mol. ID: Experimental molecule identifier. I-VII represents $\beta$-sheet numbers, g.l represents glycine-rich loop, c.l represents catalytic loop, GK represents gatekeeper residue, and xDFG represents highly conserved kinase residues; linker represents the loop that connects the hinge domain to $\alpha$C-helix. Gray dashed lines represent Van der Waals interactions. Blue lines represent hydrogen bonds and water bridges. Green lines indicate halogen bonds. Yellow dashed lines represent salt bridges. Directional interactions were noted only if the occupancy value was found above 10\%; however, for visual clarity, occupancy values of the water bridges were stated only if they were above 30\%.
}

\vspace{1em}
    
\subsection{Investigation of the binding properties of molecules using molecular dynamics}

Here, we performed molecular dynamics (MD) simulations and evaluated the binding characteristics of five synthesised de novo molecules with AKT1. The simulations were performed on the crystal’s co-crystalized ligand, capivasertib, as well as the docked poses of Molecules 1-5. The starting orientations of all these compounds were visualised within the ATP cofactor binding site in Figures 4C and S8. \\ 

MD simulations showed that all compounds remained stable (RMSD < 1.6 Å) (Figures 4E and S8), with no exceptional movements (i.e., low RMSF) in most cases (Figures 4F and S8). Capivasertib maintained near-constant hydrogen bonds with key residues (Tyr229 and Ala230) and additional water-mediated interactions with Lys158, Asn279, and Asp292 (Figure 4D). Molecule 1, which exhibited the most potent inhibition (IC\textsubscript{50} = 1.89 µM), formed strong hydrogen bonds with hinge residue Ala230 (96\%) and interacted with surrounding acidic residues, suggesting its activity derives from a similar interaction network as capivasertib—though its lack of additional hinge contacts and robust hydrogen bond networks may slightly limit its efficacy (Figure 4D). In contrast, Molecule 2 formed halogen bonds with hinge residues Tyr229 and Ala230 and hydrogen bonds with other critical residues (Lys179, Glu234, Glu278, Thr291, Asp292), accounting for its second most potent inhibitory effect (Figure 4D). Details about the MD results of synthesised molecules are provided in Supplementary Material (S10). \\

\subsection{Attention map visualisation}

We visualised the attention scores of DrugGEN’s generator transformer-encoder as an explainability analysis to evaluate critical atomic interactions in de novo molecules. Specifically, we assessed the correspondence between ligand atoms with high attention scores and those identified as critical in molecular docking (i.e., atoms directly interacting with AKT1 residues). As shown in Figure 5, DrugGEN assigned high attention scores to key atoms involved in hydrogen bonds and salt bridges—accurately identifying ~91\% of the interacting ligand atoms—demonstrating that, despite not being trained on explicit pairwise interaction data, the generator network effectively learned substructures essential for AKT1 binding, underscoring DrugGEN’s potential for target-centric molecule design. Additional information can be found in Supplementary Material (S13).

\begin{figure*}
\vspace{-3.5em}
    \centering
    \includegraphics[width=9cm,clip=True]{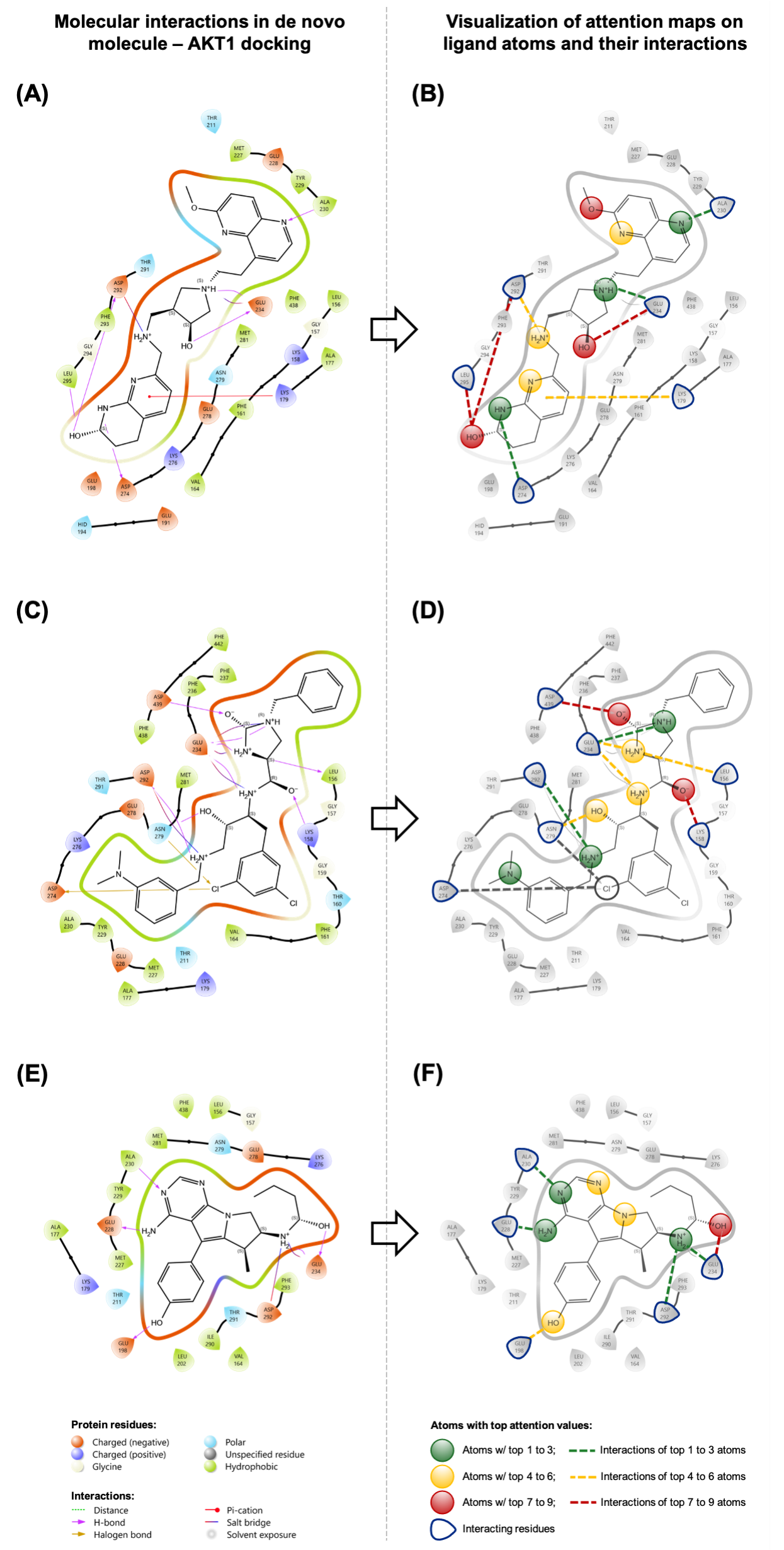}
    \caption{Visualisation of DrugGEN attention maps on three de novo molecules. The \textbf{left}-side column depicts protein-ligand interaction diagrams obtained from the molecular docking of DrugGEN’s three de novo molecules (in A, C and E) with the AKT1 protein structure (PDB id: “4GV1”) \cite{Addie_Ballard_Buttar_Crafter_Currie_Davies_Debreczeni_Dry_Dudley_Greenwood_et_al._2013}. The docked ligands are located in the binding pocket, with interactions between residues and the ligand shown as lines, coloured by the interaction type. Protein residues are coloured according to their physicochemical properties. The \textbf{right}-side column depicts the attention maps of the same three de novo molecules (in B, D and F) retrieved from the graph transformer module of the DrugGEN generator network. Atoms that receive the highest attention scores are highlighted with colours (green: 1st, 2nd and 3rd atoms; yellow: 4th, 5th and 6th atoms; and red: 7th, 8th and 9th atoms with the highest attention scores). Atoms with lower attention scores are not coloured. In the right-side plots, receptor-ligand interactions (i.e., those obtained from the docking analysis) are represented by dashed lines in the attention-score-based colour of the molecule atom involved in the respective interaction. If an interacting atom could not be retrieved (i.e., the atom received a low attention score), its interaction is given in grey colour. \textbf{(A, B)} molecule id: $MOL\_02\_045597$, docking score: -9.803 kcal/mol, \textbf{(C, D)} molecule id: $MOL\_02\_000496$, docking score: -9.693 kcal/mol, \textbf{(E, F)} molecule id: $MOL\_02\_008350$, docking score: -9.619 kcal/mol.
}
    \label{fig:img5}
\end{figure*}
\clearpage
\section{Discussion and Conclusion}

In this study, we developed the DrugGEN system to generate target-centric drug-like molecules automatically. DrugGEN combines the GAN architecture with graph transformers to create an end-to-end system that can design novel ligands given the target protein. We further advanced the graph transformer architecture by incorporating a modified attention mechanism that more effectively integrates edge information, thereby reinforcing the representation of molecular structures. DrugGEN is trained and evaluated to generate new ligands of the AKT1 and CDK2 protein. DrugGEN displayed a better performance compared to SOTA models in fundamental performance metrics, pointing to its high generation efficiency and capacity. Considering physicochemical metrics and structural/topological features, we showed that the DrugGEN model can generate de novo molecules with similar molecular characteristics to real inhibitors of the AKT1 protein. Further in silico analyses (i.e., molecular docking and MD simulations) and in vitro enzymatic assays were conducted to assess the interaction-related characteristics and potency of the de novo molecules. The results of enzymatic assays indicated two out of five compounds have high potency at low micromolar concentrations, 1.89 µM (Molecule 1) and 48.6 µM (Molecule 2). Intending to present a tool to the community, we openly shared the code base, datasets, all results and trained models in our code repository at https://github.com/HUBioDataLab/DrugGEN and as a web-based tool with a graphical interface at https://huggingface.co/spaces/HUBioDataLab/DrugGEN where users can easily generate de novo molecules via employing the desired DrugGEN model. \\

One essential point about DrugGEN is that, unlike many previous generative models, DrugGEN operates at large molecular sizes similar to actual drugs and drug-like bioactive compounds (Figure S11) at the mean and maximum number of 33 and 45 heavy atoms, respectively, so that its output can be useful under different drug development-related contexts even though this choice elevated the complexity of the modelling problem. \\

A current and essential question in the literature is what steps should be taken after obtaining the de novo molecule records from a generative model so that they could ultimately be utilised in real-world applications of drug discovery, biotechnology or material science. In this work, we explored this topic and introduced a downstream analysis pipeline—incorporating both in silico and in vitro analyses (Figure 1C)—to facilitate the early selection of promising molecules. Our hierarchical selection pipeline can be applied to the large-scale outputs of any generative model, enabling the identification of the most promising drug candidate molecules and forwarding them for subsequent (pre)clinical development steps. \\

There are studies in the literature that aim to solve the target-specific and drug-like molecule generation problem via multi-objective optimisation. Several of these studies perform this task based on predictions from external models \cite{abeer2024multi, jain2023multi, monteiro2023fsm, suzuki2024mothra}. For example, FSM-DDTR leverages the predicted bioactivity (pIC\textsubscript{50}) values of molecules to construct a predictive model, which is then used as a reward function in the molecular generation module \cite{monteiro2023fsm}. However, performing optimisation based on predicted values carries the risk of error propagation. Our model simplifies the process by removing intermediate prediction-centric steps and directly mapping random molecule distributions to those of the selected protein's inhibitors, possibly reducing the chance of error propagation. Additionally, since our model does not rely on multiple explicit optimisation steps, the complexity of training and inference processes is potentially lower. This, in turn, facilitates the practical development and implementation of our models. In future studies, these aspects can be assessed. \\

Considering limitations, first of all, the adversarial training principle of GANs presented challenges, including significant divergence between the generator and discriminator losses, mode collapse, and the vanishing gradient  \cite{Ghosh_Dutta_Totaro_Bayoumi_2020,Gui_Sun_Wen_Tao_Ye_2023,Janson_Valdes_Garcia_Heo_Feig_2023}. However, adopting the Wasserstein GAN with gradient penalty (WGAN-GP)  \cite{Arjovsky_Chintala_Bottou_2017} has partially mitigated these issues. Second, as previously stated, this study uses random real molecular graphs, rather than noise, as input to the generator network. Given that large molecular graphs (composed of up to 45 heavy atoms) significantly increase the complexity of the generative task, it is exceedingly difficult for the model to establish a robust mapping between pure noise and the high-dimensional latent space of drug-like molecules. By substituting real molecular graphs for noise, we help the model more effectively learn a mapping from these initial structures to the de novo inhibitor molecules of the selected protein. Notably, this design does not impose explicit constraints on molecular novelty. Instead, the WGAN-GP framework encourages the generation of chemically valid yet novel molecules by penalising unrealistic structures rather than enforcing similarity to the training set. As demonstrated in Table 1, the model effectively generates novel compounds while maintaining chemical validity and target specificity. \\ 

In further studies, we plan to design and train: (i) DrugGEN models for other target proteins in the human proteome; (ii) GAN models that utilise sequence-based transformers using a valid and robust molecular notation such as SELFIES \cite{Krenn_Häse_Nigam_Friederich_Aspuru-Guzik_2020}together with a successful molecular representation learning model, e.g., SELFormer \cite{Yüksel_Ulusoy_Ünlü_Doğan_2023}; (iii) models that utilise fragments/ sub-structures of molecules as its building blocks, instead of individual atoms and bonds, to be able to generate highly synthesisable de novo compounds. Another interesting direction could be incorporating constraint optimisation techniques into DrugGEN's architecture to better control the generation of molecules with desired properties. Such constraints could be integrated into the GAN's loss function to guide the generation process toward molecules that meet specific physicochemical criteria, target binding requirements, or synthesizability thresholds. This enhancement would allow for more precise control over the generated molecules' properties while maintaining DrugGEN's current advantages in handling large molecular structures. \\ 

Additionally, we plan to incorporate target protein features into the molecule generation process directly—also adopted in conventional structure-based drug design—to yield more successful learning regarding molecular structural constraints. Finally, we plan to improve the molecular generation process by incorporating the high-level functional properties of real drugs and drug candidate molecules into the model training procedure in the context of heterogeneous biomedical knowledge graphs \cite{Doğan_Atas_Joshi_Atakan_Rifaioglu_Nalbat_Nightingale_Saidi_Volynkin_Zellner_2021}. This architecture is intended to facilitate understanding the relationships between the structural and functional properties of small molecules and thereby enhance the design process. 

\vspace{1.5em}

\section{Methods}

\subsection{Datasets}

Two types of data (i.e., compounds and target-based bioactivities) were retrieved from different data sources to train our deep generative models. The compound data, which includes atomic, physicochemical, and structural properties of drug and drug candidate molecules, represent our “real” samples. The compound dataset was retrieved from ChEMBL 
\cite{Mendez_Gaulton_Bento_Chambers_De_Veij_Félix_Magariños_Mosquera_Mutowo_Nowotka_2019} (v29), an open-access chemistry database containing curated high-quality data regarding drug-like small molecules and their experimentally measured activities on biological targets. The heavy atom distribution histogram of the ChEMBL dataset is given in Figure S10, which is used to determine the threshold for the maximum number of heavy atoms in a molecule to be utilised in our model. Based on the median value and standard deviation (std) of this distribution, we selected 45 heavy atoms as our threshold (i.e., mean + 2 std) and finalised our dataset with 1,588,865 small molecules. \\

Experimental bioactivities, i.e., quantitative measurements of physical interactions between drug-like compounds and their target proteins, are the second data type used in the training of DrugGEN. The bioactivity data was also retrieved from the ChEMBL database. We applied various filters for standardisation, such as target type: “single protein”, assay type: “binding assay”, standard type: “=”, and pChEMBL value: “not null” (i.e., roughly correspond to curated activity data points). Then, bioactivity data belonging to the AKT1 target protein were selected from the filtered bioactivity dataset. The finalised dataset contains ligand interactions of the human AKT1 (“CHEMBL4282”) protein with a pChEMBL value equal to or greater than 6 (i.e., IC50 <= 1 µM) as well as SMILES notations of these ligands. This activity dataset was extended by adding the drug molecules sourced from the DrugBank database \cite{Wishart_Feunang_Guo_Lo_Marcu_Grant_Sajed_Johnson_Li_Sayeeda_et_al._2018} that are known to interact with human AKT1 protein. The filtering process excluded molecules larger than 45 heavy atoms and resulted in a dataset of 2,405 active/interacting small molecule ligands to train generative models tailored to AKT. Their heavy atom size distribution is given in Figure S10. The same procedure was applied for the human CDK2 (CHEMBL301) protein, resulting in 1,817 compounds. \\

\subsection{Graph Encodings (Featurization)}

In DrugGEN, input molecules are represented as graphs, each composed of two parts: an annotation matrix (which contains information about the atom features) and an adjacency matrix (which includes information on the presence of atomic bonds and their features). The annotation and adjacency matrices of the compounds were generated using the RDKit \cite{Landrum_2013} 67 library and the SMILES representation of molecules. The annotation matrix has a size of 45 by 9, representing the maximum length (the number of heavy atoms) of the molecule and nine types of atoms (i.e., C, O, N, F, S, P, Cl, I, and the null / no atoms case), respectively. These one-hot-encoded atom types are used as node features of the molecule graphs. The adjacency matrix is 45 by 45 by 5, displaying the bond information of the respective atoms of the molecule (cells of the one hot encoded third dimension of the tensor; 0th: no bond between the corresponding atoms, 1st: single, 2nd: double, 3rd: triple, and 4th: aromatic bond). For CDK2-targeting molecules, the annotation matrix has dimensions 38 by 10, with 38 denoting the maximum number of heavy atoms and 10 corresponding to the atom types (C, N, O, F, P, S, Cl, Br, I, and a null token). The adjacency matrix for these molecules is 38 by 38 by 5, with the third dimension similarly representing bond types as described for the AKT1 case.  \\

\subsection{The Architecture of DrugGEN}

The DrugGEN model is built on Generative Adversarial Networks (GAN) \cite{Goodfellow_Pouget-Abadie_Mirza_Xu_Warde-Farley_Ozair_Courville_Bengio_2020}. Figure 1D shows the overall workflow of the default DrugGEN model. The generator $G$, a graph transformer encoder, transforms given input z to new annotation and adjacency matrices. These matrices are then fed to the discriminator network $D$ together with the matrices of real small molecules to assign them to the “real” and “fake” groups. The details of each module are provided below.

\paragraph{Generator:} The generator ($G$) module (Figure 1D) employs transformer encoder blocks \cite{Vaswani_Shazeer_Parmar_Uszkoreit_Jones_Gomez_Kaiser_Polosukhin_2017} which operate on graph-based data \cite{Dwivedi_Bresson_2021}, a method that has been shown to be highly effective for molecular data analysis \cite{li2021effective, rong2020self, li2022kpgt, Dwivedi_Bresson_2021}. For this, both the annotation and adjacency matrices are required to be processed in the same module (details regarding the dimensions and the context of annotation and adjacency matrices are given above, in the section entitled “Graph Encodings - Featurization”). $G$ takes randomly selected real drug-like molecules from the training dataset as input instead of random noise which is usually employed as the input of the generator modules of GANs. This approach aligns with studies in the literature that use real samples as input for GAN generators instead of random noise \cite{zhu2017unpaired, kim2017learning} This approach helps the model effectively handle the high complexity and sparsity of molecular graphs (i.e., 45*9 + 45*45*5 = 18,630 elements to represent a molecule). The input is fed through individual multi-layered perceptrons (MLPs) for annotation and adjacency matrices, both of which consists of four layers (i.e., input: 16, hidden1: 64, hidden2: 64, and output: 128 dimensions). These MLPs are utilized to create embedings of annotation and adjacency matrices with $d_k$ (default: 128) dimensions. Afterwards, the input is fed to the transformer encoder module. In the classic transformer architecture \cite{Vaswani_Shazeer_Parmar_Uszkoreit_Jones_Gomez_Kaiser_Polosukhin_2017}, self-attention is computed by deriving query ($QKV$, applying a softmax, and finally multiplying by $V$ to form the final output. Graph transformers extend this mechanism to graph-structured data by incorporating the molecule’s adjacency matrix $A_m$ to account for the connectivity between nodes (atoms). In this setting, $Q_m$, Km and $V_m$ are derived from the annotation matrix of the molecule.  Unlike the original transformer, the attention weights here are obtained by multiplying the scaled dot product of $Q_m$ and $K_m$ with the $A_m$.  The resulting attention is then multiplied by $V_m$ to create the updated representation of the annotation matrix, while the updated representation of the adjacency matrix is formed through the concatenation of these attention weights as described in the study by Dviwedi et al. (2020) and Vignac et al. (2022) \cite{Dwivedi_Bresson_2021, Vignac_Krawczuk_Siraudin_Wang_Cevher_Frossard_2022}. The purpose of performing element-wise multiplication of the attention with the adjacency matrix is to enhance the attention scores by incorporating bond information into them. The attention matrix aims to identify how much each atom attends to other atoms in a pairwise manner, and this is based on both short-range (direct) and long-range (indirect) relationships/interactions between the atoms in the molecule. Therefore, attention values can be enhanced by introducing readily available bond information between the corresponding atoms (i.e., whether an atomic bond exists or not, and if so, what specific kind of bond). In our model, we begin with the graph transformer’s standard attention formulation. Our key modification is an additional multiplication step involving a shifted adjacency matrix ($A_m+1$). Specifically, we first multiply the attention by $A_m$ and then multiply by ($A_m+1$) to amplify the edge signal, thereby enhancing the model’s ability to capture subtle structural features. The calculation of the final attention is formulated below:

\begin{equation}
    Attention(Q_{m}, K_{m}, V_{m}) = softmax(\frac{Q_{m}K_{m}^{T}}{\sqrt{d_{k}}}A_{m}(A_{m}+1))V_{m}
\end{equation} \\

Where $Q_{m}$, $K_{m}$, and $V_{m}$ denote the annotation matrix of the molecules while $A_{m}$ denotes their adjacency matrix. $d_k$ is the dimension of the transformer encoder module and it is used to scale the attention weights. The reason behind multiplying the attention with the adjacency matrix via element wise multiplication is to improve the attention scores by injecting bond information into it. The attention matrix aims to identify how much each atom attends to other atoms in a pairwise manner and this is based on both short range (direct) and long range (indirect) relationships/interactions between the atoms in the molecule. Therefore, attention values can be enhanced by introducing readily available bond information between the corresponding atoms (i.e., whether an atomic bond exists or not, and if so, what specific kind of bond). This idea was first proposed in Dwivedi and Bresson (2020) \cite{Dwivedi_Bresson_2021}. After the attention layer, the final annotation  and adjacency matrices undergo a process of addition and normalization. This transformation involves two main steps. First, they are passed through layer normalization, which ensures that the values within each matrix are scaled and centered appropriately. Layer normalization is formulated as follows: \\

\begin{equation}
    \hat{X}^{l+1} = LayerNorm({X}^{l} + \overline{X}^{l+1}, 
    \hat{A}^{l+1} = LayerNorm({A}^{l} + \overline{A}^{l+1})
\end{equation} \\

X and A correspond to annotation and adjacency matrices, respectively, LayerNorm represents the layer normalisation, and $l$ is the layer number. $\hat{X}^{l+1}$represents the intermediate product after the layer normalisation. ${X}^{l}$  is the annotation matrix before attention and $\overline{X}+{l+1}$ is the product of the attention mechanism. The same annotations are used for A (adjacency matrix). \\

Following this, the matrices are further processed via a feed-forward network (FFN). The FFN introduces non-linearity and learns complex representations from the input matrices, allowing higher-level feature extraction. The output of this network is then added back to the previously normalized matrices, creating a residual connection. Final matrices are passed through layer normalization at the end. The second operation can be expressed as: \\

\begin{equation}
    {X}^{l+1} = LayerNorm(\hat{X}^{l+1} + FFN(\hat{X}^{l+1}), 
    {A}^{l+1} = LayerNorm(\hat{A}^{l+1} + FFN(\hat{A}^{l+1})
\end{equation} \\

\paragraph{Discriminator:} The purpose of the discriminator (D) in GANs is to compare the synthetic (fake) data, G(z), generated by the generator with the real molecule data, x, and classify its input samples as either fake or real. The discriminator of DrugGEN (Figure 1D) is constructed using graph transformer encoder blocks, which function similarly to those of the generator. It starts by processing the annotation and adjacency matrices through linear layers in the node and edge layers to obtain the embeddings. These embeddings are then fed into the transformer encoder blocks (as described in the generator section) to transform the representations further. The discriminator is configured with a single transformer layer, featuring 8 attention heads and a hidden dimension of 128. Finally, the output node representations are processed in a prediction head composed of an MLP (input: 64, hidden layer 1: 32, hidden layer 2: 16, and output: 1) to produce the discriminator output (i.e., prediction scores utilised for real/fake evaluation). \\

\subsection{MLP Baseline Models}

The baseline model created for comparison against DrugGEN in the ablation study is a GAN with an MLP-based generator and discriminator. The architectural difference between DrugGEN and the MLP baseline is the simplified generator module. This MLP-based generator module is a modified version of the generator featured in the MolGAN study \cite{De_Cao_Kipf_2018} and consists of two dense layers coupled with ReLU activation function ending with a dropout layer. The modification involves directly processing the starting molecules instead of employing linear projections on the noise data, as was done in MolGAN. As a result, the first dense layer, where the noise was initially processed to align its dimensions with those of flattened adjacency and annotation matrices, has been omitted. Apart from that, there is a readout layer (formed by a single dense layer) to adjust the dimension size. The MLP generator is structured with two parallel linear layers designed to handle input molecules’ annotation and adjacency matrices independently:  \\

\begin{equation}
    X =  ReadOut(DropOut(ReLU(DenseLayer(ReLU(DenseLayer(X))))))
\end{equation}

\begin{equation}
    A =  ReadOut(DropOut(ReLU(DenSeLayer(ReLU(DenseLayer(A))))))
\end{equation} \\

Where X and A represent annotation and adjacency matrices, respectively. Subsequently, the processed molecules follow a similar pipeline to that of the DrugGEN system, where they are passed to the discriminator for “real/fake” evaluation. \\

The MLP-based discriminator of the baseline model takes its input as one-dimensional vectors. Vectors representing the annotation and adjacency matrices are first flattened and then concatenated. The MLP discriminator is configured with hidden layers composed of 256, 128, 64, 32, and 16 neurons, followed by an output layer with a single neuron. This is followed by the tanh activation function to map each sample to a value between [-1,1]. \\

We trained two baseline models called MLP baseline and MLP-NoTarget baseline, analogous to DrugGEN and DrugGEN-NoTarget, respectively. The implementation and training of the MLP baseline and MLP-NoTarget baseline adhere to the protocols used for DrugGEN, including the datasets, to ensure a fair comparison. The results obtained from baseline models are discussed in the ablation study. \\

The second baseline replaces the transformer encoder blocks in DrugGEN with graph convolutional network (GCN) layers in both the generator and discriminator. For the generator, three consecutive GCN layers are applied to the annotation and adjacency matrices to update the node embeddings. These are then reshaped into the generated molecule representation via a final readout layer, ensuring the correct output dimensions. In the discriminator, three GCN layers similarly encode the input molecule (real or generated) into node embeddings, which are then aggregated into a single feature vector. This vector is passed to a four-layer MLP (with an input size of 64, followed by hidden layers of 32 and 16 neurons, and a single output neuron), same with the DrugGEN discriminator, to produce the final scalar score.\\ 

We developed both MLP and GCN baselines (targeted and non-targeted versions) under the same training, hyperparameter, and evaluation regimes as DrugGEN to ensure a fair comparison. The results obtained from these baseline models are discussed in the ablation study. \\

\subsection{Loss Function}

Both DrugGEN and the baseline models utilize the WGAN loss in model training \cite{Arjovsky_Chintala_Bottou_2017}. The formulation of the WGAN loss for the end-to-end training of DrugGEN, is given below: \\

\begin{equation} 
\begin{split}
    L = (\mathbb{E}_{x\sim{p_{r}(x)}}[D_1(x)] - &\mathbb{E}_{z\sim{p_{g}(z)}}[D_1(G_1(z))])
\end{split}
\end{equation} \\

where $x$ denotes real molecules, which are experimentally validated inhibitors of the target of interest, that has been used in the discriminator of DrugGEN; $z$ denotes the input distribution of the generator; $p_{r}$ denotes real data distribution and $p_{g}$ the generated data distribution. It has been shown in the literature that using gradient penalty (GP) improves the performance of WGAN \cite{Gulrajani_Ahmed_Arjovsky_Dumoulin_Courville_2017}. Due to this, we utilized GP, and its loss is formulated as: \\

\begin{equation}
    L_{GP} = \lambda  \mathbb{E}_{\hat{x}\sim{p_{\hat{x}}(\hat{x})}}[(|| \nabla_{\hat{x}} \tilde{D}({\hat{x}})||_2 - 1)^2]
\end{equation} \\

where $\lambda$ denotes a penalty coefficient; $\hat{x}$ denotes data coming from: (i) $x$ (real data), and (ii) generated samples. $p_{\hat{x}}(\hat{x})$ refers to sampling uniformly along straight lines between pairs of points from the data distribution $p_r$ and generator distribution $p_g$ \cite{Gulrajani_Ahmed_Arjovsky_Dumoulin_Courville_2017}. We obtained our finalised loss function as via combining the losses given in Eqn. 6 and Eqn. 7: \\

\begin{equation}
    L_{total} = L + L_{GP}
\end{equation} \\

\subsection{The Training Scheme and Hyperparameters}

DrugGEN was trained with the ChEMBL compounds dataset (used as the real molecule input of the model). The ChEMBL dataset was divided randomly into training and test partitions, maintaining a ratio of 90\% and 10\%, respectively. DrugGEN commences the training with $D$ (using randomly sampled molecules from the untrained generator and real samples), and subsequently progresses to $G$. The AdamW optimizer \cite{Loshchilov_Hutter_2017} was employed with default parameters and a batch size of 128 was used. The entire training spanned 100 epochs; however, we utilised early stopping based on the validity and novelty metrics. This was mainly applied to prevent mode collapse. During the hyperparameter optimisation phase, various learning rates (1e-05, 5e-06, 1e-07, 0.0005, 0.0001, 0.005, and 0.001) were independently tested for each module. Finally, a learning rate of 1e-05 was chosen for all modules (both for G and D), mainly due to yielding module stability. The impact of different attention head numbers (4, 8, and 16) was explored to gauge their effect on model performance, and findings supported using 8 attention heads consistently across all models. Various embedding dimension sizes (16, 32, 64, 128, and 256) were evaluated for their impact on model behaviour. The analysis indicated that the embedding dimension size of 128 were optimal for both DrugGEN and DrugGEN-NoTarget models. Model depth was subject to experimentation with full model training, exploring the values of 1, 2, 4, and 8 to understand their influence on performance (results can be found at Table S1). We observed that higher depth values impeded the model’s convergence. As a result, a transformer depth of 1 was selected for the primary model. Training the reported DrugGEN and DrugGEN-NoTarget models required approximately 2-3 days each, using a single NVIDIA A5000 GPU (24 Gb VRAM). All analyses mentioned above were conducted as short/quick tests due to long training durations on our infrastructure. During inference, de novo molecules underwent correction to eliminate SMILES-based errors, facilitated by the UnCorrupt SMILES model \cite{schoenmaker2023uncorrupt}. Due to the requirement of the molecular analysis libraries we utilised, such as RDKit \cite{Landrum_2013}, back and forth conversion between graphs and SMILES was necessary at various stages of model training and validation. \\

\subsection{Performance Metrics}

The performance of the models was evaluated using two sets of metrics. The first set includes four fundamental molecular generation metrics presented from the MOSES benchmark platform \cite{Polykovskiy_Zhebrak_Sanchez_Lengeling_Golovanov_Tatanov_Belyaev_Kurbanov_Artamonov_Aladinskiy_Veselov_et_al._2020}, validity, uniqueness, internal diversity (IntDiv), and novelty. These metrics are used to assess the efficiency of the generative capabilities of models and are generally concerned with the structural properties of de novo generated molecules. Validity is calculated as the percent of the data that the RDKit's SMILES conversion function can successfully parse  \cite{Landrum_2013}. Uniqueness is the metric that checks whether there are redundant molecules in the same generated batch (i.e., molecules that are identical to each other), whereas IntDiv is the measurement of the mean pairwise dissimilarity (based on Tanimoto coefficient calculated using ECFP \cite{Rogers_Hahn_2010} based molecular fingerprints) between each molecule pair in a particular generated batch. Novelty is the ratio of the generated molecules not presented in the real molecules dataset (used in training) to all generated molecules. This study also introduced a second novelty-related metric, "novelty against the inference set (NI), " in the ablation study. NI measures the ratio of the generated molecules that are not present among the starting molecules (i.e., real molecules used as input to the generator module during an inference run). Higher values of validity, uniqueness, IntDiv, novelty, and NI indicate better performance. \\

The second set of metrics concerns the physicochemical properties of de novo molecules: quantitative estimate of drug-likeness (QED) and synthetic accessibility (SA). QED computes the drug-like quality of a molecule using molecular descriptors such as molecular weight, lipophilicity, hydrogen bond donors and acceptors, polar surface area (PSA), and others. The SA score assesses the difficulty of synthesising de novo generated molecules by comparing them with labelled molecular building blocks. Higher values of QED and lower values of SA indicate better performance. In addition to these metrics, the drug-likeness of molecules is further evaluated using Lipinski’s Rule of 5 \cite{Lipinski_Lombardo_Dominy_Feeney_1997}, Veber’s rules \cite{Veber_Johnson_Cheng_Smith_Ward_Kopple_2002}, and the Pan-Assay Interference Compounds (PAINS) filter \cite{Baell_Holloway_2010}. Lipinski's Rule and Veber's rules focus on optimising pharmacokinetics, bioavailability, and membrane permeability, while the PAINS filter eliminates compounds prone to non-specific binding or false positives, ensuring the quality of biological screening. Together, these rules and filters enhance the assessment of a molecule’s drug-like potential.  We also included the Frechet ChemNet Distance (FCD) metric to assess the physicochemical similarity between molecular datasets. This metric uses the Frechet distance applied to molecular structure embeddings generated by the ChemNet model. Finally, to evaluate structural similarity, we employed scaffold and fragment similarity metrics from MOSES, which utilise Murcko scaffolds and BRICS decomposition for comparative analysis.  Detailed descriptions of the metrics can be found in Polykovskiy et al. \cite{Polykovskiy_Zhebrak_Sanchez_Lengeling_Golovanov_Tatanov_Belyaev_Kurbanov_Artamonov_Aladinskiy_Veselov_et_al._2020} and Landrum \cite{Landrum_2013}. \\

\subsection{Molecule Docking}

For the docking study, the crystal structure of the selected target protein, AKT1 (PDB ID: 4GV1) \cite{Addie_Ballard_Buttar_Crafter_Currie_Davies_Debreczeni_Dry_Dudley_Greenwood_et_al._2013}, was prepared by Protein Preparation Wizard Madhavi \cite{Madhavi_Sastry_Adzhigirey_Day_Annabhimoju_Sherman_2013} in the Schrödinger Suite 2022-4 \cite{Release_2022} with the OPLS4 force field. Missing hydrogen atoms were added, and water molecules were removed. The pH value was set to 7.4 ± 1.0 for atom typing. The binding site of AKT1 was defined based on the literature data and included Ala177, Lys179, Lys182, Ala212, Glu228, Ala230, Glu234, Glu278, Thr291, and Asp292 \cite{Addie_Ballard_Buttar_Crafter_Currie_Davies_Debreczeni_Dry_Dudley_Greenwood_et_al._2013}. Docking studies were performed using Glide \cite{Friesner_Murphy_Repasky_Frye_Greenwood_Halgren_Sanschagrin_Mainz_2006}to determine the top-scoring binding poses for each ligand.  The Van der Waals radius scaling factor was set to 1.0, and the partial charge cut-off value was set to 0.25. Docking calculations were conducted in the Standard Precision (SP) mode using the GlideScore scoring function. The same docking procedure was applied for CDK2 (PDB ID: 4KD1) \cite{martin2013cyclin}. The crystal structure of CDK2 was prepared similarly, and its binding site was confirmed based on literature data \cite{martin2013cyclin}.  All results were visualized via PyMOL \cite{Schrodinger_2015}. \\

\subsection{Dimensionality reduction with UMAP and t-SNE}

Figure 2B present the UMAP (parameters: $n_neighbors$=50, $min_dist$=0.8 and metric=”dice”) and t-SNE (parameters: perplexity: 500, iteration: 2000) projections, respectively, of a randomly selected subset of de novo molecules from DrugGEN and DrugGEN-NoTarget models (1,000 from each), real AKT1 inhibitors (all molecules used in model training); and 50,000 ChEMBL molecules from the training dataset. In these plots, each dot represents a molecule, colours indicate their source, and the Euclidean distances roughly indicate structural similarities based on Tanimoto applied on molecular fingerprints using the MACCS (Molecular ACCess System) descriptors \cite{Durant_Leland_Henry_Nourse_2002}. \\

\subsection{DEEPScreen Model Training and Evaluation Procedure}

Both real and de novo molecules were subjected to deep learning-based drug-target interaction (DTI) prediction against the selected protein (AKT1). We employed our previously developed DTI prediction system, entitled DEEPScreen, for this purpose. Briefly, DEEPScreen employs 2-D image-based structural Kakule representations (300-by-300 pixels) of compounds (created by RDKit, Draw.MolToFile function) as input and processes them via deep convolutional neural networks to classify them as active or inactive against the target of interest \cite{Rifaioglu_Nalbat_Atalay_Martin_Cetin-Atalay_Doğan_2020}. We preferred DEEPScreen for this analysis since (i) it is straightforward to prepare input data and train/test a target-specific DTI prediction model, (ii) it only requires bioactivity data, (iii) it has high predictive performance \cite{Rifaioglu_Nalbat_Atalay_Martin_Cetin-Atalay_Doğan_2020}. DEEPScreen is completely separate/independent from DrugGEN in terms of the modelling approach, datasets, and output. Therefore, this DTI prediction-based validation is objective. \\

For this, we first trained a DEEPScreen model for the AKT1 protein using experimental bioactivity data available in ChEMBL (v35). The training data comprised 1932 active and 2414 inactive small molecules (activity threshold was selected as pChEMBL value: 6, equivalent to 1 µM concentration in terms of IC50). We split the compound dataset into train, validation and test folds using the balanced scaffold split function from the chemprop library \cite{yang2019analyzing}. By ensuring a balanced label ratio across the folds the dataset is partitioned into training, validation, and test sets containing 3.090, 774, and 968 molecules, respectively. We augmented the dataset by adding transformed versions of each active and inactive molecule created by 10-degree rotations (the images were rotated around the centre of the image), making a total of 36 images for each input sample. This was done to render the model rotation invariant. Other types of image transformations, such as translation, were irrelevant to our case since the 2-D image generation function of RDKit standardises the molecular drawings. We evaluated the model with a confidence score threshold of 0.5 for the binary classification task (active/inactive), classifying molecules with scores $\geq$ 0.5 as active which corresponds to $\geq$ 18 augmented images. We optimised the hyper-parameters of the model with respect to the classification metrics on the validation fold and measured the finalised performance of the model on the independent hold-out test fold. Afterwards, the 2-D structural images of the real and de novo molecules were generated using the same parameters of RDKit Draw.MolToFile. The images were run on the trained AKT1 model in the prediction/inference mode. Details regarding the DEEPScreen system and its training can be obtained from https://github.com/HUBioDataLab/DEEPScreen2. \\

\subsection{Molecular Dynamics Simulations}

To conduct molecular dynamics (MD) simulations, firstly, AKT1 bound complexes of both compounds (capivasertib and $MOL\_01\_027820$) were aligned. Later, the simulation system was prepared with the builtin System Builder utility of Desmond 7.3. Water molecules were placed, and Na+ ions were added to neutralize the system. A five step simulation relaxation protocol was applied based on the generated complex: (i) Brownian Dynamics was utilized at 10 K with NVT ensemble and Berendsen thermostat with small timesteps and restraints were applied on solute heavy atoms with 50.0 kcal/(mol Å2) force constant for 1 ns; (ii) the relaxation proceeded with Langevin dynamics with the NVT ensemble and Berendsen thermostat at 10 K for 120 ps, harmonic restraints continued to be applied; (iii) Langevin dynamics proceeded with NPT ensemble using Berendsen thermostat and barostat at 10 K for 120 ps with solvent restraints; (iv) the simulation started to be annealed with NPT ensemble using Berendsen thermostat and barostat for 120 ps by keeping the restraints; and (v) the solute heavy atoms were unrestrained and heating proceeded and reached to 300K with NPT ensemble using Nosé–Hoover thermostat, and Martyna–Tobias–Klein barostat for 240 ps with Langevin dynamics. Following the relaxation steps, 500ns MD simulations were run with NPT ensemble at 300K with Desmond 7.3. In total, four copies of the simulations were run for each system. The trajectory visualization was done with Maestro 13.5 \cite{Release_2022}. The simulation analyses were done with the Simulation Interactions Diagram module integrated in Maestro 13.5 \cite{Release_2023}. Desmond Trajectory Clustering utility was applied to obtain the highest occurring conformers of the ligands during each simulation. \\

\subsection{Experimental procedures for enzymatic assays}

Our experimental validation analyses were designed to assess the direct binding of de novo generated molecules to AKT1, leading to the specific inhibition of the enzyme. To accomplish this, we outsourced the enzymatic assays to Reaction Biology Europe GmbH in Germany. Using Reaction Biology’s radiometric \textsuperscript{33}PanQinase\textsuperscript{TM} activity assay, we obtained the IC\textsubscript{50} profiles of the synthesised compounds against AKT1. The detailed assay methodology can be found in the Supplementary Material (S12). \\

\section{Data availability}

The data and code to reproduce the experiments are available at the DrugGEN GitHub repo https://github.com/HUBioDataLab/DrugGEN, together with the results obtained. We used open-access data for input as described in Methods. \\

\section{Code availability}

The source code and ready-to-use trained models are available in the archived DrugGEN repository (DOI:10.5281/zenodo.15014579) and can be accessed at https://github.com/HUBioDataLab/DrugGEN. DrugGEN is also available as an online tool with a graphical interface at https://huggingface.co/spaces/HUBioDataLab/DrugGEN, where users can generate de novo molecules by employing the desired model. \\

\section{Acknowledgments}

This project was supported by TUBITAK-BIDEB 2247-A National Outstanding Researchers Program under project number 120C123. The authors thank Heval Ataş Güvenilir for guidance during the preparation of datasets and Altay Koyaş for aiding the target protein selection process. \\

\section{Author Information \& Contributions}

AU: Atabey Ünlü (atabeyunlu36@gmail.com), \\
EC: Elif Çevrim (candaselif@gmail.com), \\
AS: Ahmet Sarıgün (ahmet.sarigun@metu.edu.tr), \\
MGY: Melih Gokay Yigit (gokay.yigit@metu.edu.tr), \\
HC: Hayriye Çelikbilek (hayriye.celikbilek@gmail.com), \\
OB: Osman Bayram (osmanfbayram@gmail.com), \\
DCK: Deniz Cansen Kahraman (cansen@metu.edu.tr), \\
AO: Abdurrahman Olğaç (aolgac@gazi.edu.tr), \\
ASR: Ahmet Rifaioğlu (ahmet.rifaioglu@uni-heidelberg.de), \\
EB: Erden Banoğlu (banoglu@gazi.edu.tr) \\
TD: Tunca Doğan (tuncadogan@gmail.com).  \\ 

TD conceptualised the study and designed the general methodology. EC prepared the datasets. AS, AU, ASR and TD determined the technical details of the fundamental model architecture. AU and AS prepared the original codebase and designed and implemented initial models. AU and MGY designed, implemented, trained, tuned and evaluated numerous model variants and constructed the finalised DrugGEN models. AO and EC conducted the molecular filtering operations and physics-based (docking and molecular dynamics) experiments. AU and HC analysed the de novo generated molecules in the context of deep learning-based drug-target interaction prediction. MGY, EC and TD conducted the attention map analysis. AU, EC, AO, EB and TD evaluated and discussed the findings. EC, AU, AS, AO and TD visualised the results and prepared the figures in the manuscript. AU, EC, MGY, AS, DCK, AO and TD wrote the manuscript. AU, EC, AS, MGY and TD prepared the repository. OB and MGY constructed the online tool. TD supervised the overall study. All authors approved the manuscript. \\

\bibliographystyle{unsrt}  
\bibliography{main}

\begin{thebibliography}{10}

\bibitem{Rifaioglu_Atas_Martin_Cetin-Atalay_Atalay_Doğan_2019}
Ahmet~Sureyya Rifaioglu, Heval Atas, Maria~Jesus Martin, Rengul Cetin-Atalay,
  Volkan Atalay, and Tunca Doğan.
\newblock Recent applications of deep learning and machine intelligence on in
  silico drug discovery: methods, tools and databases.
\newblock {\em Briefings in Bioinformatics}, 20(5):1878–1912, September 2019.

\bibitem{Paul_Mytelka_Dunwiddie_Persinger_Munos_Lindborg_Schacht_2010}
Steven~M. Paul, Daniel~S. Mytelka, Christopher~T. Dunwiddie, Charles~C.
  Persinger, Bernard~H. Munos, Stacy~R. Lindborg, and Aaron~L. Schacht.
\newblock How to improve r\&d productivity: the pharmaceutical industry’s
  grand challenge.
\newblock {\em Nature reviews Drug discovery}, 9(3):203–214, 2010.

\bibitem{Bhisetti_Fang_2022}
Govinda Bhisetti and Cheng Fang.
\newblock Artificial intelligence–enabled de novo design of novel compounds
  that are synthesizable.
\newblock {\em Artificial Intelligence in Drug Design}, page 409–419, 2022.

\bibitem{Elton_Boukouvalas_Fuge_Chung_2019}
Daniel~C. Elton, Zois Boukouvalas, Mark~D. Fuge, and Peter~W. Chung.
\newblock Deep learning for molecular design—a review of the state of the
  art.
\newblock {\em Molecular Systems Design\& Engineering}, 4(4):828–849, 2019.

\bibitem{Walters_2018}
W.~Patrick Walters.
\newblock Virtual chemical libraries: miniperspective.
\newblock {\em Journal of medicinal chemistry}, 62(3):1116–1124, 2018.

\bibitem{Mouchlis_Afantitis_Serra_Fratello_Papadiamantis_Aidinis_Lynch_Greco_Melagraki_2021}
Varnavas~D. Mouchlis, Antreas Afantitis, Angela Serra, Michele Fratello,
  Anastasios~G. Papadiamantis, Vassilis Aidinis, Iseult Lynch, Dario Greco, and
  Georgia Melagraki.
\newblock Advances in de novo drug design: from conventional to machine
  learning methods.
\newblock {\em International journal of molecular sciences}, 22(4):1676, 2021.

\bibitem{Kingma_Welling_2013}
Diederik~P. Kingma and Max Welling.
\newblock Auto-encoding variational bayes.
\newblock {\em arXiv preprint arXiv:1312.6114}, 2013.

\bibitem{Gómez-Bombarelli_Wei_Duvenaud_Hernández-Lobato_Sánchez-Lengeling_Sheberla_Aguilera-Iparraguirre_Hirzel_Adams_Aspuru-Guzik_2018}
Rafael Gómez-Bombarelli, Jennifer~N. Wei, David Duvenaud, José~Miguel
  Hernández-Lobato, Benjamín Sánchez-Lengeling, Dennis Sheberla, Jorge
  Aguilera-Iparraguirre, Timothy~D. Hirzel, Ryan~P. Adams, and Alán
  Aspuru-Guzik.
\newblock Automatic chemical design using a data-driven continuous
  representation of molecules.
\newblock {\em ACS Central Science}, 4(2):268–276, February 2018.

\bibitem{Goodfellow_Pouget-Abadie_Mirza_Xu_Warde-Farley_Ozair_Courville_Bengio_2020}
Ian Goodfellow, Jean Pouget-Abadie, Mehdi Mirza, Bing Xu, David Warde-Farley,
  Sherjil Ozair, Aaron Courville, and Yoshua Bengio.
\newblock Generative adversarial networks.
\newblock {\em Communications of the ACM}, 63(11):139–144, October 2020.

\bibitem{De_Cao_Kipf_2018}
Nicola De~Cao and Thomas Kipf.
\newblock Molgan: An implicit generative model for small molecular graphs.
\newblock {\em arXiv preprint arXiv:1805.11973}, 2018.

\bibitem{Zou_Yu_Hu_Zhao_Shi_2023}
Jinping Zou, Jialin Yu, Pengwei Hu, Long Zhao, and Shaoping Shi.
\newblock Stagan: An approach for improve the stability of molecular graph
  generation based on generative adversarial networks.
\newblock {\em Computers in Biology and Medicine}, 167:107691, 2023.

\bibitem{Mahmood_Mansimov_Bonneau_Cho_2021}
Omar Mahmood, Elman Mansimov, Richard Bonneau, and Kyunghyun Cho.
\newblock Masked graph modeling for molecule generation.
\newblock {\em Nature Communications}, 12(1):3156, May 2021.

\bibitem{Ho_Jain_Abbeel_2020}
Jonathan Ho, Ajay Jain, and Pieter Abbeel.
\newblock Denoising diffusion probabilistic models.
\newblock In H.~Larochelle, M.~Ranzato, R.~Hadsell, M.~F. Balcan, and H.~Lin,
  editors, {\em Advances in Neural Information Processing Systems}, volume~33,
  page 6840–6851. Curran Associates, Inc., 2020.

\bibitem{Sohl-Dickstein_Weiss_Maheswaranathan_Ganguli_2015}
Jascha Sohl-Dickstein, Eric Weiss, Niru Maheswaranathan, and Surya Ganguli.
\newblock Deep unsupervised learning using nonequilibrium thermodynamics.
\newblock In {\em Proceedings of the 32nd International Conference on Machine
  Learning}, page 2256–2265. PMLR, June 2015.

\bibitem{Hoogeboom_Satorras_Vignac_Welling_2022}
Emiel Hoogeboom, Victor~Garcia Satorras, Clément Vignac, and Max Welling.
\newblock Equivariant diffusion for molecule generation in 3d.
\newblock In {\em Proceedings of the 39th International Conference on Machine
  Learning}, page 8867–8887. PMLR, June 2022.

\bibitem{peng2022pocket2mol}
Xingang Peng, Shitong Luo, Jiaqi Guan, Qi~Xie, Jian Peng, and Jianzhu Ma.
\newblock Pocket2mol: Efficient molecular sampling based on 3d protein pockets.
\newblock In {\em International Conference on Machine Learning}, pages
  17644--17655. PMLR, 2022.

\bibitem{schneuing2022structure}
Arne Schneuing, Yuanqi Du, Charles Harris, Arian Jamasb, Ilia Igashov, Weitao
  Du, Tom Blundell, Pietro Li{\'o}, Carla Gomes, Max Welling, et~al.
\newblock Structure-based drug design with equivariant diffusion models.
\newblock {\em arXiv preprint arXiv:2210.13695}, 2022.

\bibitem{Mitton_Senn_Wynne_Murray-Smith_2021}
Joshua Mitton, Hans~M. Senn, Klaas Wynne, and Roderick Murray-Smith.
\newblock A graph vae and graph transformer approach to generating molecular
  graphs.
\newblock {\em arXiv preprint arXiv:2104.04345}, 2021.

\bibitem{Nemoto_Kaneko_2023}
Kohei Nemoto and Hiromasa Kaneko.
\newblock De novo direct inverse qspr/qsar: Chemical variational autoencoder
  and gaussian mixture regression models.
\newblock {\em Journal of Chemical Information and Modeling}, 63(3):794–805,
  2023.

\bibitem{Richards_Groener_2022}
Ryan~J. Richards and Austen~M. Groener.
\newblock Conditional $\beta $-vae for de novo molecular generation.
\newblock {\em arXiv preprint arXiv:2205.01592}, 2022.

\bibitem{Kadurin_Nikolenko_Khrabrov_Aliper_Zhavoronkov_2017}
Artur Kadurin, Sergey Nikolenko, Kuzma Khrabrov, Alex Aliper, and Alex
  Zhavoronkov.
\newblock drugan: An advanced generative adversarial autoencoder model for de
  novo generation of new molecules with desired molecular properties in silico.
\newblock {\em Molecular Pharmaceutics}, 14(9):3098–3104, September 2017.

\bibitem{Xie_Valiente_Kim_2023}
Xuezhi Xie, Pedro~A. Valiente, and Philip~M. Kim.
\newblock Helixgan a deep-learning methodology for conditional de novo design
  of $\alpha$-helix structures.
\newblock {\em Bioinformatics}, 39(1):btad036, 2023.

\bibitem{Arús-Pous_Johansson_Prykhodko_Bjerrum_Tyrchan_Reymond_Chen_Engkvist_2019}
Josep Arús-Pous, Simon~Viet Johansson, Oleksii Prykhodko, Esben~Jannik
  Bjerrum, Christian Tyrchan, Jean-Louis Reymond, Hongming Chen, and Ola
  Engkvist.
\newblock Randomized smiles strings improve the quality of molecular generative
  models.
\newblock {\em Journal of cheminformatics}, 11(1):1–13, 2019.

\bibitem{Bagal_Aggarwal_Vinod_Priyakumar_2022}
Viraj Bagal, Rishal Aggarwal, P.~K. Vinod, and U.~Deva Priyakumar.
\newblock Molgpt: Molecular generation using a transformer-decoder model.
\newblock {\em Journal of Chemical Information and Modeling},
  62(9):2064–2076, May 2022.

\bibitem{Blaschke_Arús-Pous_Chen_Margreitter_Tyrchan_Engkvist_Papadopoulos_Patronov_2020}
Thomas Blaschke, Josep Arús-Pous, Hongming Chen, Christian Margreitter,
  Christian Tyrchan, Ola Engkvist, Kostas Papadopoulos, and Atanas Patronov.
\newblock Reinvent 2.0: an ai tool for de novo drug design.
\newblock {\em Journal of chemical information and modeling},
  60(12):5918–5922, 2020.

\bibitem{Wang_Gao_Han_Li_Chen_Rodríguez_Patón_Wang_Zheng_2023}
Xun Wang, Changnan Gao, Peifu Han, Xue Li, Wenqi Chen, Alfonso
  Rodríguez~Patón, Shuang Wang, and Pan Zheng.
\newblock Petrans: De novo drug design with protein-specific encoding based on
  transfer learning.
\newblock {\em International Journal of Molecular Sciences}, 24(2):1146, 2023.

\bibitem{yang2023cmgn}
Minjian Yang, Hanyu Sun, Xue Liu, Xi~Xue, Yafeng Deng, and Xiaojian Wang.
\newblock Cmgn: a conditional molecular generation net to design
  target-specific molecules with desired properties.
\newblock {\em Briefings in bioinformatics}, 24(4):bbad185, 2023.

\bibitem{zhang2023resgen}
Odin Zhang, Jintu Zhang, Jieyu Jin, Xujun Zhang, RenLing Hu, Chao Shen, Hanqun
  Cao, Hongyan Du, Yu~Kang, Yafeng Deng, et~al.
\newblock Resgen is a pocket-aware 3d molecular generation model based on
  parallel multiscale modelling.
\newblock {\em Nature Machine Intelligence}, 5(9):1020--1030, 2023.

\bibitem{guan20233d}
Jiaqi Guan, Wesley~Wei Qian, Xingang Peng, Yufeng Su, Jian Peng, and Jianzhu
  Ma.
\newblock 3d equivariant diffusion for target-aware molecule generation and
  affinity prediction.
\newblock {\em arXiv preprint arXiv:2303.03543}, 2023.

\bibitem{Guan_Qian_Peng_Su_Peng_Ma_2023}
Jiaqi Guan, Wesley~Wei Qian, Xingang Peng, Yufeng Su, Jian Peng, and Jianzhu
  Ma.
\newblock 3d equivariant diffusion for target-aware molecule generation and
  affinity prediction.
\newblock (arXiv:2303.03543), March 2023.
\newblock arXiv:2303.03543 [cs, q-bio].

\bibitem{Perron_Mirguet_Tajmouati_Skiredj_Rojas_Gohier_Ducrot_Bourguignon_Sansilvestri‐Morel_Do_Huu_2022}
Quentin Perron, Olivier Mirguet, Hamza Tajmouati, Adam Skiredj, Anne Rojas,
  Arnaud Gohier, Pierre Ducrot, Marie-Pierre Bourguignon, Patricia
  Sansilvestri‐Morel, and Nicolas Do~Huu.
\newblock Deep generative models for ligand‐based de novo design applied to
  multi‐parametric optimization.
\newblock {\em Journal of Computational Chemistry}, 43(10):692–703, 2022.

\bibitem{zhou2019optimization}
Zhenpeng Zhou, Steven Kearnes, Li~Li, Richard~N Zare, and Patrick Riley.
\newblock Optimization of molecules via deep reinforcement learning.
\newblock {\em Scientific reports}, 9(1):10752, 2019.

\bibitem{Gebauer_Gastegger_Hessmann_Müller_Schütt_2022}
Niklas W.~A. Gebauer, Michael Gastegger, Stefaan S.~P. Hessmann, Klaus-Robert
  Müller, and Kristof~T. Schütt.
\newblock Inverse design of 3d molecular structures with conditional generative
  neural networks.
\newblock {\em Nature Communications}, 13(1):973, February 2022.

\bibitem{Li_Zhang_Wang_Zou_Yang_Luo_Wu_Yang_Tian_Xu_2022}
Yueshan Li, Liting Zhang, Yifei Wang, Jun Zou, Ruicheng Yang, Xinling Luo,
  Chengyong Wu, Wei Yang, Chenyu Tian, and Haixing Xu.
\newblock Generative deep learning enables the discovery of a potent and
  selective ripk1 inhibitor.
\newblock {\em Nature Communications}, 13(1):6891, 2022.

\bibitem{Liu_Luo_Uchino_Maruhashi_Ji_2022}
Meng Liu, Youzhi Luo, Kanji Uchino, Koji Maruhashi, and Shuiwang Ji.
\newblock Generating 3d molecules for target protein binding.
\newblock (arXiv:2204.09410), May 2022.
\newblock arXiv:2204.09410 [cs, q-bio].

\bibitem{Rozenberg_Freedman_2023}
Eyal Rozenberg and Daniel Freedman.
\newblock Semi-equivariant conditional normalizing flows, with applications to
  target-aware molecule generation.
\newblock {\em Machine Learning: Science and Technology}, 4(3):035037, 2023.

\bibitem{Shi_Singha_Srivastava_Pu_Ramanujam_Brylinski_2022}
Wentao Shi, Manali Singha, Gopal Srivastava, Limeng Pu, J.~Ramanujam, and
  Michal Brylinski.
\newblock Pocket2drug: an encoder-decoder deep neural network for the
  target-based drug design.
\newblock {\em Frontiers in pharmacology}, 13:837715, 2022.

\bibitem{Uludoğan_Ozkirimli_Ulgen_Karalı_Özgür_2022}
Gökçe Uludoğan, Elif Ozkirimli, Kutlu~O Ulgen, Nilgün Karalı, and Arzucan
  Özgür.
\newblock Exploiting pretrained biochemical language models for targeted drug
  design.
\newblock {\em Bioinformatics}, 38(Supplement2), 2022.

\bibitem{Wang_Hsieh_Wang_Wang_Weng_Shen_Yao_Bing_Li_Cao_et_al._2022}
Mingyang Wang, Chang-Yu Hsieh, Jike Wang, Dong Wang, Gaoqi Weng, Chao Shen,
  Xiaojun Yao, Zhitong Bing, Honglin Li, Dongsheng Cao, and Tingjun Hou.
\newblock Relation: A deep generative model for structure-based de novo drug
  design.
\newblock {\em Journal of Medicinal Chemistry}, 65(13):9478–9492, July 2022.

\bibitem{Zhang_Li_Xing_Yuan_He_Sun_2023}
Yunjiang Zhang, Shuyuan Li, Miaojuan Xing, Qing Yuan, Hong He, and Shaorui Sun.
\newblock Universal approach to de novo drug design for target proteins using
  deep reinforcement learning.
\newblock {\em ACS omega}, 8(6):5464–5474, 2023.

\bibitem{Vaswani_Shazeer_Parmar_Uszkoreit_Jones_Gomez_Kaiser_Polosukhin_2017}
Ashish Vaswani, Noam Shazeer, Niki Parmar, Jakob Uszkoreit, Llion Jones,
  Aidan~N. Gomez, Łukasz Kaiser, and Illia Polosukhin.
\newblock Attention is all you need.
\newblock {\em Advances in neural information processing systems}, 30, 2017.

\bibitem{Kipf_Welling_2016}
Thomas~N. Kipf and Max Welling.
\newblock Semi-supervised classification with graph convolutional networks.
\newblock {\em arXiv preprint arXiv:1609.02907}, 2016.

\bibitem{li2021effective}
Pengyong Li, Jun Wang, Yixuan Qiao, Hao Chen, Yihuan Yu, Xiaojun Yao, Peng Gao,
  Guotong Xie, and Sen Song.
\newblock An effective self-supervised framework for learning expressive
  molecular global representations to drug discovery.
\newblock {\em Briefings in Bioinformatics}, 22(6):bbab109, 2021.

\bibitem{rong2020self}
Yu~Rong, Yatao Bian, Tingyang Xu, Weiyang Xie, Ying Wei, Wenbing Huang, and
  Junzhou Huang.
\newblock Self-supervised graph transformer on large-scale molecular data.
\newblock {\em Advances in neural information processing systems},
  33:12559--12571, 2020.

\bibitem{li2022kpgt}
Han Li, Dan Zhao, and Jianyang Zeng.
\newblock Kpgt: knowledge-guided pre-training of graph transformer for
  molecular property prediction.
\newblock In {\em Proceedings of the 28th ACM SIGKDD Conference on Knowledge
  Discovery and Data Mining}, pages 857--867, 2022.

\bibitem{zhu2017unpaired}
Jun-Yan Zhu, Taesung Park, Phillip Isola, and Alexei~A Efros.
\newblock Unpaired image-to-image translation using cycle-consistent
  adversarial networks.
\newblock In {\em Proceedings of the IEEE international conference on computer
  vision}, pages 2223--2232, 2017.

\bibitem{kim2017learning}
Taeksoo Kim, Moonsu Cha, Hyunsoo Kim, Jung~Kwon Lee, and Jiwon Kim.
\newblock Learning to discover cross-domain relations with generative
  adversarial networks.
\newblock In {\em International conference on machine learning}, pages
  1857--1865. Pmlr, 2017.

\bibitem{Addie_Ballard_Buttar_Crafter_Currie_Davies_Debreczeni_Dry_Dudley_Greenwood_et_al._2013}
Matt Addie, Peter Ballard, David Buttar, Claire Crafter, Gordon Currie,
  Barry~R. Davies, Judit Debreczeni, Hannah Dry, Philippa Dudley, Ryan
  Greenwood, Paul~D. Johnson, Jason~G. Kettle, Clare Lane, Gillian Lamont,
  Andrew Leach, Richard W.~A. Luke, Jeff Morris, Donald Ogilvie, Ken Page,
  Martin Pass, Stuart Pearson, and Linette Ruston.
\newblock Discovery of 4-amino- n -[(1 s
  )-1-(4-chlorophenyl)-3-hydroxypropyl]-1-(7 h -pyrrolo[2,3- d
  ]pyrimidin-4-yl)piperidine-4-carboxamide (azd5363), an orally bioavailable,
  potent inhibitor of akt kinases.
\newblock {\em Journal of Medicinal Chemistry}, 56(5):2059–2073, March 2013.

\bibitem{Polykovskiy_Zhebrak_Sanchez_Lengeling_Golovanov_Tatanov_Belyaev_Kurbanov_Artamonov_Aladinskiy_Veselov_et_al._2020}
Daniil Polykovskiy, Alexander Zhebrak, Benjamin Sanchez-Lengeling, Sergey
  Golovanov, Oktai Tatanov, Stanislav Belyaev, Rauf Kurbanov, Aleksey
  Artamonov, Vladimir Aladinskiy, Mark Veselov, Artur Kadurin, Simon Johansson,
  Hongming Chen, Sergey Nikolenko, Alán Aspuru-Guzik, and Alex Zhavoronkov.
\newblock Molecular sets (moses): A benchmarking platform for molecular
  generation models.
\newblock {\em Frontiers in Pharmacology}, 11:565644, December 2020.

\bibitem{Guimaraes_Sanchez-Lengeling_Outeiral_Farias_Aspuru-Guzik_2018}
Gabriel~Lima Guimaraes, Benjamin Sanchez-Lengeling, Carlos Outeiral, Pedro
  Luis~Cunha Farias, and Alán Aspuru-Guzik.
\newblock Objective-reinforced generative adversarial networks (organ) for
  sequence generation models.
\newblock (arXiv:1705.10843), February 2018.
\newblock arXiv:1705.10843 [cs, stat].

\bibitem{Fang_Pan_Shen_2023}
Yi~Fang, Xiaoyong Pan, and Hong-Bin Shen.
\newblock De novo drug design by iterative multiobjective deep reinforcement
  learning with graph-based molecular quality assessment.
\newblock {\em Bioinformatics}, 39(4):btad157, April 2023.

\bibitem{Grisoni_Moret_Lingwood_Schneider_2020}
Francesca Grisoni, Michael Moret, Robin Lingwood, and Gisbert Schneider.
\newblock Bidirectional molecule generation with recurrent neural networks.
\newblock {\em Journal of Chemical Information and Modeling},
  60(3):1175–1183, March 2020.

\bibitem{Xie_Shi_Zhou_Yang_Zhang_Yu_Li_2021}
Yutong Xie, Chence Shi, Hao Zhou, Yuwei Yang, Weinan Zhang, Yong Yu, and Lei
  Li.
\newblock Mars: Markov molecular sampling for multi-objective drug discovery.
\newblock (arXiv:2103.10432), March 2021.
\newblock arXiv:2103.10432 [cs, q-bio].

\bibitem{Olivecrona_Blaschke_Engkvist_Chen_2017}
Marcus Olivecrona, Thomas Blaschke, Ola Engkvist, and Hongming Chen.
\newblock Molecular de-novo design through deep reinforcement learning.
\newblock {\em Journal of Cheminformatics}, 9(1):48, December 2017.

\bibitem{matsukiyo2023novo}
Yuki Matsukiyo, Chikashige Yamanaka, and Yoshihiro Yamanishi.
\newblock De novo generation of chemical structures of inhibitor and activator
  candidates for therapeutic target proteins by a transformer-based variational
  autoencoder and bayesian optimization.
\newblock {\em Journal of Chemical Information and Modeling}, 64(7):2345--2355,
  2023.

\bibitem{tadesse2020targeting}
Solomon Tadesse, Abel~T Anshabo, Neil Portman, Elgene Lim, Wayne Tilley,
  C~Elizabeth Caldon, and Shudong Wang.
\newblock Targeting cdk2 in cancer: challenges and opportunities for therapy.
\newblock {\em Drug discovery today}, 25(2):406--413, 2020.

\bibitem{Rifaioglu_Nalbat_Atalay_Martin_Cetin-Atalay_Doğan_2020}
Ahmet~Sureyya Rifaioglu, Esra Nalbat, Volkan Atalay, Maria~Jesus Martin, Rengul
  Cetin-Atalay, and Tunca Doğan.
\newblock Deepscreen: high performance drug–target interaction prediction
  with convolutional neural networks using 2-d structural compound
  representations.
\newblock {\em Chemical science}, 11(9):2531–2557, 2020.

\bibitem{McInnes_Healy_Melville_2018}
Leland McInnes, John Healy, and James Melville.
\newblock Umap: Uniform manifold approximation and projection for dimension
  reduction.
\newblock {\em arXiv preprint arXiv:1802.03426}, 2018.

\bibitem{Van_der_Maaten_Hinton_2008}
Laurens Van~der Maaten and Geoffrey Hinton.
\newblock Visualizing data using t-sne.
\newblock {\em Journal of machine learning research}, 9(11), 2008.

\bibitem{ertl2009estimation}
Peter Ertl and Ansgar Schuffenhauer.
\newblock Estimation of synthetic accessibility score of drug-like molecules
  based on molecular complexity and fragment contributions.
\newblock {\em Journal of cheminformatics}, 1:1--11, 2009.

\bibitem{bickerton2012quantifying}
G~Richard Bickerton, Gaia~V Paolini, J{\'e}r{\'e}my Besnard, Sorel Muresan, and
  Andrew~L Hopkins.
\newblock Quantifying the chemical beauty of drugs.
\newblock {\em Nature chemistry}, 4(2):90--98, 2012.

\bibitem{abeer2024multi}
ANM~Nafiz Abeer, Nathan~M Urban, M~Ryan Weil, Francis~J Alexander, and
  Byung-Jun Yoon.
\newblock Multi-objective latent space optimization of generative molecular
  design models.
\newblock {\em Patterns}, 5(10), 2024.

\bibitem{jain2023multi}
Moksh Jain, Sharath~Chandra Raparthy, Alex Hern{\'a}ndez-Garc{\i}a, Jarrid
  Rector-Brooks, Yoshua Bengio, Santiago Miret, and Emmanuel Bengio.
\newblock Multi-objective gflownets.
\newblock In {\em International conference on machine learning}, pages
  14631--14653. PMLR, 2023.

\bibitem{monteiro2023fsm}
Nelson~RC Monteiro, Tiago~O Pereira, Ana Catarina~D Machado, Jos{\'e}~L
  Oliveira, Maryam Abbasi, and Joel~P Arrais.
\newblock Fsm-ddtr: End-to-end feedback strategy for multi-objective de novo
  drug design using transformers.
\newblock {\em Computers in Biology and Medicine}, 164:107285, 2023.

\bibitem{suzuki2024mothra}
Takamasa Suzuki, Dian Ma, Nobuaki Yasuo, and Masakazu Sekijima.
\newblock Mothra: Multiobjective de novo molecular generation using monte carlo
  tree search.
\newblock {\em Journal of Chemical Information and Modeling},
  64(19):7291--7302, 2024.

\bibitem{Ghosh_Dutta_Totaro_Bayoumi_2020}
Bhaskar Ghosh, Indira~Kalyan Dutta, Michael Totaro, and Magdy Bayoumi.
\newblock A survey on the progression and performance of generative adversarial
  networks.
\newblock In {\em 2020 11th International Conference on Computing,
  Communication and Networking Technologies (ICCCNT)}, page 1–8, Kharagpur,
  India, July 2020. IEEE.

\bibitem{Gui_Sun_Wen_Tao_Ye_2023}
Jie Gui, Zhenan Sun, Yonggang Wen, Dacheng Tao, and Jieping Ye.
\newblock A review on generative adversarial networks: Algorithms, theory, and
  applications.
\newblock {\em IEEE Transactions on Knowledge and Data Engineering},
  35(4):3313–3332, April 2023.

\bibitem{Janson_Valdes_Garcia_Heo_Feig_2023}
Giacomo Janson, Gilberto Valdes-Garcia, Lim Heo, and Michael Feig.
\newblock Direct generation of protein conformational ensembles via machine
  learning.
\newblock {\em Nature Communications}, 14(1):774, 2023.

\bibitem{Arjovsky_Chintala_Bottou_2017}
Martin Arjovsky, Soumith Chintala, and Léon Bottou.
\newblock Wasserstein generative adversarial networks.
\newblock In {\em Proceedings of the 34th International Conference on Machine
  Learning}, page 214–223. PMLR, July 2017.

\bibitem{Krenn_Häse_Nigam_Friederich_Aspuru-Guzik_2020}
Mario Krenn, Florian Häse, AkshatKumar Nigam, Pascal Friederich, and Alan
  Aspuru-Guzik.
\newblock Self-referencing embedded strings (selfies): A 100
  string representation.
\newblock {\em Machine Learning: Science and Technology}, 1(4):045024, December
  2020.

\bibitem{Yüksel_Ulusoy_Ünlü_Doğan_2023}
Atakan Yüksel, Erva Ulusoy, Atabey Ünlü, and Tunca Doğan.
\newblock Selformer: Molecular representation learning via selfies language
  models.
\newblock {\em Machine Learning: Science and Technology}, 2023.

\bibitem{Doğan_Atas_Joshi_Atakan_Rifaioglu_Nalbat_Nightingale_Saidi_Volynkin_Zellner_2021}
Tunca Doğan, Heval Atas, Vishal Joshi, Ahmet Atakan, Ahmet~Sureyya Rifaioglu,
  Esra Nalbat, Andrew Nightingale, Rabie Saidi, Vladimir Volynkin, and Hermann
  Zellner.
\newblock Crossbar: comprehensive resource of biomedical relations with
  knowledge graph representations.
\newblock {\em Nucleic acids research}, 49(16):e96–e96, 2021.

\bibitem{Mendez_Gaulton_Bento_Chambers_De_Veij_Félix_Magariños_Mosquera_Mutowo_Nowotka_2019}
David Mendez, Anna Gaulton, A.~Patrícia Bento, Jon Chambers, Marleen De~Veij,
  Eloy Félix, María~Paula Magariños, Juan~F. Mosquera, Prudence Mutowo, and
  Michał Nowotka.
\newblock Chembl: towards direct deposition of bioassay data.
\newblock {\em Nucleic acids research}, 47(D1):D930–D940, 2019.

\bibitem{Wishart_Feunang_Guo_Lo_Marcu_Grant_Sajed_Johnson_Li_Sayeeda_et_al._2018}
David~S Wishart, Yannick~D Feunang, An~C Guo, Elvis~J Lo, Ana Marcu, Jason~R
  Grant, Tanvir Sajed, Daniel Johnson, Carin Li, Zinat Sayeeda, Nazanin
  Assempour, Ithayavani Iynkkaran, Yifeng Liu, Adam Maciejewski, Nicola Gale,
  Alex Wilson, Lucy Chin, Ryan Cummings, Diana Le, Allison Pon, Craig Knox, and
  Michael Wilson.
\newblock Drugbank 5.0: a major update to the drugbank database for 2018.
\newblock {\em Nucleic Acids Research}, 46(D1):D1074–D1082, January 2018.

\bibitem{Landrum_2013}
Greg Landrum.
\newblock Rdkit documentation.
\newblock {\em Release}, 1(1–79):4, 2013.

\bibitem{Dwivedi_Bresson_2021}
Vijay~Prakash Dwivedi and Xavier Bresson.
\newblock A generalization of transformer networks to graphs.
\newblock (arXiv:2012.09699), January 2021.
\newblock arXiv:2012.09699 [cs].

\bibitem{Vignac_Krawczuk_Siraudin_Wang_Cevher_Frossard_2022}
Clement Vignac, Igor Krawczuk, Antoine Siraudin, Bohan Wang, Volkan Cevher, and
  Pascal Frossard.
\newblock Digress: Discrete denoising diffusion for graph generation.
\newblock {\em arXiv preprint arXiv:2209.14734}, 2022.

\bibitem{Gulrajani_Ahmed_Arjovsky_Dumoulin_Courville_2017}
Ishaan Gulrajani, Faruk Ahmed, Martin Arjovsky, Vincent Dumoulin, and Aaron~C
  Courville.
\newblock Improved training of wasserstein gans.
\newblock In {\em Advances in Neural Information Processing Systems},
  volume~30. Curran Associates, Inc., 2017.

\bibitem{Loshchilov_Hutter_2017}
Ilya Loshchilov and Frank Hutter.
\newblock Decoupled weight decay regularization.
\newblock {\em arXiv preprint arXiv:1711.05101}, 2017.

\bibitem{schoenmaker2023uncorrupt}
Linde Schoenmaker, Olivier~JM B{\'e}quignon, Willem Jespers, and Gerard~JP van
  Westen.
\newblock Uncorrupt smiles: a novel approach to de novo design.
\newblock {\em Journal of Cheminformatics}, 15(1):22, 2023.

\bibitem{Rogers_Hahn_2010}
David Rogers and Mathew Hahn.
\newblock Extended-connectivity fingerprints.
\newblock {\em Journal of chemical information and modeling}, 50(5):742–754,
  2010.

\bibitem{Lipinski_Lombardo_Dominy_Feeney_1997}
Christopher~A. Lipinski, Franco Lombardo, Beryl~W. Dominy, and Paul~J. Feeney.
\newblock Experimental and computational approaches to estimate solubility and
  permeability in drug discovery and development settings.
\newblock {\em Advanced drug delivery reviews}, 23(1–3):3–25, 1997.

\bibitem{Veber_Johnson_Cheng_Smith_Ward_Kopple_2002}
Daniel~F. Veber, Stephen~R. Johnson, Hung-Yuan Cheng, Brian~R. Smith, Keith~W.
  Ward, and Kenneth~D. Kopple.
\newblock Molecular properties that influence the oral bioavailability of drug
  candidates.
\newblock {\em Journal of medicinal chemistry}, 45(12):2615–2623, 2002.

\bibitem{Baell_Holloway_2010}
Jonathan~B. Baell and Georgina~A. Holloway.
\newblock New substructure filters for removal of pan assay interference
  compounds (pains) from screening libraries and for their exclusion in
  bioassays.
\newblock {\em Journal of medicinal chemistry}, 53(7):2719–2740, 2010.

\bibitem{Madhavi_Sastry_Adzhigirey_Day_Annabhimoju_Sherman_2013}
G.~Madhavi~Sastry, Matvey Adzhigirey, Tyler Day, Ramakrishna Annabhimoju, and
  Woody Sherman.
\newblock Protein and ligand preparation: parameters, protocols, and influence
  on virtual screening enrichments.
\newblock {\em Journal of computer-aided molecular design}, 27:221–234, 2013.

\bibitem{Release_2022}
Schrödinger Release.
\newblock {\em 1: Maestro, Schrödinger, LLC, New York, NY. 2021.[(accessed on
  10 December 2021)]}.
\newblock 2022.

\bibitem{Friesner_Murphy_Repasky_Frye_Greenwood_Halgren_Sanschagrin_Mainz_2006}
Richard~A. Friesner, Robert~B. Murphy, Matthew~P. Repasky, Leah~L. Frye,
  Jeremy~R. Greenwood, Thomas~A. Halgren, Paul~C. Sanschagrin, and Daniel~T.
  Mainz.
\newblock Extra precision glide: Docking and scoring incorporating a model of
  hydrophobic enclosure for protein-ligand complexes.
\newblock {\em Journal of Medicinal Chemistry}, 49(21):6177–6196, October
  2006.

\bibitem{martin2013cyclin}
Mathew~P Martin, Sanne~H Olesen, Gunda~I Georg, and Ernst Schonbrunn.
\newblock Cyclin-dependent kinase inhibitor dinaciclib interacts with the
  acetyl-lysine recognition site of bromodomains.
\newblock {\em ACS chemical biology}, 8(11):2360--2365, 2013.

\bibitem{Schrodinger_2015}
L.~L.~C. Schrodinger.
\newblock The pymol molecular graphics system.
\newblock {\em Version}, 1:8, 2015.

\bibitem{Durant_Leland_Henry_Nourse_2002}
Joseph~L. Durant, Burton~A. Leland, Douglas~R. Henry, and James~G. Nourse.
\newblock Reoptimization of mdl keys for use in drug discovery.
\newblock {\em Journal of Chemical Information and Computer Sciences},
  42(6):1273–1280, November 2002.

\bibitem{yang2019analyzing}
Kevin Yang, Kyle Swanson, Wengong Jin, Connor Coley, Philipp Eiden, Hua Gao,
  Angel Guzman-Perez, Timothy Hopper, Brian Kelley, Miriam Mathea, et~al.
\newblock Analyzing learned molecular representations for property prediction.
\newblock {\em Journal of chemical information and modeling}, 59(8):3370--3388,
  2019.

\bibitem{Release_2023}
Schrödinger Release.
\newblock {\em 1: Desmond Molecular Dynamics System, DE Shaw Research, New
  York, NY, 2021. Maestro-Desmond Interoperability Tools, Schrödinger}.
\newblock 2023.

\end{thebibliography}

\end{document}


\maketitle

\vspace{-8em}
\section*{Supplementary Text}

\subsection*{S1. A short review of the fundamental literature on AI-based molecule design}

Most of the initial AI-based de novo design methods focused on designing valid molecules that are typically small in terms of molecular size (via training the models on datasets such as QM9 (~9 heavy atoms on average) \cite{Ramakrishnan_Dral_Rupp_Von_Lilienfeld_2014, Ruddigkeit_Van_Deursen_Blum_Reymond_2012} and ZINC250K (~23 heavy atoms on average) \cite{Gómez-Bombarelli_Wei_Duvenaud_Hernández-Lobato_Sánchez-Lengeling_Sheberla_Aguilera-Iparraguirre_Hirzel_Adams_Aspuru-Guzik_2018}, which is an easier task compared to designing molecules that have the same size distribution as the existing drugs and bioactive molecules (~29 heavy atoms in ChEMBL molecules, on average).

One of the first generative modelling architectures used in de novo molecule design was variational autoencoders (VAE) \cite{Kingma_Welling_2013}. Gomez-Bombarelli et al. devised a molecule generation method based on VAE, where the encoding network converts discrete SMILES expressions of molecules into continuous vectors, with the decoder network subsequently reconstructing SMILES from this continuous space. Furthermore, an additional network was implemented to the system that guides the decoder by predicting properties such as drug-likeness and synthetic accessibility of the representations in the latent space \cite{Gómez-Bombarelli_Wei_Duvenaud_Hernández-Lobato_Sánchez-Lengeling_Sheberla_Aguilera-Iparraguirre_Hirzel_Adams_Aspuru-Guzik_2018}. Another generative modelling architecture, Generative Adversarial Networks (GAN) \cite{Goodfellow_Pouget-Abadie_Mirza_Xu_Warde-Farley_Ozair_Courville_Bengio_2020},  originally developed for image generation, has been employed to design de novo molecules. GANs are trained via a battle between generator and discriminator networks in a zero-sum game, where each agent tries to beat the other one by performing better at each move. The model called MolGAN uses a multilayer perceptron-based generator and graph convolutional discriminator to handle the molecule generation process \cite{De_Cao_Kipf_2018}. This method was one of the first studies to implement GANs for de novo drug design. A few GAN-based models, similar to MolGAN, were implemented in the following years \cite{Zou_Yu_Hu_Zhao_Shi_2023}. Another study set the training objective as predicting the masked node and edge labels on molecular graphs to enhance the efficiency of the molecular design process \cite{Mahmood_Mansimov_Bonneau_Cho_2021}. Denoising diffusion probabilistic models have surfaced as a promising approach as a part of the recent advancements in generative modelling \cite{Ho_Jain_Abbeel_2020,Sohl-Dickstein_Weiss_Maheswaranathan_Ganguli_2015}. Diffusion models learn the underlying data distribution by constructing a forward diffusion process and utilising a denoising model. One notable study by Hoogeboom et al. \cite{Hoogeboom_Satorras_Vignac_Welling_2022} proposes an equivariant diffusion model (EDM) for molecule generation in 3-D. The model uses forward diffusion to add noise to input atom coordinates and types and reverse diffusion to denoise the positions and types. This approach enables the generation of diverse molecular structures in the learned Euclidean space. Although diffusion models have the potential to provide high-quality samples, training and inference with them can be challenging due to difficulties in convergence, lengthy iterations, and the resource-intensive nature of computations. While other 3-D graph-based models exist \cite{peng2022pocket2mol, schneuing2022structure}, their tendency to grapple with higher computational requirements and lower success rates in solving complex problems underscores the favorability of 2-D graph-based models for providing reliable approximations. This causes an overall preference for 2-D graph-based models over their 3-D counterparts, due to their ability to offer cost-effective modelling with high generalisation capabilities, facilitating rapid analysis of extensive datasets.

\subsection*{S2. About protein kinase B (PKB) / AKT}

One of the intriguing proteins with biomedical relevance to be targeted in this context can be the AKT protein, also known as protein kinase B (PKB). AKT is an evolutionarily conserved serine-threonine kinase that plays a critical role in the development and progression of cancer \cite{Mroweh_Roth_Decaens_Marche_Lerat_Macek_Jílková_2021}. AKT is a key component of the PI3K/AKT signalling pathway, which regulates various cellular processes such as proliferation, cell growth, metabolism, and survival \cite{Manning_Cantley_2007}. The three isoforms—AKT1, AKT2, and AKT3—have distinct functions and tissue expressions, sharing a structurally conserved composition (85-90\% sequence similarity) \cite{Matheny_Jr_Adamo_2009}. The shared compositions comprise an N-terminus PH domain, a central kinase domain, and a C-terminus regulatory domain encompassing a hydrophobic motif \cite{Martorana_Motta_Pavone_Motta_Stella_Vitale_Manzella_Vigneri_2021}. In numerous cancer types, the activation of AKT signalling is frequently observed, leading to increased cell survival and uncontrolled proliferation \cite{george2022akt1,Rychahou_Kang_Gulhati_Doan_Chen_Xiao_Chung_Evers_2008, Banno_Togashi_De_Velasco_Mizukami_Nakamura_Terashima_Sakai_Fujita_Kamata_Kitano_et_al._2017, Bellacosa_De_Feo_Godwin_Bell_Cheng_Altomare_Wan_Dubeau_Scambia_Masciullo_1995, stahl2006integrating,Chen_Liang_Liu_Song_Guo_Shen_2018}. Therefore, inhibition of AKT signalling has emerged as a promising therapeutic approach for various cancers, as targeting this pathway could inhibit tumour growth and improve the efficacy of other treatments. AKT1 inhibitors may compete with adenosine triphosphate (ATP) for its binding site in the kinase domain (Figure S1). Alternatively, they may go for allosteric binding sites (i.e., the pleckstrin homology - PH domain) \cite{Thomas_Deak_Alessi_van_Aalten_2002} (Figure S1). Some important ATP-competent AKT inhibitors that reached clinical trials include uprosertib (GSK2141795), ipatasertib (GDC-0068), and capivasertib (AZD5363) 15. Recently, the latter was approved by the FDA \cite{Turner_Oliveira_Howell_Dalenc_Cortes_Gomez_Moreno_Hu_Jhaveri_Krivorotko_Loibl_2023}. Capivasertib and ipatasertib were also crystallised within the ATP binding site of the AKT1 protein (PDB ids: “4GV1” \cite{Addie_Ballard_Buttar_Crafter_Currie_Davies_Debreczeni_Dry_Dudley_Greenwood_et_al._2013}  and “4EKL” \cite{Lin_Lin_Wu_Ballard_Lee_Gloor_Vigers_Morales_Friedman_Skelton_2012}, respectively). Numerous studies in industry and academia highlight the current need for novel inhibitors that effectively target the AKT pathway.
\vspace{-0.4em}

\subsection*{S3. Overall performance evaluation and comparison (details)}

Considering individual metrics, DrugGEN displayed high performance on most fronts and did not suffer from issues such as low validity, novelty, or uniqueness—problems that afflict many competing models. Among the competing approaches, some are based on GANs. ORGAN relies on a GAN structure composed of an RNN (serving as the generator) and a CNN (serving as the discriminator) to produce conditioned molecules, while STAGAN was introduced to mitigate the mode collapse observed in conventional GANs; however, it exhibits diminished validity when processing considerably larger molecular structures within the ZINC250k dataset (with approximately 23 heavy atoms on average). In comparison, DrugGEN achieves a superior overall ranking (1 versus 12 for ORGAN and 3 for STAGAN) by maintaining robust validity (0.931, compared to 0.379 for ORGAN and 0.482 for STAGAN) alongside nearly perfect novelty (0.991, compared to 0.687 for ORGAN). On the other hand, graph-based methods offer an alternative route: MARS, which combines a graph neural network with MCMC sampling for multi-objective optimisation, shows extremely high validity (0.997), yet its novelty (0.333) remains low, yielding an overall ranking of 8, while MGM—based on an MPNN and operating as a masked graph model—achieves moderate novelty (0.722) and perfect uniqueness, although its validity (0.849) is lower than that of DrugGEN, resulting in an overall ranking of 4. \\

Furthermore, moving to transformer-based models, MolGPT employs the conventional sequence-based transformer decoder architecture—originally developed for natural language processing and trained on SMILES representations of molecules—although its novelty score (0.782) remains subpar. Additionally, CMGN, a transformer encoder–decoder model designed to incorporate target-specific information for generating inhibitor candidates, faces limitations in terms of uniqueness (0.409) and novelty (0.585), resulting in an overall ranking considerably lower than that of DrugGEN. Lastly, TRIOMPHE-BOA, a transformer-based variational autoencoder enhanced with Bayesian optimisation, demonstrates a relatively high QED score (0.819). However, its low uniqueness (0.445) and limited internal diversity (0.586) suggest that it captures only a narrow segment of the molecular space, resulting in an overall performance that does not match the balanced metrics achieved by DrugGEN. \\

Similarly, RL-based approaches such as QADD, MolDQN, and REINVENT typically aim to optimise specific properties (for example, QED) but often exhibit low novelty because they tend to generate molecules already present in the training set. DrugGEN, on the other hand, produces highly original compounds (novelty 0.991, compared to 0.307 for REINVENT, 0.341 for QADD, and 0.360 for MolDQN) while also balancing multiple metrics to secure the top overall ranking. Related to this, we also calculated the average maximum and mean pairwise Tanimoto-based similarity values between DrugGEN’s de novo molecules and all real molecules in the ChEMBL dataset and found 55\% and 11\% similarity, respectively—underscoring the high originality of our de novo molecules. DrugGEN utilises graph transformers with modified attention modules, likely contributing to its robust performance. \\

In addition, a separate group of architectures directly integrates 3D structural information of both drugs and/or proteins. For instance, RELATION employs an encoder–decoder framework that uses 3D grid representations of ligand–receptor complexes along with bidirectional transfer learning to generate target-specific molecules, achieving competitive performance (overall ranking 2) but with slightly lower validity (0.854) than DrugGEN; Pocket2Mol leverages an E(3)-equivariant graph neural network combined with an efficient conditional sampling algorithm to generate molecules from detailed 3D protein pocket features—although it achieves excellent validity (0.962), its low uniqueness (0.365) and modest QED (0.400) imply that it tends to generate similar, non-drug-like compounds; ResGen, which employs a hierarchical autoregression approach via a residue–atom parallel multiscale modelling scheme with SE(3)-equivariant modules, shows markedly lower uniqueness (0.289) and QED (0.218) and thus falls short relative to other models; and TargetDiff adopts a non-autoregressive 3D diffusion framework with an SE(3)-equivariant network to jointly generate atom coordinates and types, ensuring spatial consistency with the protein binding site, yet its overall performance (ranking 3) remains below that of DrugGEN, particularly in terms of validity. Finally, EDM (equivariant diffusion model), which failed to generate valid molecules (0.106) when trained on ChEMBL molecules in our analysis—likely due to the elevated complexity of the drug-like molecules in our ChEMBL training dataset (with an average of ~28.5 heavy atoms compared to ~20 in the GEOM dataset originally used for training EDM)—demonstrates that even models specifically designed for 3D data can struggle when faced with complex molecular structures. Moreover, the remaining architectures include BIMODAL, which generate molecules from SMILES strings using recurrent neural network architecture incorporating leveraging bidirectional sequence generation)—attain high validity (0.997) but suffer from low novelty (0.314), resulting in overall ranking of  9 that are considerably inferior to DrugGEN’s balanced performance. \\

The QED metric highlights a model’s ability to generate compounds with desirable drug-like properties; while some models (e.g., QADD, molDQN, and MARS) in Table 1 specifically aim to maximise QED values of their de novo molecules—hence those methods are omitted from the QED-based comparison—DrugGEN does not explicitly optimise for QED. However, it still generates molecules with relatively high QED values. Finally, the internal diversity (IntDiv) metric indicates the structural diversity among generated samples. DrugGEN exhibited notably high performance in IntDiv, demonstrating its ability to learn varied molecular structures from the training dataset. 

\vspace{-0.2em}

\subsection*{S4. Molecular docking and Drug-likeness analysis (details)}

To examine target binding characteristics of molecules on the AKT1 and CDK2 proteins, we conducted a molecular docking analysis (details are provided in the Methods section) on the molecules generated by DrugGEN and competing targeted generation methods, i.e., RELATION, TRIOMPHE-BOA, ResGen, Pocket2Mol and TargetDiff. Table 2 and Figure 2A display median docking scores (i.e., binding free energies - $\delta$G in kcal/mol) of the top 10\% of molecules with the best binding properties from each group/model, where a lower binding free energy value indicates a stronger affinity. Binding free energies for all molecules are given as histograms in Figure S4. \\

In Figure 2A, the percentages shown on top of each bar represent the docking performance of generative models in comparison to the real AKT1/CDK2 inhibitors. For this, we applied min-max normalisation to the median docking score of each model (parameters: maximum=-8.39 and minimum=-3.56 kcal/mol, which are the median of the top and bottom 10\% docking scores of real AKT1 inhibitors, respectively). DrugGEN achieved the highest performance with 99.98\% and -8.39 kcal/mol (Figure 2A), exhibiting a noteworthy contrast to other target-based models, especially TRIOMPHE-BOA and ResGen. Furthermore, the difference between the DrugGEN and the next best model, RELATION (i.e., -8.39 and -8.10 kcal/mol, respectively), was found to be statistically significant according to the Mann-Whitney U rank test (p-value < 0.0001). Here, DrugGEN struck a good balance between drug-like properties and target binding, making it a highly competitive model. Following DrugGEN and RELATION, TargetDiff and Pocket2Mol exhibit satisfactory docking scores (91.59\% and -7.98 kcal/mol for TargetDiff, and 91.10\% and -7.96 kcal/mol for Pocket2Mol). They also rank below DrugGEN and RELATION in most drug-likeness metrics. TRIOMPHE-BOA displayed the weakest docking performance (-6.92 kcal/mol) but achieved the highest scores in drug-likeness, including QED, SA, and Lipinski’s and Veber’s rules. This indicates that while the generated compounds possess favourable pharmacokinetic properties, they are not well-suited for targeting a specific protein. ResGen performed poorly in both docking and drug-likeness evaluations. Remarkably, DrugGEN surpassed the majority of the compared models and exhibited comparable performance to RELATION in terms of fragment (0.721) and scaffold (0.246) similarities, as well as the FCD score (20.416). It generated de novo molecules with structural features resembling known inhibitors (Table 2), while also being novel and diverse (Table 1).  \\

For CDK2, Pocket2Mol exhibited the strongest binding affinity with a normalised docking performance of 88.21\% and a median docking score of -8.80 kcal/mol, closely followed by DrugGEN at 86.99\% (-8.75 kcal/mol). The difference in docking scores was statistically significant (Mann-Whitney U test, p < 0.001). However, Pocket2Mol showed suboptimal drug-likeness characteristics, with lower QED and higher synthetic accessibility (SA) scores (Table 2), indicating challenging synthetic complexities. Also, DrugGEN outperformed Pocket2Mol in Lipinski’s and Veber’s rules, as well as PAINS filters, suggesting a higher potential for yielding pharmacologically promising compounds. When comparing DrugGEN to TargetDiff, DrugGEN exhibited significantly better docking performance (Mann-Whitney U test, p < 0.0001). On the other hand, TargetDiff had high QED and PAINS scores, indicating that it generated chemically attractive molecules. Furthermore, fragment and scaffold similarities and the FCD value of DrugGEN (0.931, 0.311 and 10.611, respectively) highlight its ability to perform meaningful structural transformations while preserving a high degree of novelty compared to the benchmark models. This further confirms that DrugGEN can generate molecules whose physicochemical properties closely align with known CDK2 inhibitors.

\subsection*{S5. The effect of input molecules on the generative performance}

DrugGEN operates by taking real molecules at the input level. To investigate how target-specific properties change based on molecular structures and characteristics, we employed our trained model and generated two sets of de novo molecules: one using known/real AKT1 inhibitors and the other using randomly selected ChEMBL molecules as inputs. We then compared the generated molecules in terms of their structural similarity to real AKT1 inhibitors and their docking properties. Molecules inferred from real AKT1 inhibitors demonstrated high fragment similarity (0.712) and moderate scaffold similarity (0.395), with an FCD of 12.723, indicating considerable structural and physicochemical overlap with existing AKT1 inhibitors. In contrast, molecules generated using random ChEMBL molecules showed comparable fragment similarity (0.714), a lower scaffold similarity (0.042), a higher FCD of 21.248 (lower is better). While both groups maintain similar fragment-level characteristics, the marked difference in scaffold similarity suggests that molecules inferred from random ChEMBL possess distinct structural cores. This architectural divergence positions these molecules as promising candidates for novel scaffold discovery, potentially expanding the structural diversity of therapeutic approaches. Whereas, utilising known AKT1 inhibitors as input during de novo generation can yield high bioactivity values, especially suitable for molecule optimisation. The advantages of these approaches differ, enabling preference based on specific objectives. \\

We also checked whether DrugGEN improves the binding characteristics of real AKT1 inhibitors when used as input molecules. For this, we compared the docking scores of de novo molecules with their corresponding input AKT1 inhibitor inputs. We found that top 10\% de novo molecules have improved the docking scores of their inputs (i.e., -8.232 and -7.680 kcal/mol, respectively), although the real AKT1 inhibitors has a slightly better score when we directly consider the top 10\% molecules on both sides, without following up with the starting molecules (i.e., -8.232 and -8.387 kcal/mol, for de novo and real inhibitors, respectively). This indicates that, in general, DrugGEN can generate novel molecules that dock as well as the known, real inhibitors of the target. Figure S3 shows three example real AKT1 inhibitors used as input, the corresponding de novo molecules generated from them, and their respective docking, QED and SA scores. In these examples, the de novo molecules achieved better docking scores than their references; however, reference molecules tend to be more synthesisable, as reflected in their lower synthetic accessibility (SA) scores. Also, drug-likeness (QED) varies. 

\subsection*{S6. Ablation study (details)}

In this analysis, we evaluated the impact of (i) the graph transformer architecture and (ii) its specialised attention mechanism on molecular generation performance. We compared the outputs of various DrugGEN and DrugGEN-NoTarget model variants with each other, as well as with the molecules in the training datasets. For consistency, we employed the same set of fundamental benchmarking metrics as in previous sections (Validity, Novelty, Uniqueness, IntDiv, QED, and SA). All models were trained using identical hyperparameter value ranges and datasets, with modifications applied only to the core module responsible for learning molecular representations. Notably, for each model variant, the best-performing model was selected based on training metrics. Furthermore, no additional post-processing steps—i.e., SMILES correction and filtering—were performed during the molecule generation process. Consequently, the results reported here may differ from those presented in other sections of the text. In addition, all metrics, with the exception of validity, were calculated solely on the subset of valid molecules. We generated 10,000 de novo molecules from each model and ran them through the metrics mentioned above. Table S1 summarises the performance of each model variant in terms of these fundamental benchmarking metrics. \\ 

To evaluate the impact of the graph transformer architecture, we utilised six distinct models: (i) DrugGEN: the default model; (ii) DrugGEN-NoTarget: a DrugGEN model that generates only random molecules featuring the same architecture as DrugGEN, but with a discriminator that accepts random real molecules as input instead of target-specific inhibitors (with the sole objective of generating drug-like molecules); (iii) MLP baseline: a targeted baseline model employing multi-layer perceptron (MLP) modules for both the generator and discriminator; (iv) MLP-NoTarget baseline: a non-targeted baseline model based on MLPs; (v) GCN baseline: a targeted baseline model employing graph convolutional network (GCN) modules, for both the generator and discriminator; and (vi) GCN-NoTarget baseline: a non-targeted baseline model based on GCNs. These baseline models are differentiated from DrugGEN by replacing the graph transformer encoder blocks with either simple MLP layers or GCN layers (please refer to the “Methods” section for further details). The purpose of comparing the primary DrugGEN model and its non-targeted version with the MLP- and GCN-based baselines is to highlight the advantages of employing graph transformers for molecule generation.
DrugGEN’s generator module takes random molecules as the initial input (i.e., starting molecules). Therefore, measuring a second novelty score against the starting molecules used during the generation process is also essential. We named this metric novelty at inference (NI) and provided the scores in Table S1 in the same column as the regular novelty measure (i.e., novelty against training data). We observed that NI scores are slightly lower than the standard novelty values, which was expected since the input of the inference run has a more direct effect on generated samples. It is also important to note that the NI scores of DrugGEN are still higher than the regular novelty scores of most of the models listed in Table 1. \\ 

In the target-specific setting, although the MLP-based targeted baseline exhibits relatively favourable quantitative estimation of drug-likeness (QED) scores, its validity score (0.512) is notably lower than that of DrugGEN (0.713) (Table S1). In contrast, the GCN-based targeted baseline achieves a higher validity score (0.754) but demonstrates poor drug-likeness metrics, as indicated by a markedly low QED (0.271) and a high synthetic accessibility (SA) score (5.215), suggesting that the GCN layers are less effective at capturing the drug-likeness related features of molecular structures. In the non-targeted setting, the DrugGEN-NoTarget model exhibits moderate performance. In contrast, the corresponding MLP-NoTarget baseline displays an even lower NI value (0.201), indicating a tendency to reproduce its input molecules, while the GCN-NoTarget variant suffers from relatively low validity (0.639) along with suboptimal QED (0.135) and SA (6.429) scores. These comparisons underscore that the graph transformer architecture within DrugGEN provides a more effective trade-off between structural validity and desirable molecular properties. Clearly, the use of graph transformer blocks contributes substantially to generating valid molecules with balanced drug-like properties. \\ 

Secondly, we performed an attention mechanism-based analysis to assess how various approaches for incorporating molecular graphs’ edges (i.e., the atomic bonds) into attention affect the generative performance. We tested five attention variants: (i)  $Att\bigodot A_m\bigodot(A_m+1)$ (the official DrugGEN model), (ii) $Att\bigodot A_m$ (the original graph transformer), (iii) $Att\bigodot(A_m+1)$, (iv) $Att \bigodot A_mul+A_add$ (FiLM-inspired modulation \cite{perez2018film}), and (v) Att (without edge-based modification). Where Att denotes the standard attention matrix computed from the transformer’s scaled dot product of query and key matrices, before applying a softmax, $A_m$ denotes the adjacency matrix, and $A_mul$ , $A_add$  are multiplicative and additive terms used in FiLM-based edge modulation. The purpose of comparing different attention mechanism variants—including the official DrugGEN model, FiLM-inspired modulation, edge modulation, edge+1 modulation, and the version without edge-based modification—is to systematically evaluate how incorporating molecular bond information into the transformer’s attention matrix influences the generative performance of the models, thereby demonstrating that an effective edge incorporation scheme is essential for capturing long-range structural dependencies and generating high-quality, target-specific molecules. 

As shown in Table S1, the official DrugGEN model exhibits a significantly higher validity (0.713) compared to other attention variants (0.110, 0.464, and 0.117 for the film, edge, and edge+1 variants, respectively). This finding suggests that the edge incorporation scheme is effective in learning molecular structures. Additionally, the variant without edge-based modifications shows a dramatic drop in validity (0.081), underscoring the critical role of bond information in generating valid molecular structures. Although some alternative variants may yield higher QED scores, these results should be interpreted with caution because the scores were computed solely on valid molecules, which was considerably low for models with reduced validity. Furthermore, the lower NI scores observed for these variants (0.503, 0.527, and 0.677 for the film, edge, and edge+1 variants, respectively) suggest that the apparent improvements in QED and SA do not necessarily reflect a genuine increase in the diversity or quality of the generated molecules. In contrast, the official DrugGEN model, with its notably high NI value of 0.924, more reliably produces molecules that are distinct from their starting points. Overall, these results demonstrate that the specific design of the attention mechanism, which incorporates the $Att\bigodot A_m\bigodot(A_m+1)$ formulation, is pivotal in enabling the model to capture long-range structural dependencies and generate high-quality, target-specific molecules.
Collectively, the ablation results confirm that the graph transformer architecture, along with the optimised attention mechanism design, is essential for achieving superior targeted molecular generation performance relative to both MLP- and GCN-based alternatives. 

\vspace{-0.2em}

\subsection*{S7. Analysis of physicochemical distributions}

We drew histograms (density plots) displaying the physicochemical property value distributions of real molecules in our training dataset and de novo generated molecules from DrugGEN models and other target-based generation models (Figure S5). Details of this analysis are given in Supplementary Material (S7). In summary, de novo molecules belonging to the DrugGEN model and real AKT1 inhibitors share a similar region of the physicochemical property space in general (i.e., DrugGEN's distribution shape and mean more closely resemble those of real AKT1 inhibitors compared to other models). This finding indicates that DrugGEN has effectively learned the physicochemical properties of the real inhibitors of the selected target protein. The details are provided below. \\

In Figure S5, each panel displays a different physicochemical property. In the left side plot of each panel, non-targeted molecules are compared (i.e., all ChEMBL molecules in our training dataset vs non-targeted de novo molecules generated by DrugGEN-NoTarget). In contrast, on the right side plot of each panel, targeted molecules are compared (i.e., the real AKT1 inhibitors vs targeted de novo molecules generated by RELATION, TargetDiff, Pocket2Mol, TRIOMPHE-BOA, ResGen and DrugGEN models). The results given for the DrugGEN model are based on real AKT1 inhibitor starting molecules. It is observed from Figure S5 that, in general, property distributions of non-targeted de novo molecules (i.e., DrugGEN-NoTarget) are similar to ChEMBL molecules, which represent this model’s training dataset, and distributions of targeted de novo molecules (i.e., DrugGEN) resemble real AKT1 inhibitors. The real AKT1 inhibitors exhibit higher molecular weights and a greater number of heavy atoms and aromatic rings than ChEMBL molecules (e.g., Figure S5A, B and H). The same pattern is also evident in DrugGEN molecules when compared with DrugGEN-Notarget. Among the target-specific models, DrugGEN has the closest distributions to AKT1 inhibitors, followed by Pocket2Mol, which demonstrated mean values closely aligned with AKT1 inhibitors for heavy atoms, aromatic rings and molecular weight. RELATION and TargetDiff exhibited slight deviations from the AKT1 inhibitor distributions, while TRIOMPHE-BOA’s molecules are concentrated within a narrow range due to how molecule generation is handled by this model (i.e., producing derivatives of a single molecule). Finally, the ResGen model exhibits the distributions most distant from the real AKT1 inhibitors. TargetDiff also exhibited a substantial divergence, particularly considering the aromatic rings. \\

Considering individual properties, DrugGEN and Pocket2Mol successfully learned the hydrogen bond acceptor (HBA). At the same time, DrugGEN and TargetDiff most accurately captured the hydrogen bond donor (HBD) value distributions of AKT1 inhibitors, as demonstrated in Figure S5E and F. Apart from that, Figure S5C displays that ChEMBL and AKT1 molecules exhibit almost identical AlogP values (i.e., a measure of the relative solubility of a molecule in two liquids: octanol and water); however, DrugGEN and DrugGEN-Notarget models both yielded a shifted mean compared to the distributions of their training datasets. RELATION achieved the closest AlogP values to AKT1 inhibitors, whereas TargetDiff and TRIOMPHE-BOA stand out as outliers. Furthermore, polar surface area (PSA) (Figure S5D) and rotatable bond (RB) (Figure S5G) value distributions of DrugGEN and DrugGEN-Notarget do not reflect the ratio between AKT1 and ChEMBL datasets; therefore, we conducted a statistical analysis to observe whether there is a notable difference between DrugGEN and DrugGEN-Notarget, although the distributions of their training datasets could not be captured. According to the Mann-Whitney U rank test, PSA distributions (Figure S5D) of DrugGEN and DrugGEN-Notarget are significantly different from each other (p-value < 0.0001); however, the same outcome was not obtained for RB distributions (Figure S5G) (p-value > 0.05). For the polar surface area, Pocket2Mol showed the greatest alignment with AKT1 inhibitors, whereas TargetDiff, TRIOMPHE-BOA and ResGen models showed higher mean values than AKT1 inhibitors. Conversely, for rotatable bonds, these models exhibited lower means compared to AKT1 inhibitors, except for TargetDiff, which achieved the best alignment considering this property. In contrast, RELATION displayed a density distribution similar to DrugGEN's. \\

The relatively modest performance of DrugGEN-NoTarget in this analysis could be explained by the fact that its training dataset (i.e., random ChEMBL molecules) is highly diverse regarding physicochemical properties since it is not limited to a particular target. 

\subsection*{S8. Evaluation of DrugGEN molecules using Lipinski, Veber and PAINS filters}
To assess their drug-related properties, de novo molecules generated by DrugGEN are subjected to the following filters: (i) Lipinski’s Rule of 5 \cite{Lipinski_Lombardo_Dominy_Feeney_1997}, (ii) Veber’s rules \cite{Veber_Johnson_Cheng_Smith_Ward_Kopple_2002}, and (iii) the pan-assay INterference compoundS (PAINS) \cite{Baell_Holloway_2010} filter, which identifies and eliminates false positives in biological screening assays such as redox cyclers, toxoflavins, polyhydroxylated natural phytochemicals, etc. In this analysis, we utilised the same 10,000 de novo generated molecules subjected to the benchmarking analysis. In the end, 41.15\% (4,115 out of 10,000) of the de novo molecules successfully passed through the filters. Subsequently, we subjected ChEMBL and AKT1 molecules to the same filters for comparison. It is observed that 57.18\% (1,342 out of 2,347) of AKT1 inhibitors and 50.80\% (50,802 out of 100,000) of randomly sampled ChEMBL dataset molecules have passed, providing roughly similar percentages to DrugGEN molecules.

\subsection*{S9. Docking analysis of the synthesised molecules (details)}

Molecule 1 ($MOL_02_045762$) exhibited the highest binding affinity (-8.948 kcal/mol), forming hydrogen bonds with Ala230 and Asp292, stabilised by a salt bridge with Asp292. Molecule 2 ($MOL_02_045795$) (-7.686 kcal/mol) demonstrated a halogen bond with Glu228, hydrogen bonds with Thr291 and Asp292, and a salt bridge with Asp292. Molecule 5 ($MOL_02_045627$) (-5.341 kcal/mol) displayed moderate binding, primarily through hydrogen bonds with Lys163, Lys179, Glu278, and Thr291, along with a salt bridge with Asp292. Molecule 4 ($MOL_02_045609$) (-8.383 kcal/mol) exhibited hydrogen bonds with Lys158, Lys179, and Asn279, salt bridges with Glu234 and Asp292, and a $\pi$-cation interaction with Lys179. Molecule 3 ($MOL_02_045758$) (-8.655 kcal/mol) formed hydrogen bonds with Thr211, Glu228, and Glu234, along with salt bridges involving Asp292. Molecule 1, 2, 4, and 5 show hydrophobic interactions as well, while Molecule 3 does not exhibit any (Figure 4B).

\subsection*{S10. Investigation of the binding properties of synthesised molecules using molecular dynamics (details)}

DFG-in motive of the structure was chosen to search for binding poses via docking. In this scenario, the polar interactions with the backbone atoms of the hinge domain residues, which help maintain the DFG motive as the -in conformation, as well as residence within the binding cavity, are important for AKT1 inhibitory effects. \\

As seen in the Root Mean Square Deviation (RMSD) values (Figures 4E and S8), all compounds remained stable, with values consistently below 1.6 Å. Similarly, Root Means Square Fluctuation (RMSF, Figures 4F and S8) values were low and no exceptional movements were observed in most simulations, except for one copy of Molecule 2. Throughout conducted simulations with 4 copies and 500 ns, capivasertib’s pyrrolopyrimidine moiety maintained hydrogen bonds for nearly the entire duration of the simulation (99\%) with both Tyr229 and Ala230. Additional hydrogen bonds were formed between the aliphatic hydroxypropyl group and Glu278 (74\%) residue in the catalytic loop. Water mediated hydrogen bonds were also observed with Lys158, Asn279, Asp292 residues, occurring on more than 30\% of the simulations (Figure 4D). 
The most potent de novo generated compound, Molecule 1, also formed hydrogen bonds between isoquinoline moiety and hinge domain residue Ala230 (96\%). Its quaternary amine interacted with the surrounding acidic residues; Glu234 (29\%), Glu278 (11\%), Asp292 (46\%). An additional hydrogen bond was observed between carbonyl group of the linker amide and the side chain of Lys179 (15\%) (Figure 4D). \\

Since the compound exhibited the most potent AKT1 inhibitory effect in the low $\mu$M range among the de novo generated compounds, we hypothesize that this is due to its shared interaction network with capivasertib. However, as its inhibitory value hasn’t reached nanomolar range, we further hypothesize that this is related to the absence of additional contacts with hinge domain and a lack of strong hydrogen bond networks within the binding cavity. Molecule 2, on the other hand, forms halogen bonds instead of hydrogen bonds with the hinge domain residues Tyr229 (p-chlorine 30\%) and Ala230 (o-chlorine 17\%, p-chlorine 19\%). Additionally, it forms hydrogen bonds with surrounding residues, including Lys179(23\%), Glu234 (17\%), Glu278 (23\%), Thr291 (13\%), Asp292 (21\%) (Figure 4D). Molecule 2 exhibits the second most potent inhibitory effect within the $\mu$M range against the AKT1 enzyme. \\

Molecule 3 include an aminopurine moiety and oxolane ring, similar to ATP. While its structure shares the hydrogen bond network with the residues surrounding the cavity, it hasn’t shown efficacy under 100 $\mu$M (Figure S8). We hypothesize that this is due to the compound’s short length and the presence of agonist-like fragmentation. Molecule 4 lacks heterocycles capable of  forming polar interactions with the hinge domain, with all its polar interactions concentrated within the core of the binding cavity. This observation, combined with the absence of hinge domain interactions, may explain its lack of efficacy below 100 $\mu$M (Figure S8). Similarly, Molecule 5 was found incapable of forming hydrogen bonds with the hinge domain residues. Its structure features aminopurine ring at the core, which forms hydrogen bonds with the surrounding polar residues, and oxolanediol moiety forms hydrogen bonds with the glycine-rich loop. However, the weak hydrogen bond network within the cavity and absence of interactions with the hinge domain might be the reason to show its lack of efficacy below 100 $\mu$M (Figure S8). \\

We also conducted the same MD analysis with an additional de novo showcase molecule from the list of 30 promising compounds apart from the in vitro tested ones, to show that our positive results were not isolated cases. For this, we selected $MOL\_01\_027820$ (Figure 3) and evaluated its binding characteristics with AKT1. We compared it with the simulation of the reference co-crystal complex structure “AKT1 - capivasertib”. According to results, both compounds maintained stable binding conformations within the AKT1 binding site. Further details about this experiment are provided in Supplementary Material (S11) and Figure S9. 

\subsection*{S11. Investigation of binding properties of the additional showcase molecule using molecular dynamics (details)}

We showcase another molecule among the 30 (shown with the identifier: “$MOL\_01\_027820$” in Figure 3), which can be denoted as a “Pyrrolo[1,2-a]pyrimidin-4(1H)-one” derivative. Figure S9A displays the reference crystal complex structure of AKT1 with the ligand capivasertib (model PDB ID: 4GV1) 26, for which the binding free energy was measured as -9.517 kcal/mol in the molecular docking analysis (on the same ligand conformer). Figure S9B shows the best docking pose of $MOL\_01\_027820$ with the binding free energy -9.686 kcal/mol obtained via the same docking protocol. This molecule is investigated further below to comprehend its binding-related dynamic properties. \\

Capivasertib (Figure S9A) and the showcased de novo molecule “AKT1 - $MOL\_01\_027820$” (Figure S9B) share the binding pattern with Ala177, Lys179 and Thr291. Hydrophobic interactions are more prominent in the AKT1 - $MOL\_01\_027820$ complex, whereas hydrogen bonds are evident in the crystal structure. This may contribute to the difference in binding affinities. The replacement of the adenine ring of the ATP with various heterocyclic moieties resulted in selective and potent inhibition of kinases, and this mode of binding constitutes more than 90\% of kinase inhibitors \cite{Sharma_Gupta_2022}. These inhibitors bind the same binding pocket as ATP and mimic the adenine group’s interactions with the hinge residues. Here, we see a similar orientation with capivasertib and $MOL\_01\_027820$ in the AKT1 kinase binding site. However, the hydrogen bond interactions with the backbones of the hinge residues were not directly observed in the docking-derived binding pose of $MOL\_01\_027820$ (Figure S9B). To further evaluate the relaxation and adaptation of the binding site and to reveal binding characteristics of capivasertib and $MOL\_01\_027820$, 500 ns long molecular dynamics (MD) simulations were conducted with four copies (please see “Methods” for details). \\

The output of MD simulations is shown in Figure S9, panels C to I. The AKT1 structure mostly consists of $\alpha$ helices, and visual inspection of the simulations displayed the conservation of secondary structures during the simulations of both complexes. As seen in RMSD plots (Figure S9 panels F to I), both simulations (with their simulation copies) reached stable binding patterns. Hydrogen bonding networks between the hinge domain residues Glu228 and Ala230 occurred with high ratios in all simulations. Van der Waals interactions were also observed with the hydrophobic moieties of the surrounding residues. The gatekeeper residue Met227 remained closed and supported the residence of the ligand within the binding cavity for both complexes. Additionally, Lys179 in $\beta$-sheet continued to form a salt bridge with Glu234 in the $\alpha$C-helix located at the N terminal domain of AKT1. This was a supporting factor to keep the ligand within the binding site. With the help of this ionic interaction, the protein stayed in DFG-in and $\alpha$C-helix in conformation in the simulations. \\

Capivasertib remained stable during all simulations (RMSD <1, Figure S9 panels F and H). The interaction analysis of capivasertib showed that hydrogen bonds with the hinge domain (via Glu228 -99\%- and Ala230 -99\%- residues) were conserved during the simulation. An additional hydrogen bond was seen with Glu278 (74\%). $MOL\_01\_027820$ showed two different binding patterns (pattern-1 and -2, shown in Figure S9 panels D and E, respectively). Those patterns were seen in two individual copies of the simulations. Pattern-1 was observed in the second and the third replica simulations, and pattern-2 was observed in the first and the fourth replicas (Figure S9G). The first pattern was similar to the docking-derived starting conformation (RMSDaverage = 1.2, Figure S9 panels D, G, and I) and relatively close to the binding pattern with the compound capivasertib. However, the second pattern showed 110.2 degrees of dihedral tilting (RMSDaverage 3.1, Figure S9 panels E, and G) following atom number 15 and above (Figure S9I) but still conserved the essential interactions with the hinge domain and remained within the binding cleft and conserved the protein’s general conformation. Further comments about MD results of the showcase molecule can be found below. \\

\underline{Capivasertib–AKT1:} \\

Capivasertib remained stable during all simulations (RMSD <1, Figure S9 panels F and H). The interaction analysis of capivasertib showed that hydrogen bonds with the hinge domain (via Glu228 -99\%- and Ala230 -99\%- residues) were conserved during the simulation. An additional hydrogen bond was seen with Glu278 (74\%). Also, several water bridges were observed between the amine group located on the piperidine ring and Glu278 (74\%), Asp292 (90\%) residues, amide group and Lys158 (33\%), Glu234 (11\%), Asn279 (28\%), Asp292 (36\%) residues, and hydroxyl group and the Lys276 (19\%) residue. The quaternary amine group formed a salt bridge with Glu294 (94\%), and the p-Chlorobenzyl group showed cation-$\pi$ interactions with Lys179 (11\%) (Figure S9C). \\

\underline{$MOL\_01\_027820$–AKT1:} \\

$MOL\_01\_027820$ showed two different binding patterns (pattern-1 and -2, shown in Figure S9 panels D and E, respectively). Those patterns were seen in two individual copies of the simulations. Pattern-1 was observed in the second and the third replica simulations and pattern-2 was observed in the first and the fourth replicas (Figure S9G). The first pattern was similar to the docking-derived starting conformation (RMSDaverage = 1.2, Figure S8 panels D, G, and I) and relatively close to the binding pattern with the compound capivasertib. However, the second pattern showed 110.2 degrees of dihedral tilting ($RMSD_average$ 3.1, Figure S9 panels E, and G) following atom number 15 and above (Figure S9I) but still conserved the essential interactions with the hinge domain and remained within the binding cleft and conserved the protein’s general conformation. The hydrogen interactions with Glu228 (54\% and 65\%, respectively given for pattern-1 and -2) and Ala230 (83\% and 66\%, respectively given for pattern-1 and -2) backbone atoms were formed (which did not exist in the docking-derived pose) with high occupancy values in the hinge domain. However, the remaining part of the molecule showed torsional rotations observed after the 12th and, more importantly, after the 16th atoms (Figure S9I). Therefore, slight differences in the bound clefts and the interaction networks were seen during simulations. Pattern-1 showed that the hydroxyl group connected to the fluorobenzyl moiety forms hydrogen bonds with Glu234 (44\%) and water bridges with Lys158 (48\%), Glu234 (20\% and 63\%), and Tyr437 (52\%) with the same residue. The second hydroxyl group located at the center of the ligand showed hydrogen bonds with Leu156 (22\%) and Glu234 (16\%) and water bridges with Glu234 (16\%) and Leu156 (19\%) residues. $\pi$-$\pi$ interactions were seen with fluorophenyl and Phe237 (53\%) and Phe442 (18\%) residues (Figure S9D). In pattern-2, the water network around the acid hole (Glu234 and Asp292) was mostly broken and only two water bridges remained with Leu158 (37\%) and Glu234 (10\%). Hydrogen bonds were observed between the central hydroxyl group and Leu156 (11\%), Glu234 (67\%) residues and amine group in the heterocyclic ring and Asp292 (10\%) residue. The rotated conformation was found to form Van der Waals interactions with the hydrophobic portions of the surrounding lysine residues and orthogonal multipolar interactions with fluorine atoms of the ligand and carbonyl group of Lys163 (Figure S9E). 

\vspace{-0.3em}

\subsection*{S12. Experimental procedures for enzymatic assays (details)}

Based on Reaction Biology’s radiometric \textsuperscript{33}PanQinase\textsuperscript{TM} activity assay, the IC\textsubscript{50}
profiles of the synthesised de novo compounds were retrieved against AKT1 and the IC\textsubscript{50} values were determined by testing ten different concentrations of each compound, ranging from 100 $\mu$M to 3 nM with semi-logarithmic dilution, in singlicate. The compounds were later dissolved to 0.01 M with DMSO. 100 $\mu$l of those stock solutions were transferred into the master plate. 100\% DMSO were also used as control groups. \\

As a parameter for assay quality, the Z factor \cite{zhang1999simple} for the low and high controls of each assay plate (n = 8) was used. While the low control includes substrate but no enzyme and no test compound, the high control includes the enzyme, substrate but not the test compound. RBE s criterion for repetition of an assay plate is a Z - factor below 0.4 \cite{iversen2006comparison}. The Z - factors for this study was calculated as 0.79. Additionally, staurosporine, a protein kinase inhibitor, was tested in parallel as a quality control. Staurosporine’s IC50 value was in the expected range for the protein kinase. AKT1 enzyme was produced from human cDNAs and affinity tags were removed during the purification. The purity of the enzyme was examined by SDS-PAGE/Coomassie staining and the structure was controlled by mass spectroscopy. \\

The assay was performed on 96-well ScientiPlatesTM from PerkinElmer (Boston, MA, USA) in 50 $\mu$L reaction volume containing 11.9 nM AKT1. The assay environment was also composed of 25 $\mu$L of assay buffer (standard buffer/[$\gamma$-33P]-ATP), 10 $\mu$L of ATP solution in H2O (1 $\mu$M), 5 $\mu$L of test compound (in 10 \% DMSO), 10 $\mu$L of AKT1/GSK3(14-27) mixture where the amount of the substrate is 2 $\mu$g. The reaction cocktails were incubated at 30°C for 1 hour. The reaction was stopped with 50$\mu$L of 2\% (v/v) H3PO4. The plates were aspirated and washed two times with 200 $\mu$L 0.9\% (w/v) NaCl. Incorporation of 33Pi was determined with microplate scintillation counter (Microbeta, Wallac). The assay buffer contains 70 mM HEPES-NaOH pH 7.5, 3 mM MgCl2, 3 mM MnCl2, 3 $\mu$M Na-orthovanadate, 1.2 mM DTT, 50 $\mu$g/ml PEG20000 and [$\gamma$-33P]-ATP contains approximately 8x105 cpm per well. Below is Supplementary Equation (SE1), which details how the response is calculated for each concentration. \\

Remaining Activity(\%) = 100 x [(cpm of compound – low control) / 			(high control – low control)]							        (SE1)
The remaining activities of each concentration of the compound and the IC50 values were calculated using Quattro Workflow V3.1.1 (Quattro Research GmbH, Munich, Germany; www.quattro-research.com). The IC50 values were determined with Sigmoidal response (variable slope) with parameters 100\% (for the top value) to 0\% (for the bottom value) and fit with least-squares. Hill slopes were reported and the ones with higher than -0.4 is indicative that the curve is not sigmoidal, very flat or not descending. 

\subsection*{S13. Attention map visualisation (additional information)}

Transformer-based models weigh intrinsic pairwise relationships between data tokens through the attention mechanism, which can be utilised to visualise the important relationships in the data and, therefore, interpret what the model learns. Adopting this strategy, we visualised the average attention scores (produced by the transformer encoder module of the generator network) for each atom in the selected de novo generated molecules to assess critical atoms and bonds according to the model's perspective. We aimed to observe the correspondence between the atoms that received high attention scores and those interacting with AKT1 protein’s binding pocket residues according to the molecular docking analysis. A high correspondence would indicate that the model learned which atoms are critical for enabling physical interaction with the target. \\

In Figure 5, interactions between the atoms of de novo molecules and residues in the AKT1 binding pocket are visualised via 2-D protein-ligand interaction diagrams for three different de novo molecules (Figure 5A, C, and E), using a 4 Angstrom cutoff. Also, in Figure 5, the same de novo molecules’ atoms with the highest average attention scores are visualised with green, yellow and red, indicating the importance appointed by the model (Figure 5B, D, and F). The de novo molecules employed in this analysis were randomly selected from the 30 promising molecules given in Figure 3. \\

Considering all three cases in Figure 5, the model identifies most of the atoms interacting with the target residues by assigning high attention scores. For example, in Figure 5A and B, N+H atoms forming both salt bridges and hydrogen bonds with the residue Glu234, and HN and N atoms forming hydrogen bonds with the residues Asp274 and Ala230, respectively, are assigned the highest attention values (i.e., the green coloured ligand atoms). The same pattern is observed for other interactions in Figure 5. Overall, 20 out of 22 (~91\%) ligand atoms in pairwise interactions were detected by the generator network of DrugGEN. Looking at the same examples from the other angle, 20 out of 24 (~\%83) ligand atoms with top attention scores were in contact with the target residues, thus playing a critical role in the interaction. Even though the model did not use any data regarding pairwise molecular interactions during training and validation, it still managed to learn substructures crucial for binding to AKT1 and designed molecules accordingly. We provided two additional examples in Figure S10 with a similar outcome.

\newpage
\section*{Supplementary Figures}

\renewcommand{\figurename}{Figure S}

\setcounter{figure}{0}

\begin{figure}[h]
    \centering
    \includegraphics[width=\textwidth,clip=True]{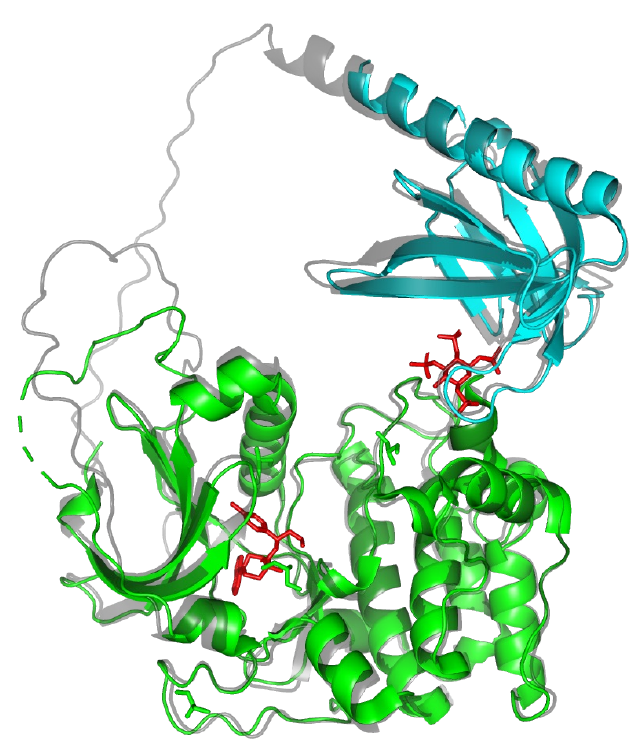}
    \let\nobreakspace\relax
    \caption{General overview of full AKT1 protein structure (gray: AlphaFold structure based on Uniprot ID: P31749, green: kinase (orthosteric) domain (PDB: 4GV1), cyan: PH (allosteric) domain (PDB: 1H10), red: ligands).}
    
    \label{fig:img1}
\end{figure}

\setcounter{figure}{1}

\begin{figure}[h]
    \centering
    \includegraphics[width=\textwidth,clip=True]{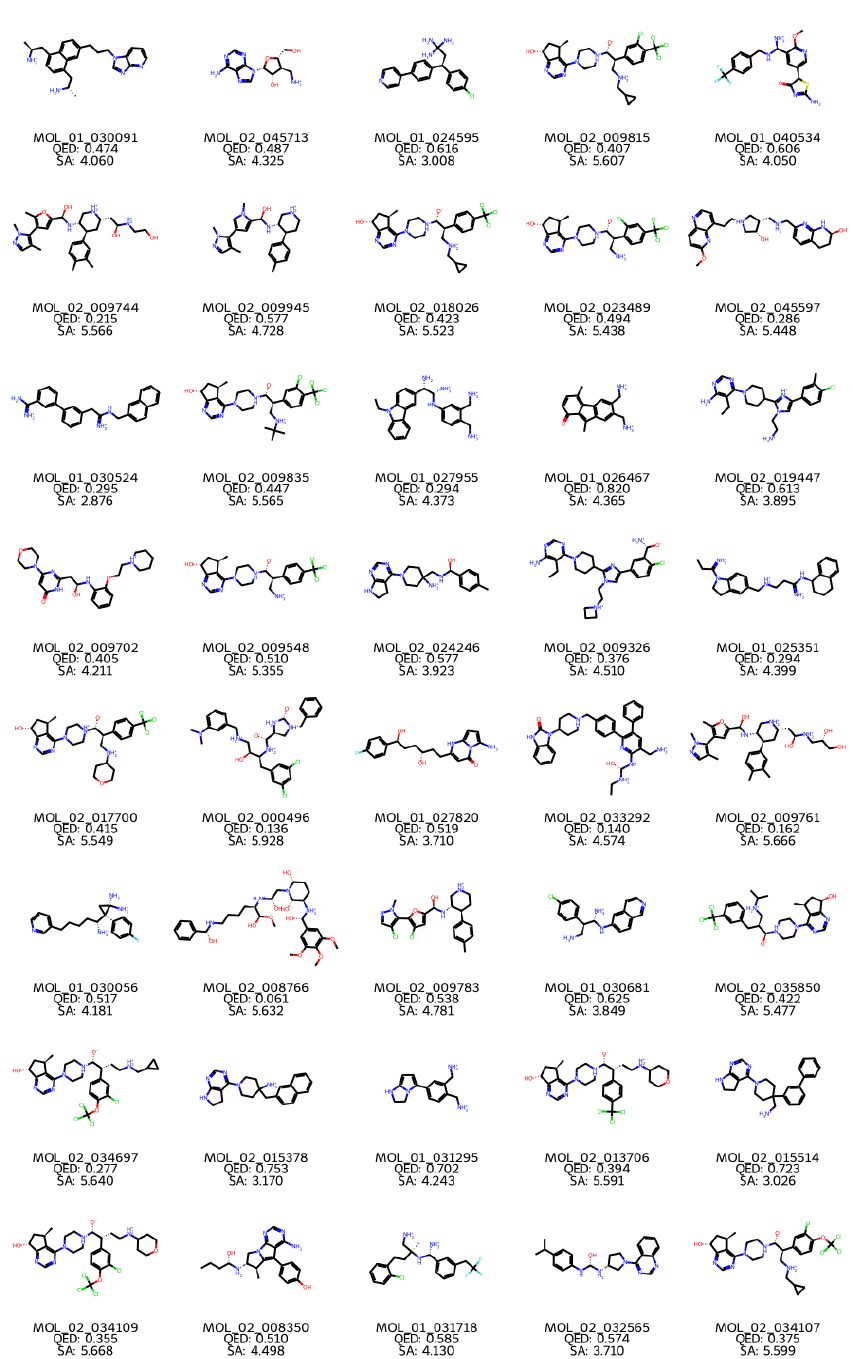}
    \let\nobreakspace\relax
    \caption{Forty molecules with predicted docking scores surpassing that of the native ligand capivasertib in the crystal complex structure of AKT1 “4GV1” (i.e., -9.517 kcal/mol).}
    
    \label{fig:img2}
\end{figure}

\setcounter{figure}{2}

\begin{figure}[h]
    \centering
    \includegraphics[width=\textwidth,clip=True]{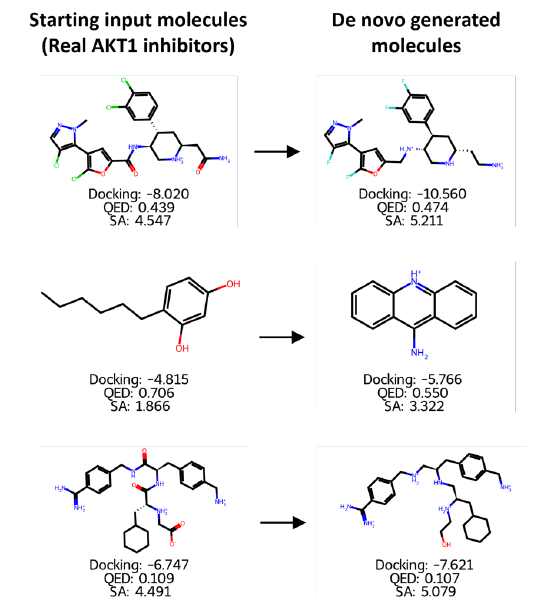}
    \let\nobreakspace\relax
    \caption{Three selected starting input molecules and their corresponding de novo-generated molecules. The left column shows the reference real AKT1 inhibitors, while the right column displays the de novo-generated molecules along with their docking scores, QED, and SA values. For the first pair, the Tanimoto similarity is 0.238 for Morgan fingerprints and 0.794 for MACCS, with Cl-to-F substitutions significantly impacting similarity. For the second pair, the lowest similarity (0.073 for Morgan and 0.109 for MACCS) indicates the highest structural novelty. For the last pair, scores of 0.435 for Morgan and 0.755 for MACCS suggest a moderate transformation, with the removal of oxygen atoms with double bonds contributing to the difference.}
    
    \label{fig:img3}
\end{figure}

\setcounter{figure}{3}

\begin{figure}[h]
    \centering
    \includegraphics[width=\textwidth,clip=True]{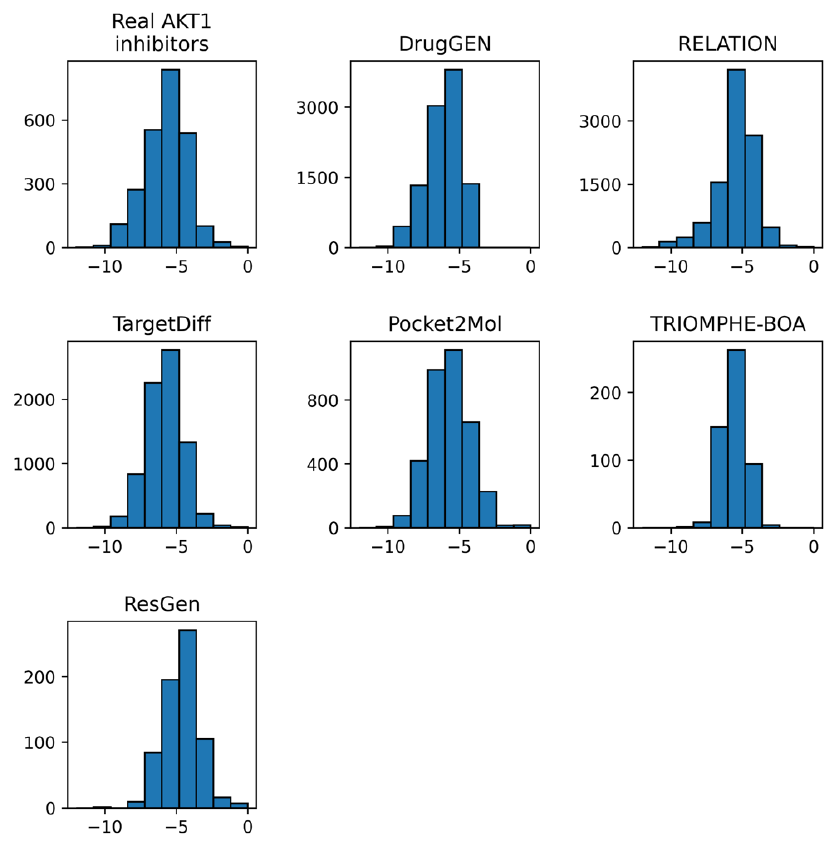}
    \let\nobreakspace\relax
    \caption{Docking score distributions of real AKT1 inhibitors, DrugGEN and other target-centric molecule generation models, including RELATION, TargetDiff, Pocket2Mol, TRIOMPHE-BOA and ResGen.}
    
    \label{fig:img4}
\end{figure}

\setcounter{figure}{4}

\begin{figure}[h]
    \centering
    \includegraphics[width=\textwidth,clip=True]{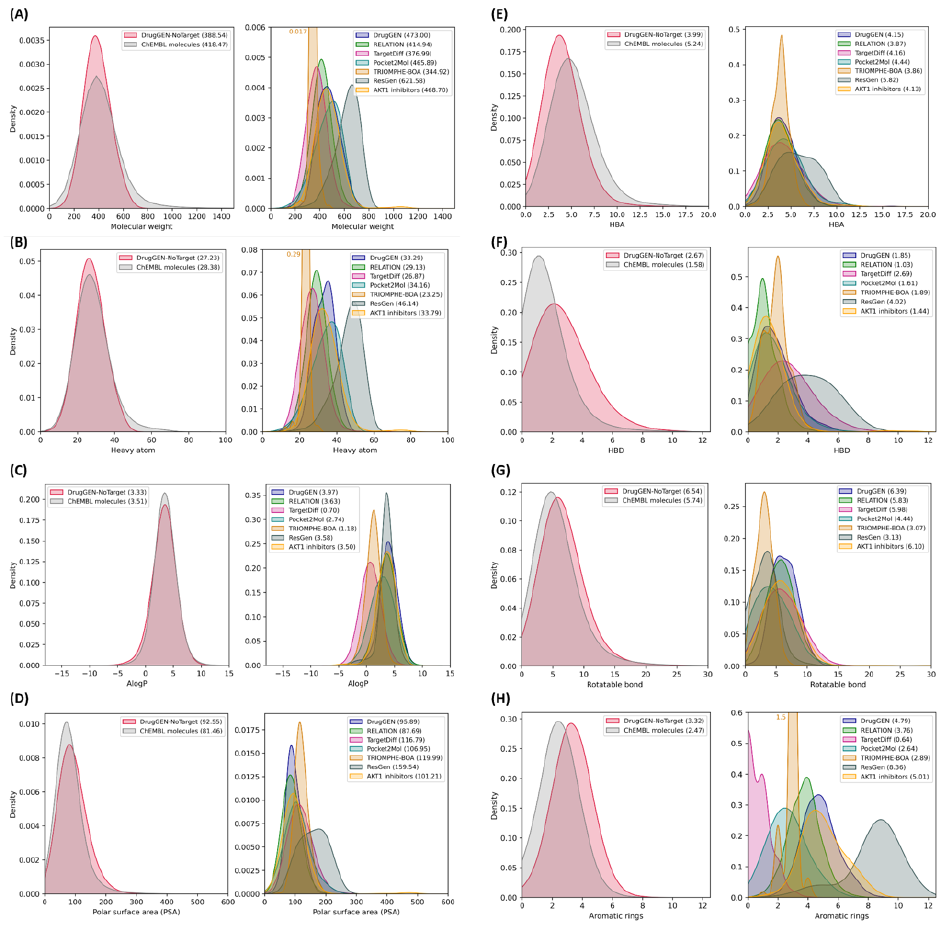}
    \let\nobreakspace\relax
    \caption{Physicochemical property value distributions of de novo molecules generated by target-centric and non-targeted models, randomly selected ChEMBL molecules and real AKT1 inhibitors for \textbf{(A)} molecular weight, \textbf{(B)} number of heavy atoms, \textbf{(C)} AlogP, \textbf{(D)} polar surface area (PSA), \textbf{(E)} hydrogen bond acceptor (HBA), \textbf{(F)} hydrogen bond donor (HBD), \textbf{(G)} rotatable bonds, and \textbf{(H)} aromatic rings. In each panel, DrugGEN-NoTarget and random ChEMBL molecules are given in the plot on the left side, whereas the targeted models’ distributions are on the right side (i.e., DrugGEN, RELATION, TargetDiff, Pocket2Mol, TRIOMPHE-BOA, ResGen and real AKT1 inhibitors). The numbers in parentheses in figure legends indicate mean values.}
    
    \label{fig:img5}
\end{figure}

\setcounter{figure}{5}

\begin{figure}[h]
\vspace{-8em}
    \centering
    \includegraphics[width=\textwidth,clip=True]{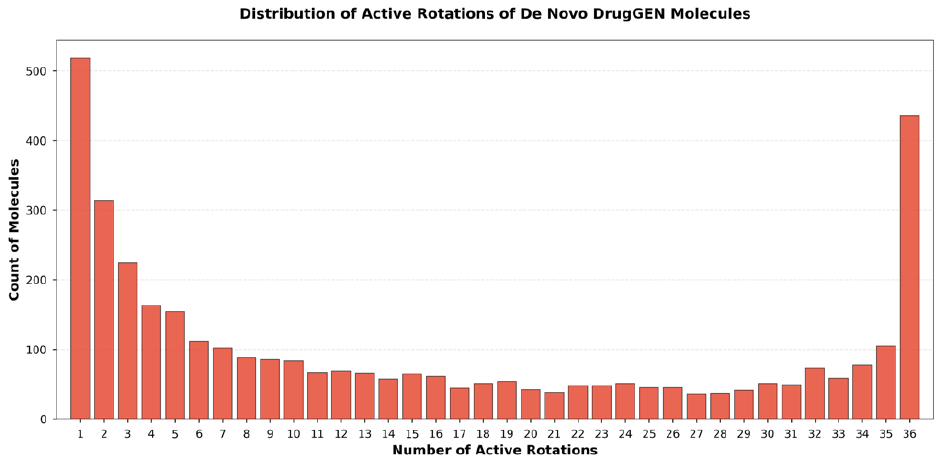}
    \let\nobreakspace\relax
    \caption{DEEPScreen drug-target interaction prediction analysis results (see “Methods” in the main text). The plot shows the confidence level histogram of 3,655 de novo compounds generated by the DrugGEN model with at least 1 rotated molecular image (out of the total of 36 rotated images generated for a single molecule) predicted as active against the AKT1 protein. 1,382 of these molecules are predicted with a sufficiently high confidence score to be accepted as bioactive molecules against AKT1 (i.e., a 50\% score cut-off, corresponding to 18 actively predicted rotated images).}
    
    \label{fig:img6}
\end{figure}

\setcounter{figure}{6}

\begin{figure}[h]
    \centering
    \includegraphics[width=12cm,clip=True]{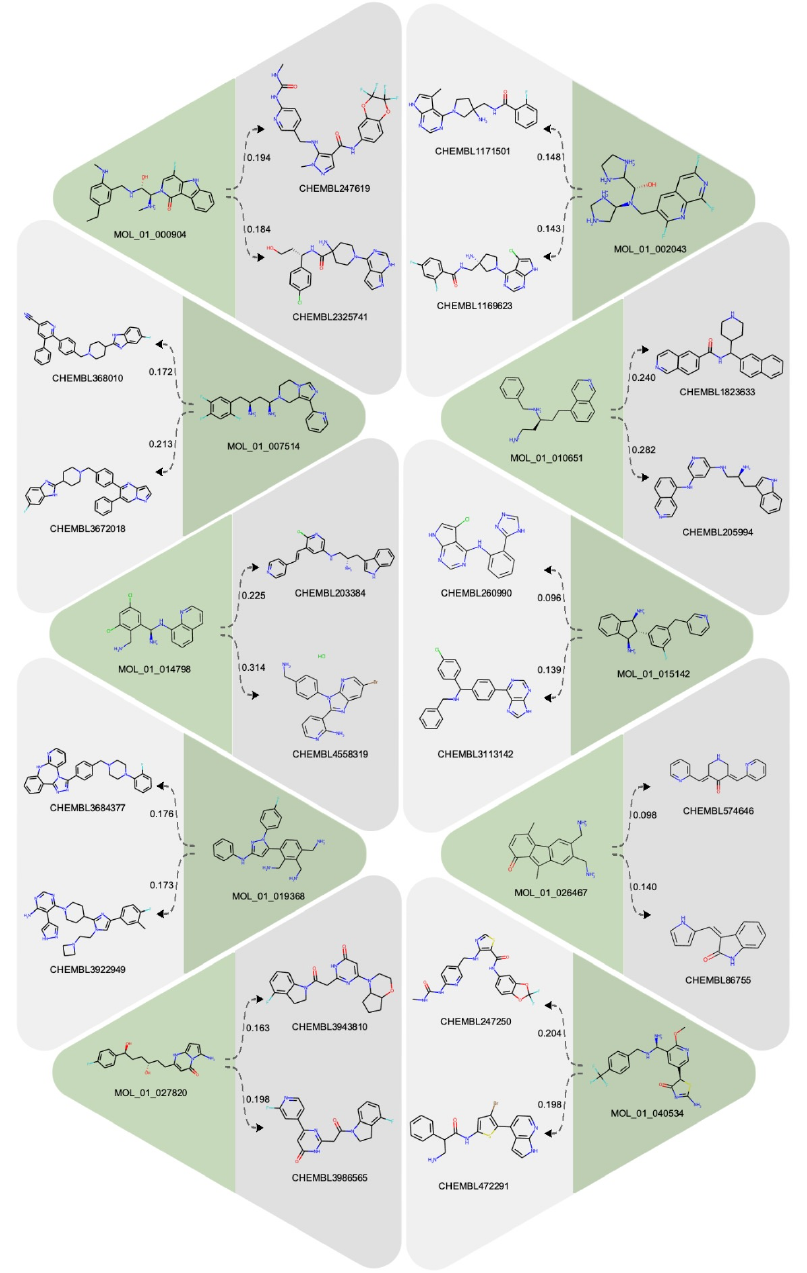}
    \let\nobreakspace\relax
    \caption{Ten randomly selected samples from 30 promising de novo molecules (displayed in Figure 3 of the main text) are highlighted in green. Each DrugGEN molecule is shown with the most structurally similar two training dataset (ChEMBL) molecules (highlighted with grey). Real (ChEMBL) molecules are selected based on their substructure similarity to the de novo designs using the Tanimoto coefficient on MACCS fingerprints.}
    
    \label{fig:img7}
\end{figure}

\setcounter{figure}{7}

\begin{figure}[h]
    \centering
    \includegraphics[width=\textwidth,clip=True]{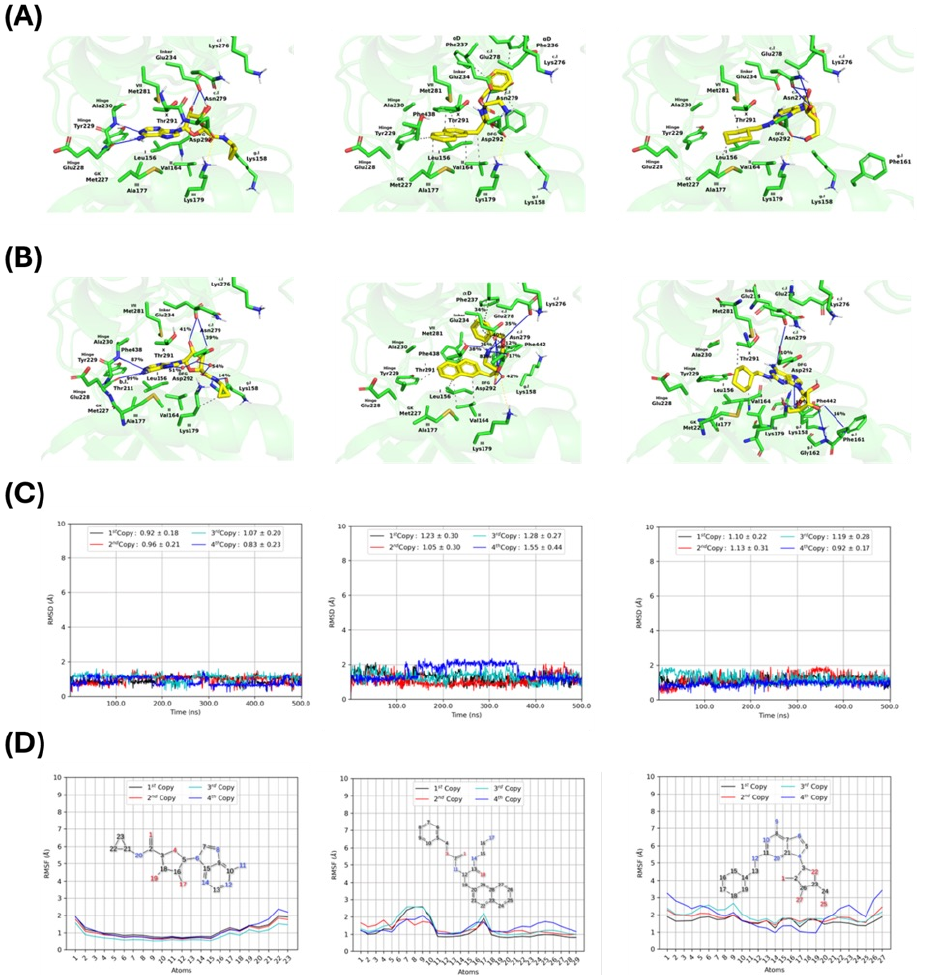}
    \let\nobreakspace\relax
    \caption{Structural analysis of three de novo generated molecules (Mol. ID 3–5) selected for experimental validation. \textbf{(A)} Initial binding orientations of Molecules 3, 4, and 5 at the starting point of molecular dynamics (MD) simulations. \textbf{(B)} Key protein-ligand interactions observed during MD simulations, visualized with interacting residues and interaction types. The depicted poses represent the most populated conformations from each simulation. \textbf{(C)} Root-mean-square deviation (RMSD) values of Molecules 3, 4, and 5 in complex with AKT1. \textbf{(D)} Root-mean-square fluctuation (RMSF) values of ligand atoms in the same complexes. Abbreviations: Exp. Mol. ID: Experimental molecule identifier. I-VII represents $\beta$-sheet numbers, g.l represents glycine-rich loop, c.l represents catalytic loop, GK represents gatekeeper residue, and xDFG represents highly conserved kinase residues, linker represents the loop that connects the hinge domain to $\alpha$C-helix. Gray dashed lines represent Van der Waals interactions. Blue lines represent hydrogen bonds and water bridges. Green lines indicate halogen bonds. Yellow dashed lines represent salt bridges. Directional interactions were noted only if the occupancy value was found above 10\%, however, for visual clarity occupancy values of the water bridges were noted only if they were above 30\%.}
    
    \label{fig:img8}
\end{figure}

\setcounter{figure}{8}

\begin{figure}[h]
    \centering
    \includegraphics[width=\textwidth,clip=True]{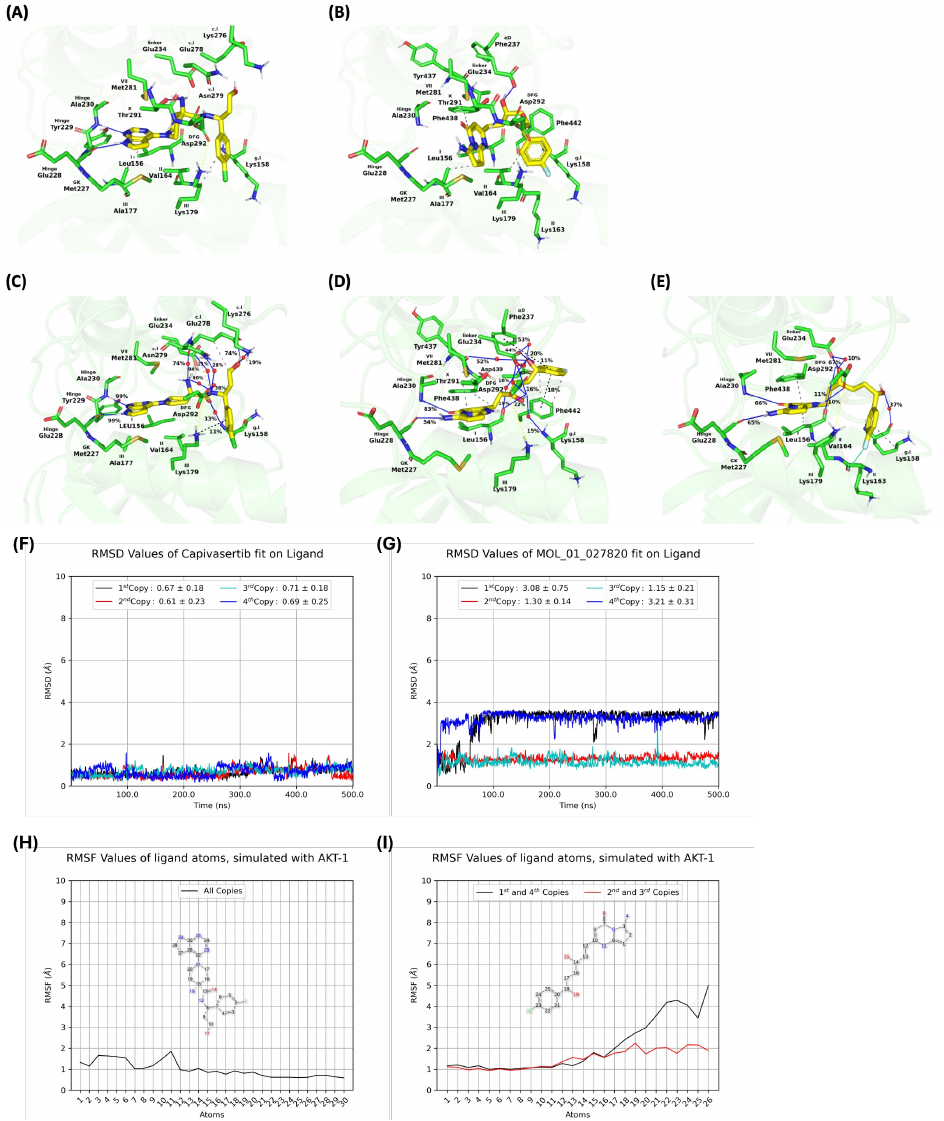}
    
    \label{fig:img9}
\end{figure}
\clearpage
\let\nobreakspace\relax
\captionof{figure}{The output of the structural analysis of selected molecules: Capivasertib (the ligand from the crystal structure PDB ID: 4GV1) 26  and the showcase de novo generated molecule: “Pyrrolo[1,2-a]pyrimidin-4(1H)-one” derivative ($MOL\_01\_027820$ in Figure 3). Block I: Molecular docking results: \textbf{(A)} AKT1 crystal complex structure with the co-crystallised ligand: Capivasertib (binding free energy in docking: -9.517 kcal/mol), and \textbf{(B)} The best pose in the molecular docking of the showcase de novo generated molecule $MOL\_01\_027820$, to the structurally resolved binding site of AKT1 (binding free energy in docking: -9.686 kcal/mol). Gray dashed lines represent Van der Waals interactions. Blue lines represent hydrogen bonds. Yellow dashed lines represent salt bridges. Block II: The molecular dynamics simulation output. The occupancy values of: \textbf{(C)} the single binding pattern of capivasertib, \textbf{(D)} the first binding pattern of $MOL\_01\_027820$, and \textbf{(E)} the second binding pattern of $MOL\_01\_027820$ in complex with AKT1 obtained from the MD simulations. Noninteracting backbone atoms are hidden for visual clarity. Only the interactions above 10\% occupancy values are reported. Domain names are noted on the figure as given in the KLIFS database \cite{van2014klifs}. Abbreviations: I-VII represents $\beta$-sheet numbers, g.l represents glycine-rich loop, c.l represents catalytic loop, GK represents gatekeeper residue, and xDFG represents highly conserved kinase residues, linker represents the loop that connects the hinge domain to $\alpha$C-helix. Gray dashed lines represent Van der Waals interactions. Blue lines represent hydrogen bonds and water bridges. Yellow dashed lines represent salt bridges. The cyan line represents orthogonal multipolar interaction. MD-derived occupancy values are shown on each interaction. Block III: RMSD plots of the compounds in MD simulations: \textbf{(F)} capivasertib and \textbf{(G)} $MOL\_01\_027820$ in each MD simulation. The illustration of rotation by RMSF plots of compounds: \textbf{(H)} capivasertib and \textbf{(I)} $MOL\_01\_027820$ for different binding patterns observed in MD simulations. For visual clarity, similar RMSF values on ligand atoms are calculated together.}

\setcounter{figure}{9}

\begin{figure}[h]
    \centering
    \includegraphics[width=\textwidth,clip=True]{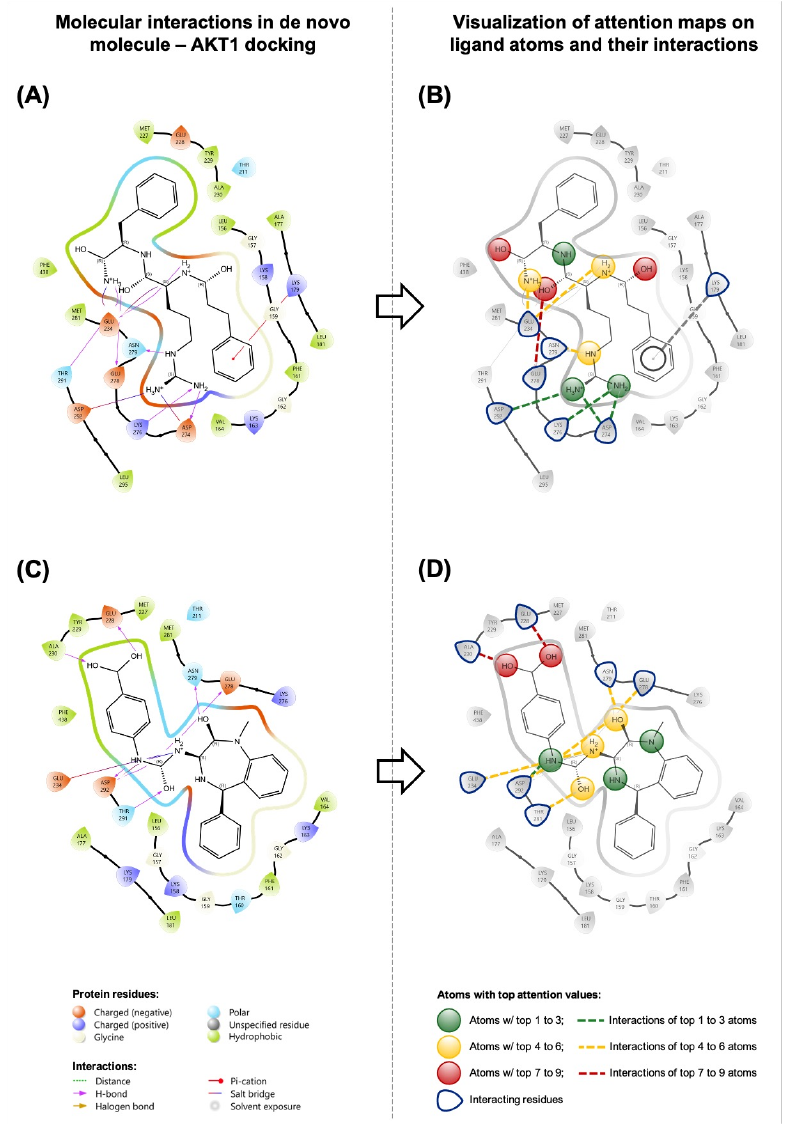}
    
    \label{fig:img10}
\end{figure}
\clearpage
\let\nobreakspace\relax   
\captionof{figure}{Visualisation of DrugGEN attention maps on two additional de novo molecules. The \textbf{left}-side column depicts protein-ligand interaction diagrams obtained from the molecular docking of DrugGEN’s three de novo molecules (in A, C and E) with the AKT1 protein structure (PDB ID: 4GV1). The docked ligands are located in the binding pocket, with interactions between residues and the ligand shown as lines, coloured by the interaction type. Protein residues are coloured according to their physicochemical properties. The \textbf{right}-side column depicts the attention maps of the same three de novo molecules (in B, D and F) retrieved from the graph transformer module of the DrugGEN generator network. Atoms that receive the highest attention scores are highlighted with colours (green: 1st, 2nd and 3rd atoms; yellow: 4th, 5th and 6th atoms; and red: 7th, 8th and 9th atoms with the highest attention scores). Atoms with lower attention scores are not coloured. In the right-side plots, receptor-ligand interactions (i.e., those obtained from the docking analysis) are represented by dashed lines in the attention-score-based colour of the molecule atom involved in the respective interaction. If an interacting atom could not be retrieved (i.e., the atom received a low attention score), its interaction is given in grey colour. \textbf{(A, B)} molecule id: $MOL\_02\_045598$, docking score: -9.055 kcal/mol, \textbf{(C, D)} molecule id: $MOL\_02\_000546$, docking score: -9.051 kcal/mol.}

\setcounter{figure}{10}

\clearpage
\begin{figure}[h]
    \vspace{15em}
    \centering
    \includegraphics[width=\textwidth,clip=True]{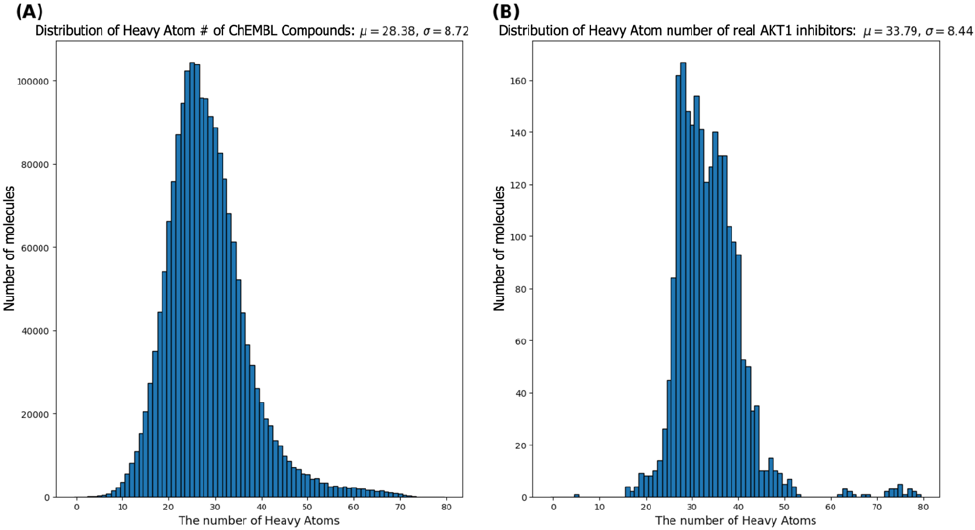}
    \let\nobreakspace\relax
    \caption{The heavy atom distribution histogram of \textbf{(A)} all of the small molecules in the ChEMBL database, and \textbf{(B)} recorded AKT1 binding small molecules in the ChEMBL database (v30).}
    
    \label{fig:img11}
\end{figure}

\setcounter{figure}{11}

\begin{figure}[h]
    \centering
    \includegraphics[width=\textwidth,clip=True]{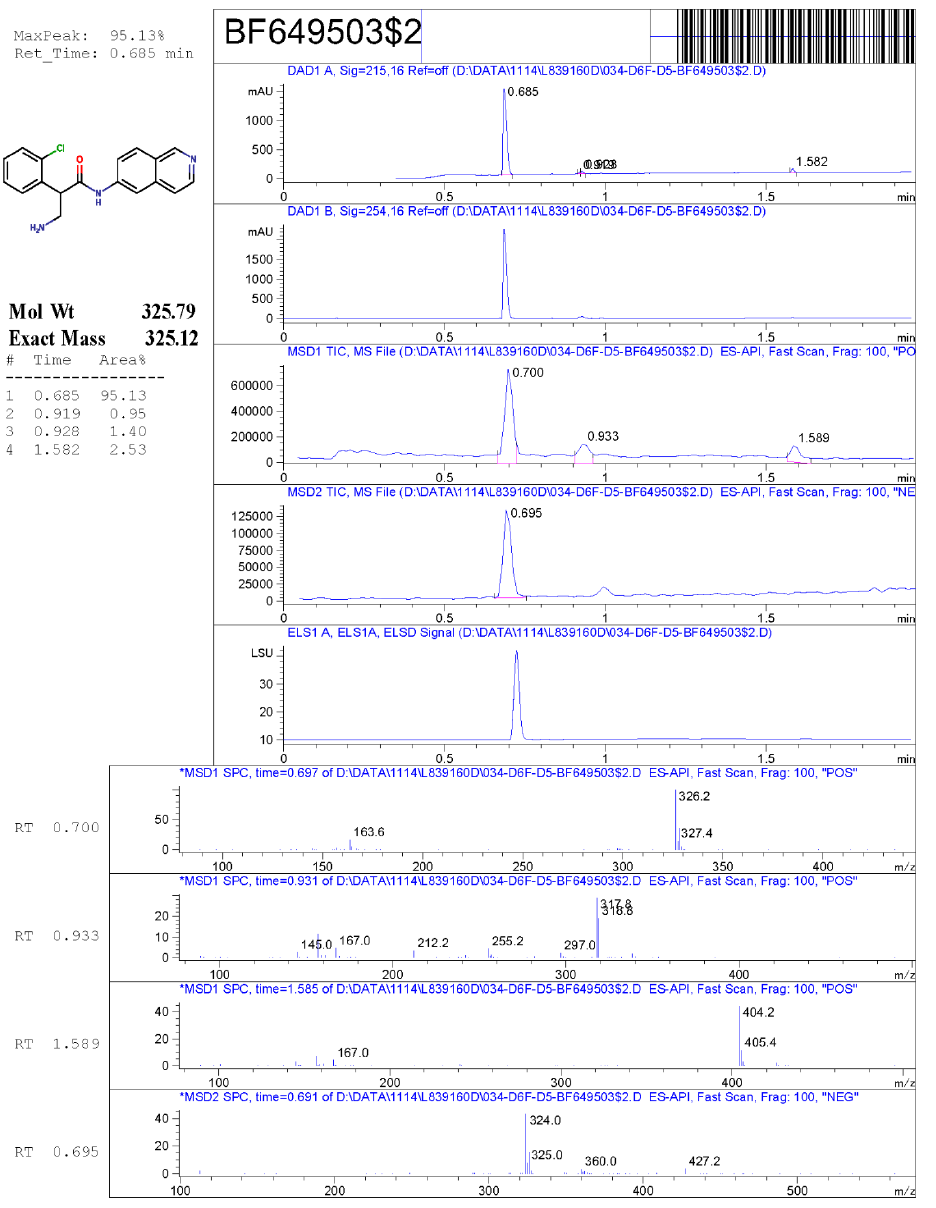}
    \let\nobreakspace\relax
    \caption{\raggedright LC-MS analysis result of \textbf{Molecule 1}.}
    
    \label{fig:img12}
\end{figure}

\setcounter{figure}{12}

\begin{figure}[h]
    \centering
    \includegraphics[width=\textwidth,clip=True]{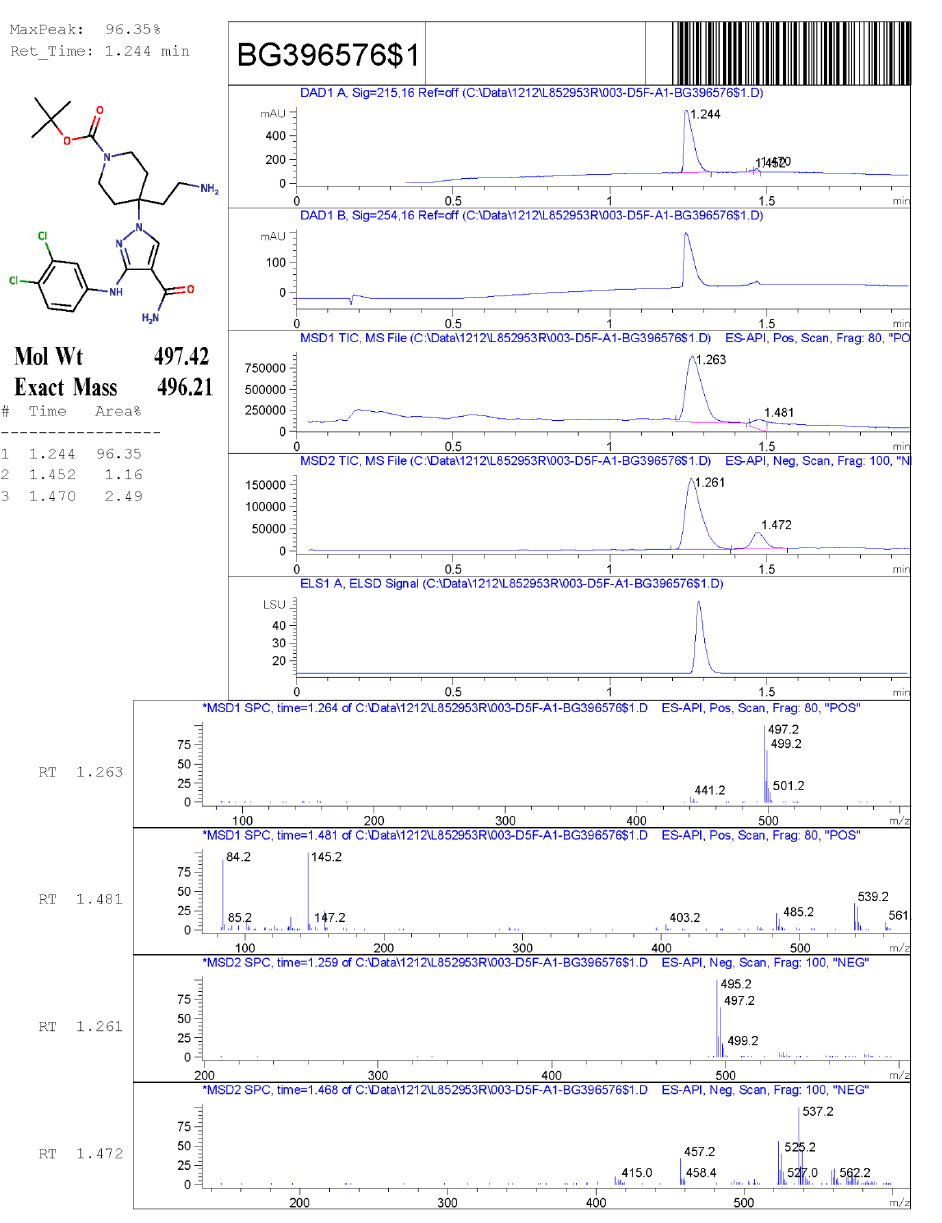}
    \let\nobreakspace\relax
    \caption{\raggedright LC-MS analysis result of \textbf{Molecule 2}.}
    
    \label{fig:img13}
\end{figure}

\setcounter{figure}{13}

\begin{figure}[h]
    \centering
    \includegraphics[width=\textwidth,clip=True]{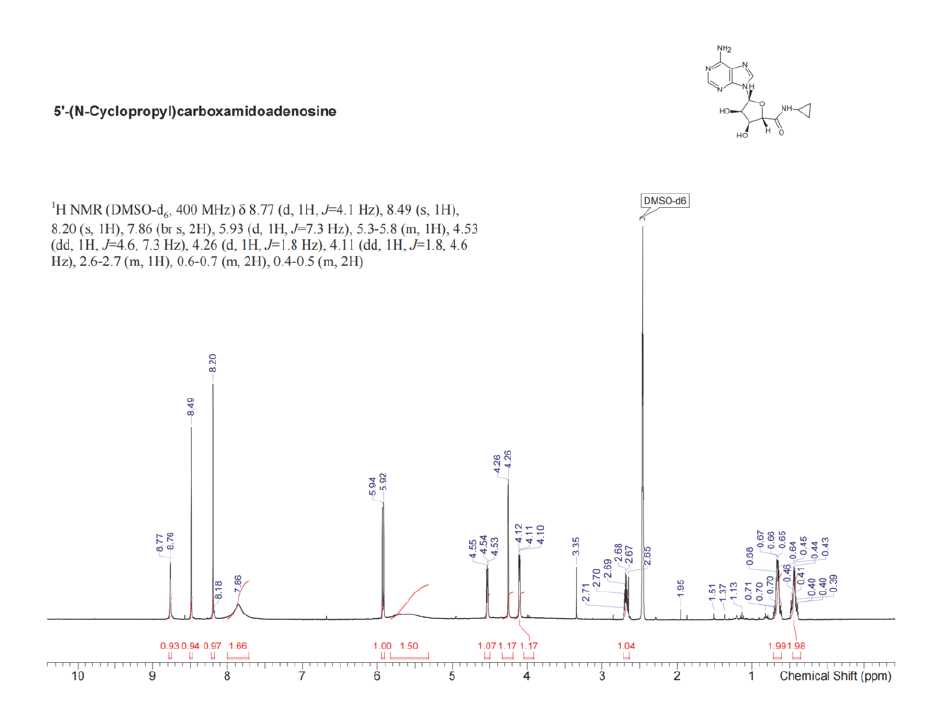}
    \let\nobreakspace\relax
    \caption{\raggedright \textsuperscript{1}H-NMR analysis result of \textbf{Molecule 3}.}
    
    \label{fig:img14}
\end{figure}

\setcounter{figure}{14}

\begin{figure}[h]
    \centering
    \includegraphics[width=\textwidth,clip=True]{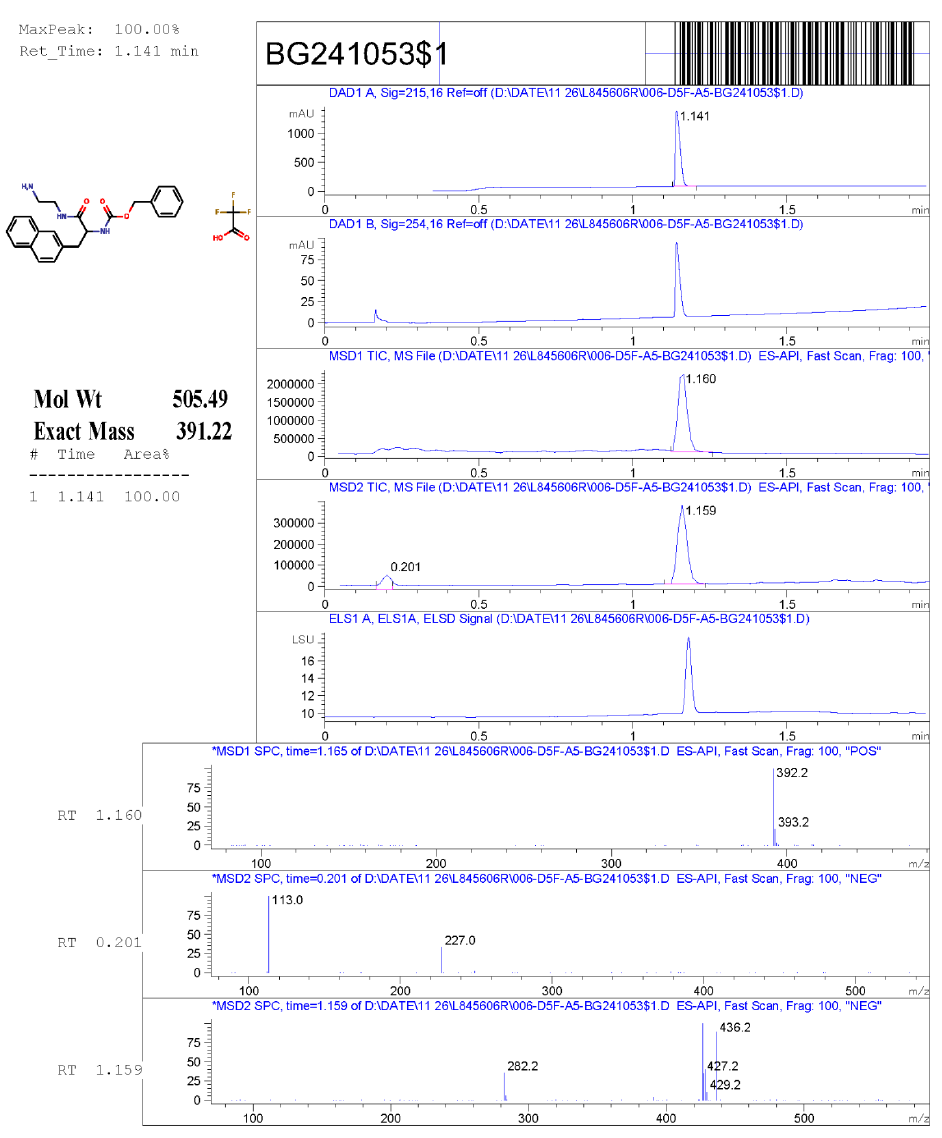}
    \let\nobreakspace\relax
    \caption{\raggedright LC-MS analysis result of \textbf{Molecule 4}.}
    
    \label{fig:img15}
\end{figure}

\setcounter{figure}{15}

\begin{figure}[h]
    \centering
    \includegraphics[width=\textwidth,clip=True]{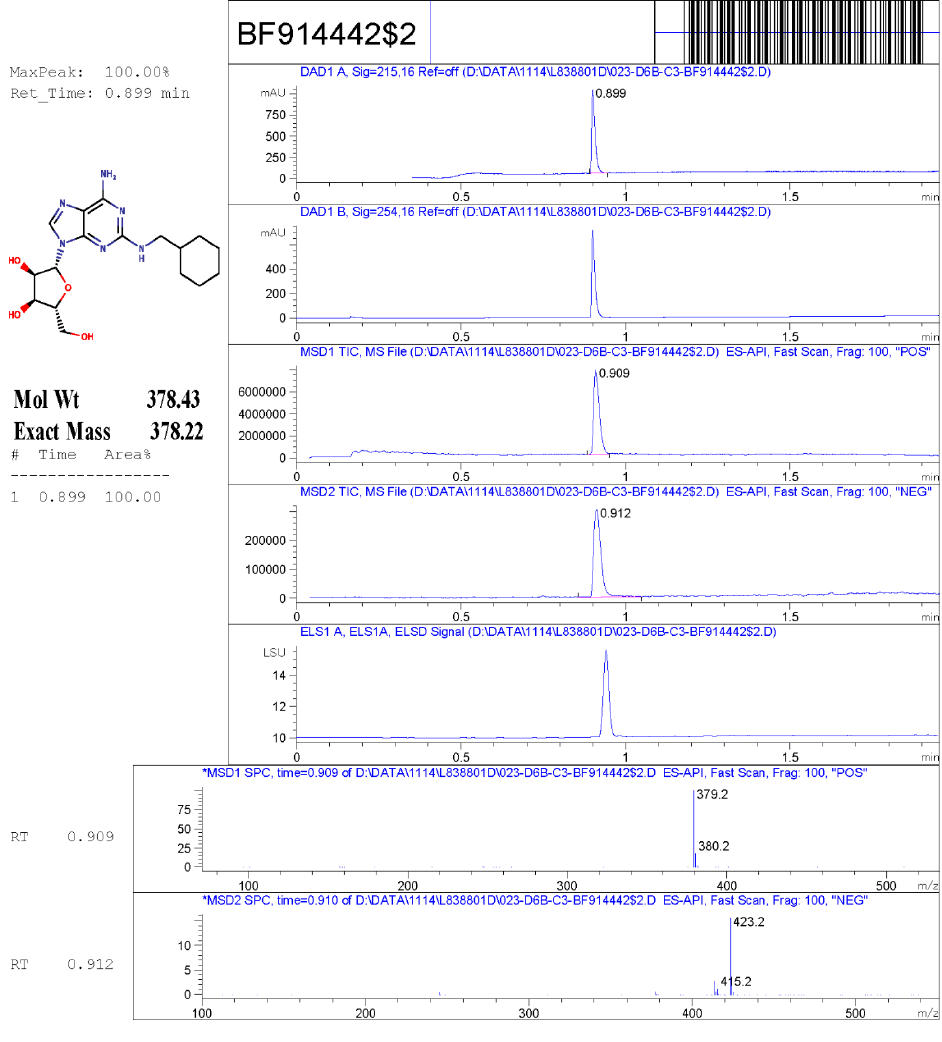}
    \let\nobreakspace\relax
    \caption{\raggedright LC-MS analysis result of \textbf{Molecule 5}.}
    
    \label{fig:img16}
\end{figure}

\setcounter{figure}{16}

\begin{figure}[h]
    \centering
    \includegraphics[width=\textwidth,clip=True]{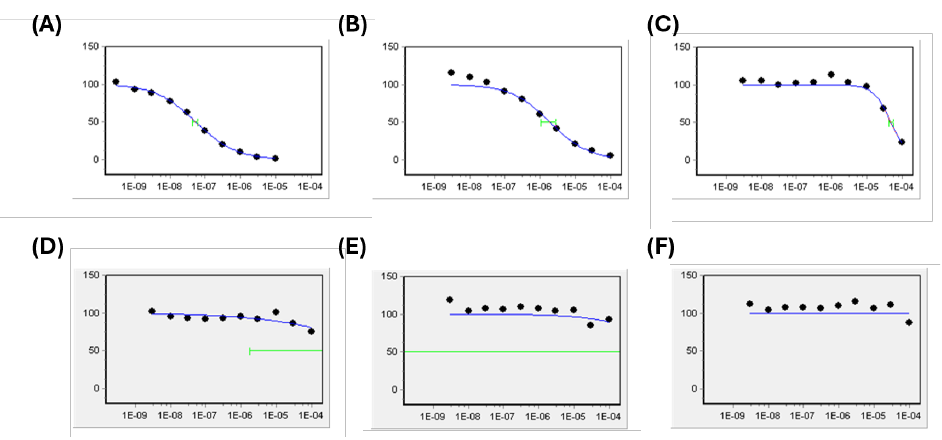}
    \let\nobreakspace\relax
    \caption{IC\textsubscript{50} curves of the compounds \textbf{(A)} Staurosporine (positive control), \textbf{(B)} Molecule 1, \textbf{(C)} Molecule 2, \textbf{(D)} Molecule 3, \textbf{(E)} Molecule 4, \textbf{(F)} Molecule 5 achieved via radiometric AKT1 binding assay. The plots with gray background indicate compounds that are inactive up to the point of 100 . X axis shows the molar concentration of the compounds and Y axis shows the residual activity of the compounds against the enzyme.}
    
    \label{fig:img17}
\end{figure}

\renewcommand{\tablename}{Table S}

\clearpage

\section*{Supplementary Tables}
\renewcommand{\theadfont}{\fontsize{7}{10}}
\begin{table}[h]
\setlength\tabcolsep{2pt}
\let\nobreakspace\relax
\caption{Ablation study performance results in terms of (i) fundamental benchmarking metrics (i.e., validity, novelty, uniqueness, and IntDiv), (ii) metrics measuring physicochemical and drug-likeness-related properties (i.e., QED, and SA), and novelty at inference (NI), which is a novelty metric measured against the starting molecules during inference, whereas the regular novelty is measured against the training dataset. NI is given together with the regular novelty (in parenthesis). Arrows show the direction in which the performance increases.}
    \centering
    \fontsize{7}{10}\selectfont
    \makebox[\linewidth]{
    \begin{tabular}{ccccccccc}
    \hline
    \thead{Model \\ objective /\\ data \\ properties} & \thead{Model/dataset\\ name} & \thead{Highlight of the\\ Model/Dataset} & \thead{Validity\\ ($\uparrow$)} & \thead{Novelty (NI*)\\ ($\uparrow$)} & \thead{Uniq.\\ ($\uparrow$)} & \thead{IntDiv\\ ($\uparrow$)} & \thead{QED\\ ($\uparrow$)} & \thead{SA\\ ($\downarrow$)} \\
    \hline
    
    \multirow{2}{*}{\thead{Training \\ Dataset\\(real \\ molecules)}}
    & ChEMBL molecules & \thead{Training data\\(random mol.*)} & - & - & - & 0.879 ± 0.018 & 0.537± 0.223 & 3.008 ± 0.965 \\
    & Real AKT1 inhibitors & \thead{Training data\\(targeted mol.)} & - & - & - & 0.862 ± 0.022 & 0.552 ± 0.189 & 3.225 ± 0.833 \\
    \hline
    
    \multirow{24}{*}{\thead{Targeted\\molecule \\ design\\(AKT1 \\ targeting)}}
    & DrugGEN & \thead{Targeted gen. \\ via graph\\transformer with \\ modified attention \\ (att * edge * (edge + 1))} & 0.713 & \textbf{0.993 (0.924)} & \textbf{0.995} & \textbf{0.878 ± 0.019} & 0.512 ± 0.175 & 2.968 ± 0.862 \\
    & \thead{DrugGEN\\-att-FiLM} & \thead{Attention changed \\ to FiLM-inspired\\modulation \\ (att * edge\_mul + edge\_add)} & 0.110 & 0.988 (0.503) & \textbf{0.996} & \textbf{0.881 ± 0.020} & \textbf{0.576 ± 0.158} & \textbf{2.847 ± 0.852} \\
    & \thead{DrugGEN\\-att-edge} & \thead{Attention changed \\ to att * edge\\(official graph \\ transformer;\\Dwivedi and Bresson, \\ 2021)} & 0.464 & \textbf{0.991 (0.527)} & \textbf{0.994} & \textbf{0.879 ± 0.018} & 0.537 ± 0.180 & 3.025 ± 0.897 \\
    & \thead{DrugGEN\\-att-edge+1} & \thead{Attention changed to\\att * (edge + 1)} & 0.117 & 0.961 (0.677) & 0.988 & \textbf{0.885 ± 0.019} & \textbf{0.580 ± 0.174} & \textbf{2.818 ± 0.929} \\
    & \thead{DrugGEN\\-att-noedge} & \thead{Attention without \\ any edge-based \\ modification} & 0.081 & 0.984 (0.959) & \textbf{0.995} & 0.850 ± 0.032 & 0.496 ± 0.143 & \textbf{2.841 ± 0.932} \\
    & \thead{MLP* \\(baseline)} & \thead{Targeted gen. \\ via MLPs} & 0.512 & \textbf{0.992 (0.952)} & \textbf{1.000} & 0.857 ± 0.023 & 0.545 ± 0.187 & 3.117 ± 0.923 \\
    & \thead{GCN* \\(baseline)} & \thead{Targeted gen. \\ via GCNs} & \textbf{0.754} & \textbf{0.999 (1.000)} & \textbf{0.997} & 0.849 ± 0.022 & 0.271 ± 0.173 & 5.215 ± 0.912 \\
    \hline
    
    \multirow{6}{*}{\thead{Non- \\ targeted\\molecule \\ design}}
    & \thead{DrugGEN- \\NoTarget} & \thead{Non-targeted gen. \\ via graph transformer \\ with modified attention \\ (att * edge * (edge + 1))} & 0.811 & \textbf{0.990 (0.552)} & \textbf{1.000} & \textbf{0.874 ± 0.020} & 0.502 ± 0.214 & 3.584 ± 0.875 \\
    & \thead{MLP-NoTarget \\ (baseline)} & \thead{Non-targeted gen. \\ via MLPs} & \textbf{0.839} & \textbf{0.999 (0.201)} & \textbf{0.998} & \textbf{0.879 ± 0.018} & \textbf{0.576 ± 0.199} & \textbf{3.215 ± 0.951} \\
    & \thead{GCN-NoTarget \\ (baseline)} & \thead{Non-targeted gen. \\ via GCNs} & 0.639 & \textbf{1.000 (1.000)} & \textbf{0.991} & 0.800 ± 0.025 & 0.135 ± 0.043 & 6.429 ± 0.429 \\
    \hline
    \end{tabular}}
    {\raggedright \footnotesize *mol: molecules, NI: Novelty at Inference (novelty measured against the real molecules used as input at inference), MLP: multi-layered perceptron, GCN: graph convolutional network.\par}  
    \hrule   
    \label{tab:tableS1}
\end{table}

\begin{table}[h]
    \let\nobreakspace\relax
    \caption{The number of transformer encoder blocks (depth) optimisation results for the DrugGEN model. Scores are provided in terms of percent-based performance loss (-) or gain (+) over the default model.}
    \centering
    \makebox[\linewidth]{
    \begin{tabular}{ccccc}

    \hline
    
     & 1-Depth  &  2-Depth & 4-Depth & 6-Depth  \\

    \hline
    
        Validity & -	& -35\%	& -34\%	& -89\% \\

    \hline
    
        Uniqueness	&-	& 0\%	& -4\%	& -8\% \\

    \hline
    
        Novelty	&-	& -2\%	& -3\%	& +9\% \\
    
    \hline    
    \end{tabular}}    
    \label{tab:tableS2}
\end{table}

\begin{table}[h]
\setlength\tabcolsep{4pt}
\let\nobreakspace\relax
\caption{IC50 values of the de novo generated compounds simulated with AKT1. Mol. ID represents the unique identifier assigned during molecule generation, while Exp. Mol. ID refers to the experimental molecule ID given to the synthesized molecules. Hill slope higher than -0.4, i.e. curve is not sigmoidal or very flat or not descending.}
    \centering
    \fontsize{8}{10}\selectfont
    \makebox[\linewidth]{
    \begin{tabular}{cccc}
    \hline
    \thead{Exp. Mol. ID} & \thead{Mol. ID} & \thead{IC50 ($\mu$M)} & \thead{Hill slope} \\
    \hline
    1 & MOL\_02\_045762 & 1.89 & -0.79 \\
    \hline
    2 & MOL\_02\_045795 & 48.55 & -1.74 \\
    \hline
    3 & MOL\_02\_045627 & >100 & 0.00 \\
    \hline
    4 & MOL\_02\_045609 & >100 & -0.51 \\
    \hline
    5 & MOL\_02\_045758 & >100 & -0.29 \\
    \hline
    \end{tabular}}
    \label{tab:tableS3}
\end{table}